\documentclass[11pt]{article}

\usepackage[preprint]{acl}
\usepackage{algorithm}
\usepackage{algorithmic}
\usepackage{times}
\usepackage{latexsym}
\usepackage{bbm}
\usepackage[T1]{fontenc}

\usepackage[utf8]{inputenc}

\usepackage{microtype}

\usepackage{inconsolata}

\usepackage{graphicx}
\usepackage{booktabs}
\usepackage{amsmath,amssymb}
\usepackage{multirow}
\title{From Sampled Evidence to Radiology Reports:
Objectives, Constraints, and Realization}

\author{ 
	Kun Zhao\textsuperscript{1},
    \textbf{Haoteng Tang}\textsuperscript{2},
	\textbf{Chenghua Lin}\textsuperscript{3},
    \textbf{Liang Zhan}\textsuperscript{1}\space\\ 
	\textsuperscript{1}University of Pittsburgh\vspace{-0.5mm} \space
	\textsuperscript{2}University of Texas Rio Grande Valley\vspace{-0.5mm} \space
    \textsuperscript{3}The University of Manchester\vspace{-0.5mm} \\
	\texttt{
		\{kun.zhao, liang.zhan\}@pitt.edu, 
	}\vspace{-0.5mm} 
	\texttt{
		haoteng.tang@utrgv.edu 
	}\vspace{-0.5mm} \\
	\vspace{-0.5mm}
	\texttt{
		chenghua.lin@manchester.ac.uk
	}\vspace{-0.5mm}
}

\providecommand{\rqcell}[2]{}
\begin{document}
\raggedbottom
\maketitle

\title{From Sampled Evidence to Radiology Reports:
Objectives, Constraints, and Realization}


\begin{abstract}
Sampling multiple reports creates an inference problem:
how should sampled evidence be converted into a final
decision, and does a better structured decision yield
a better report?
We study these questions through a framework that
separates the decision signal, scoring objective,
and feasible decision set, followed by text realization.
On MIMIC-CXR and IU-Xray, controlled comparisons
across 13 generator configurations per benchmark
under grouped development/test evaluation reveal
three findings.
First, changing the objective under fixed evidence
and rule capacity changes predictions and their
held-out metric trade-offs.
Second, relaxing candidate constraints has
objective-dependent effects under shared signals
and frozen scores.
Diagnostics connect the IU macro-oriented assertion
increase to near-zero label thresholds and show
that limited candidate coverage coexists with
selection headroom.
Third, a controlled realization study with
SFT-Qwen3-VL-8B shows that decision-level gains
can be attenuated, amplified, or reversed.
Change decomposition reveals that realization
can remove incorrect assertions while losing
correct findings, so target fidelity and reference
accuracy need not improve together.
Finding gains also do not consistently improve
reference-based report quality.
These results motivate separate evaluation of
decision rules, candidate constraints, and
realized reports.
\end{abstract}

\setlength{\itemsep}{0pt}
\setlength{\itemsep}{0pt}

\section{Introduction}
\label{sec:introduction}
Sampling multiple outputs from a fixed generator is
a common approach to improving inference.
Self-consistency aggregates agreement across sampled
reasoning paths~\citep{wang2023selfconsistency},
while minimum Bayes risk (MBR) decoding selects outputs
according to expected utility~\citep{kumar2004minimum,
eikema2022sampling}.
Quality-aware decoding introduces additional selection
signals~\citep{fernandes2022quality}, and recent analyses
highlight the roles of utility bias and sample
diversity~\citep{kovacs-etal-2024-mitigating,
kamigaito-etal-2025-diversity}.
In radiology report generation (RRG), Clinical Consensus
Selection (CCS) and Set-Distance Rewards (SDR) similarly
exploit clinical agreement or representation distances
across sampled reports~\citep{zhang2026ccs,gulluk2026sdr}.
These approaches motivate a broader question:
how should sampled evidence be converted into
a final prediction?

A sampled pool does not uniquely determine that prediction.
First, the same CheXpert finding vote rates can support
different predictions when decision rules are selected
for majority agreement, macro-F1, or micro-F1.
F-measure optimization provides established motivation
for objective-dependent decision
thresholds~\citep{lipton2014thresholding,
koyejo2014consistent}.
Second, candidate selection can return only an existing
report, whereas predicting findings individually can
produce a label combination absent from the pool.
Lexically distinct reports may express identical findings,
so textual diversity need not expand the clinical
combinations available to selection, consistent with
the broader distinction between lexical and semantic
diversity~\citep{kuhn2023semantic}.
Third, realizing a predicted finding vector as text
may omit requested findings or introduce additional
assertions.
Prior RRG research has consequently emphasized factual
completeness and consistency~\citep{miura-etal-2021-improving}
and developed clinical evaluation methods beyond lexical
overlap~\citep{yu2023evaluating,
ostmeier-etal-2024-green,zhao-etal-2024-ratescore}.
Finding accuracy, target fidelity, and reference-based
report quality therefore warrant separate evaluation.

Method-level comparisons can obscure these distinctions
because inference procedures often change their signals,
scoring rules, output restrictions, and realization
procedures simultaneously.
Recent RRG methods also improve reports through
multi-objective training or retrieval-conditioned
generation~\citep{xiao-etal-2025-online,
park-etal-2026-ra}, making complementary changes
to different components of the pipeline.
A performance difference between systems therefore
does not identify which component produced it.
Likewise, an improvement in finding-level F1
does not establish an improvement in the final report.
These considerations motivate a central inquiry:
\emph{How do inference-time decision objectives,
candidate constraints, and text realization jointly
determine the quality of outputs produced from
the same sampled evidence?}
We separate it into three research questions:
\begin{itemize}
\setlength{\itemsep}{0pt}
\setlength{\parsep}{0pt}
\setlength{\itemsep}{0pt}
    \item \textbf{RQ1: Objective choice.}
    Given the same sampled evidence, how does choosing
    a different objective affect predictions and performance?

    \item \textbf{RQ2: Candidate constraints.}
    How does the effect of restricting predictions
    to finding combinations represented in the sampled
    pool depend on the scoring objective?

    \item \textbf{RQ3: Text realization.}
    How faithfully do retrieval and generation express
    predicted findings, and do improvements in finding
    predictions persist in the resulting reports?
\end{itemize}

To address these questions, we separate sampling-based
inference into the \emph{decision signal}, or information
used for prediction; the \emph{scoring objective},
which evaluates possible outputs; and the
\emph{feasible decision set}, which specifies permitted
outputs.
For finding-level predictions, \emph{text realization}
then converts the selected findings into a report.
This framework accommodates both report selection
and finding prediction, while guiding comparisons
that isolate particular design choices.

We study MIMIC-CXR and IU-Xray under grouped
development/test evaluation.
Across 13 generator configurations per benchmark,
RQ1 varies shared-threshold objectives and candidate
utilities, while RQ2 compares constrained, partial,
and free predictions under shared signals and
frozen objective-specific scores.
Partial replacement applies free predictions only
to five labels with the lowest training prevalence.
For RQ3,
we apply the RQ2 decision rules to obtain structured finding targets, and realize these targets through retrieval or generation using a fine-tuned Qwen3-VL-8B.

Our results reveal objective-dependent effects
of candidate relaxation, with diagnostics separating
threshold behavior, pool coverage, and selection
headroom.
Realization can correct false positives while losing
true positives, so fidelity and reference accuracy
need not improve together.
These findings motivate separate evaluation of
structured decisions and final reports.

Our contributions are:
\begin{itemize}
\setlength{\itemsep}{0pt}
\setlength{\parsep}{0pt}
\setlength{\itemsep}{0pt}

\item We develop a controlled inference framework that separates decision signals, scoring objectives, feasible decision sets, and text realization, enabling component-level comparisons under fixed sampled evidence.

\item Across 13 generator configurations per benchmark, we demonstrate objective-dependent effects of candidate relaxation: improving surrogate scores need not improve held-out F1, with diagnostics identifying threshold-driven assertion inflation, candidate coverage limitations, and selection headroom.

\item We show that finding-level gains can be attenuated, amplified, or reversed during report realization. Reference-aware change decomposition further reveals that correcting false assertions can coincide with losing correct findings, separating target fidelity from reference accuracy and report quality.

\end{itemize}

\begin{figure*}[t]
  \centering
  \includegraphics[width=0.6\linewidth]{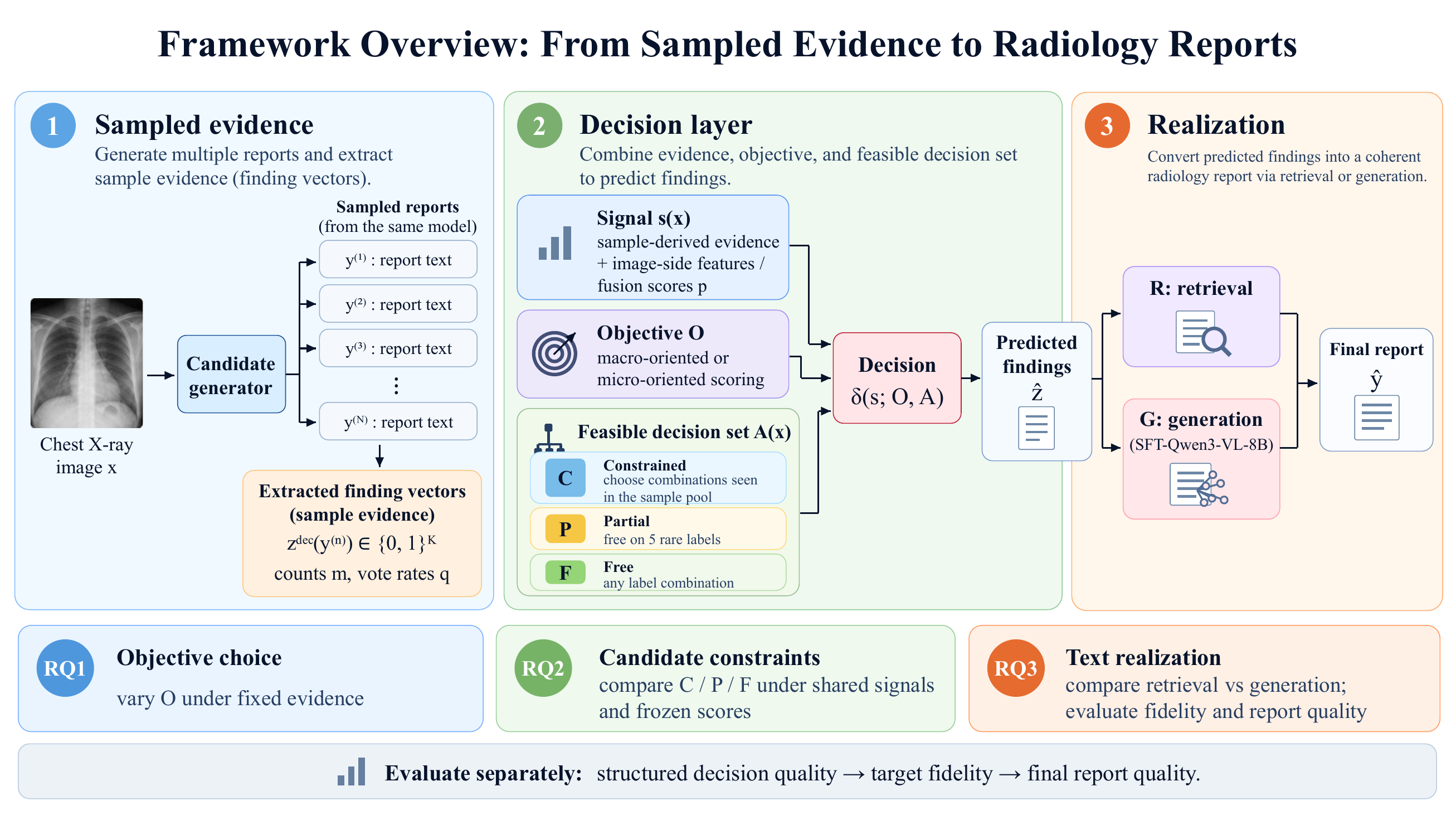}
    \caption{Overview of the proposed framework. Multiple reports are sampled to extract finding-level evidence. The framework separates decision signals, scoring objectives, and feasible decision sets to produce structured predictions. RQ1 studies objective choice, RQ2 examines candidate constraints (constrained, partial, and free), and RQ3 evaluates report realization through retrieval and generation. Decision accuracy, target fidelity, and report quality are assessed separately.}
   \label{fig:framework}
\end{figure*}

\section{Related Work}
\label{sec:related_work}

\paragraph{Sampling-based inference and decision objectives.}
Minimum Bayes risk (MBR) decoding selects outputs
by expected utility~\citep{kumar2004minimum}.
Subsequent work examines sampling approximations,
utility bias, diversity, and uncertainty~\citep{
eikema2022sampling,freitag2023epsilon,
ohashi2024true,jinnai2024model,
kamigaito-etal-2025-diversity,
kovacs-etal-2024-mitigating,ICLR2025_1588dc2b}.
Self-consistency and voting admit MBR interpretations
under specific utilities~\citep{wang2023selfconsistency,
bertsch2023mbr}, while quality-aware and structured
selection introduce additional signals~\citep{
fernandes2022quality,farinhas2024reranking,
lyu2025source,eikema2025structure}.
Established F-measure theory motivates our decision
objectives~\citep{dembczynski2013optimizing,
waegeman2014bayes,lipton2014thresholding,
koyejo2014consistent,narasimhan2015optimizing};
we examine their effects under fixed sampled evidence.

\paragraph{Sampling and structured generation in RRG.}
Sampled reports support uncertainty estimation
and clinical consensus or distance-based selection
\citep{wang2025semantic,zhang2026ccs,gulluk2026sdr}.
Lexical diversity need not imply semantic diversity
\citep{kuhn2023semantic}, motivating our analysis
of candidate finding combinations.
Related methods improve factuality through rewards,
alignment, and editing~\citep{miura-etal-2021-improving,
delbrouck-etal-2022-improving,xiao-etal-2025-online,
zhang-etal-2025-editgrpo,zhou-etal-2026-enhancing},
or structure generation using report organization
and retrieved phrases~\citep{delbrouck-etal-2025-automated,
nishino-etal-2022-factual,park-etal-2026-ra}.
We isolate objective choice, candidate constraints,
and realization while holding candidate pools fixed.

\paragraph{Clinical evaluation.}
CheXbert, RadGraph, GREEN, and RaTEScore assess
finding agreement, clinical entities and relations,
or report errors beyond lexical overlap~\citep{
smit-etal-2020-combining,jain2021radgraph,
ostmeier-etal-2024-green,zhao-etal-2024-ratescore}.
Work on evaluation artifacts and metric standardization
further highlights the importance of evaluation design
\citep{zhao-etal-2026-x,xu-etal-2025-radeval}.
We distinguish finding accuracy, target fidelity,
and reference-based report quality.
\section{Method}
\label{sec:method}

We separate sampling-based inference into a decision signal,
a scoring objective, and a feasible decision set, followed by
text realization. Controlled comparisons examine objective choice
(RQ1), candidate constraints (RQ2), and transfer to reports (RQ3).

\subsection{Framework and Finding Representations}
\label{sec:framework}

For an image $x$, a fixed generator samples
$\mathcal C(x)=\{y^{(1)},\ldots,y^{(N)}\}$.
We express inference as $\hat a=\delta(s(x);\mathcal O,\mathcal A(x))$,
where $s(x)$ is the decision signal, $\mathcal O$ is the scoring
objective, and $\mathcal A(x)$ is the feasible decision set.
Rule parameters are fitted or selected on development data
and frozen for held-out evaluation.

For $K=14$ findings \cite{irvin2019chexpert}, $z^{\mathrm{dec}}(y)\in\{0,1\}^K$
is a candidate's structured finding vector, when available,
or a vector obtained by clinical labeling.
Separately, $z^{\mathrm{text}}(y)$ is extracted from its narrative.
The two representations need not agree.
Both map positive states to one and negative, uncertain,
and unmentioned states to zero. Reference-uncertain cells
are excluded from state-model fitting and reference-based evaluation.
Constrained and free decision sets are
\begin{equation}
\begin{aligned}
\mathcal A_{\mathrm{con}}^z(x)&=\{z^{\mathrm{dec}}(y):y\in\mathcal C(x)\},\\
\mathcal A_{\mathrm{free}}^z&=\{0,1\}^K.
\end{aligned}
\end{equation}

The sampled reports supply positive vote counts and rates:
\begin{equation}
m_L(x)=\sum_{n=1}^Nz_L^{\mathrm{dec}}(y^{(n)}),
\qquad q_L(x)=m_L(x)/N.
\end{equation}
Vote rates summarize what the sampled reports assert,
rather than the probability that a finding is present.
RQ1 uses these sample-derived signals directly.
For RQ2/RQ3, we fit a separate fusion model for each
finding on development examples with nonuncertain
reference labels.
Each model combines image-side and sample-derived
features to estimate $p_L^{(1)}$, with
$p_L^{(0)}=1-p_L^{(1)}$.
These estimates are complementary by construction,
but are not assumed to be calibrated.
Feature and fitting details appear in
Appendix~\ref{app:state-estimation}.

\subsection{Objective Choice under Fixed Evidence}
\label{sec:objective}

RQ1 varies the objective for shared-threshold finding prediction
and candidate report selection, holding sampled evidence fixed.

\paragraph{Shared-threshold rules.}
Let $m_{iL}=m_L(x_i)$. Within each generator--benchmark condition,
a single threshold is shared across inputs and labels:
\begin{equation}
\hat z_{iL}(\tau)=\mathbf1[m_{iL}\ge\tau],
\qquad \tau\in\{1,\ldots,N\}.
\end{equation}
Majority fixes $\tau_{\mathrm{maj}}=\lfloor N/2\rfloor+1$,
with tied votes yielding non-assertion.
For $M\in\{F_{\mathrm{macro}},F_{\mathrm{micro}}\}$,
we evaluate every threshold on development data and select
one with the highest score:
\begin{equation}
\tau_M^*\in\arg\max_{\tau\in\{1,\ldots,N\}}
M^{\mathrm{dev}}(\hat z(\tau)).
\end{equation}
The threshold is frozen for test. Macro-F1 averages per-label F1;
micro-F1 uses pooled confusion counts. Selection uses fourteen labels.
The macro--micro comparison changes only the development objective,
fixing pools, votes, parameter count, and search space.
Label-specific and constant controls appear in Appendix~\ref{app:rq1-controls}.

\paragraph{Candidate utilities.}
Each sampled report is scored by its average pairwise utility
against the other $N-1$ candidates:
\begin{equation}
\hat j_u\in\arg\max_{j\in\{1,\ldots,N\}}
\frac{1}{N-1}\sum_{j'\ne j}u(y^{(j)},y^{(j')}).
\end{equation}
We return $y^{(\hat j_u)}$ and compare CheXbert finding F1 \cite{smit-etal-2020-combining},
RadGraph F1 \cite{jain2021radgraph}, and ROUGE-L \cite{lin-2004-rouge} utilities under fixed pools,
averaging, and tie-breaking. Selection uses candidate agreement
without held-out references; evaluation scores the selected narrative.

\subsection{Scoring Objectives and Candidate Constraints}
\label{sec:constraints}

RQ2 crosses macro/micro scoring with constrained (C) and free (F) decisions
and additionally evaluates partial replacement (P).
Within each generator--benchmark condition, all arms share fitted
state estimates; objective-specific coefficients are frozen across C/P/F.

\paragraph{Scoring objectives.}
Motivated by F-measure optimization results
\citep{lipton2014thresholding,koyejo2014consistent,narasimhan2015optimizing},
we alternate constrained development selection and coefficient updates.
For selections $z_i$ under the current coefficients, define soft counts
\begin{equation}
\begin{aligned}
T_L&=\sum_{i\in\mathrm{dev}}z_{iL}p_{iL}^{(1)},\\
P_L&=\sum_{i\in\mathrm{dev}}z_{iL}p_{iL}^{(0)},\\
Q_L&=\sum_{i\in\mathrm{dev}}(1-z_{iL})p_{iL}^{(1)}.
\end{aligned}
\end{equation}
These are probability-weighted TP, FP, and FN, not counts
computed directly against references.
For macro scoring, $D_L=2T_L+P_L+Q_L$ gives
\begin{equation}
a_L^{\mathrm{ma}}=\frac{2(P_L+Q_L)}{D_L^2},
\qquad b_L^{\mathrm{ma}}=\frac{2T_L}{D_L^2}.
\end{equation}
The common $1/K$ factor is omitted.
For micro scoring, $T=\sum_LT_L$, $P=\sum_LP_L$,
$Q=\sum_LQ_L$, and $D=2T+P+Q$ give shared coefficients
\begin{equation}
a_L^{\mathrm{mi}}=\frac{2(P+Q)}{D^2},
\qquad b_L^{\mathrm{mi}}=\frac{2T}{D^2}.
\end{equation}
Either objective uses the additive score
\begin{equation}
\begin{aligned}
S(z;p,a,b)=\sum_L\bigl[
&z_L(a_Lp_L^{(1)}-b_Lp_L^{(0)})\\
&-(1-z_L)b_Lp_L^{(1)}\bigr].
\end{aligned}
\end{equation}
This is an F-measure-informed surrogate: an F1 ratio of expected
counts is not the exact expected F1. Derivation appears in
Appendices~\ref{app:local-f1}--\ref{app:action-scores}.

\paragraph{Fitting and inference.}
Starting from $a_L=2$, $b_L=1$, we perform twelve development
iterations and freeze the coefficients that produced the last selection,
not the subsequent update. Normalization, numerical floors, and
update order are specified in Algorithm~\ref{alg:constraints}
and Appendix~\ref{app:coefficient-fitting}.
For $o\in\{\mathrm{ma},\mathrm{mi}\}$,
\begin{equation}
\hat z^{o,\mathrm{con}}\in
\arg\max_{z\in\mathcal A_{\mathrm{con}}^z(x)}S(z;p,a^o,b^o).
\end{equation}
Free maximization is label-wise because the score is additive:
\begin{equation}
\hat z_L^{o,\mathrm{free}}
=\mathbf1[(a_L^o+b_L^o)p_L^{(1)}\ge b_L^op_L^{(0)}].
\end{equation}
Complementarity makes this equivalent to
$p_L^{(1)}\ge b_L^o/(a_L^o+2b_L^o)$.
Development fitting does not guarantee held-out F1 optimality.

\paragraph{Partial replacement.}
Let $R_5$ contain the five lowest-prevalence training labels:
\begin{equation}\label{eq:partial}
\hat z_L^{o,\mathrm{partial}}=
\begin{cases}
\hat z_L^{o,\mathrm{free}},&L\in R_5,\\
\hat z_L^{o,\mathrm{con}},&L\notin R_5.
\end{cases}
\end{equation}
Partial does not reselect the constrained candidate or optimize
an intermediate feasible set. Additivity gives
\begin{equation}
S(\hat z^{o,\mathrm{free}})\ge
S(\hat z^{o,\mathrm{partial}})\ge S(\hat z^{o,\mathrm{con}}),
\end{equation}
checked per input with numerical tolerance.
This ordering need not hold for F1; $R_5$ differs from
the five-label evaluation subset.

\paragraph{Interaction and diagnostics.}
For metric $M$ and $t\in\{\mathrm{partial},\mathrm{free}\}$,
define $\Delta_M^{o,t}=M(\hat z^{o,t})-M(\hat z^{o,\mathrm{con}})$.
The same-metric interaction is
\begin{equation}
I_M^t=\Delta_M^{\mathrm{ma},t}-\Delta_M^{\mathrm{mi},t}.
\end{equation}
It compares the effect of changing the decision condition under
macro versus micro scoring. Bootstrap details appear in
Appendix~\ref{app:statistics-evaluation}.

\subsection{Text Realization and Report Evaluation}
\label{sec:realization}

RQ3 reuses the RQ2 targets from Qwen2.5-VL-7B candidate pools without refitting.
Two objectives, three target constructions, and two routes
produce twelve configurations.

\paragraph{Realization routes.}
For target $\hat z$, route $r$ produces a report and extracted findings:
\begin{equation}
\hat y_r=R_r(x,\hat z),\qquad
\tilde z_r=z^{\mathrm{text}}(\hat y_r).
\end{equation}
We compare retrieval from training findings sections with target- and image-conditioned generation using a fine-tuned Qwen3-VL-8B. Both routes receive the same structured finding targets derived from the fixed Qwen2.5-VL-7B candidate pools.
Our constrained-verbatim control emits the macro-constrained candidate;
its narrative-derived findings may differ from its decision vector.
Unlike retrieval, which searches training reports
for a given target, constrained verbatim directly
returns the selected report from the sampled pool.
Implementation details appear in Appendix~\ref{app:realization-implementation}.

\paragraph{Fidelity and performance transfer.}
Target accuracy, realized finding accuracy, fidelity, and report quality
are evaluated separately. Retention measures preserved target positives;
insertion measures realized positives absent from the target;
Hamming measures disagreement over all fourteen label cells.
Fidelity is pooled over test inputs without reference masking.

For reference-based finding metric $M$, define
$M_{\mathrm{dec}}^o(t)=M(\hat z^{o,t})$ and
$M_r^o(t)=M(\tilde z_r^{o,t})$, suppressing common test references
and masks. Relative to the same-objective constrained baseline $C$,
\begin{equation}
\begin{aligned}
L_r^o(t)&=M_r^o(t)-M_{\mathrm{dec}}^o(t),\\
\Delta M_{\mathrm{dec}}^o(t)&=M_{\mathrm{dec}}^o(t)-M_{\mathrm{dec}}^o(C),\\
\Delta M_r^o(t)&=M_r^o(t)-M_r^o(C),\\
A_r^o(t)&=\Delta M_r^o(t)-\Delta M_{\mathrm{dec}}^o(t).
\end{aligned}
\end{equation}
$L_r$ measures the realization gap; $A_r$ measures the change
in the target contrast after realization. These differ from fidelity:
omitting a false-positive target can improve accuracy while lowering retention.
Reference-masked deletion/addition decomposition and sign conventions
appear in Appendices~\ref{app:rq3-decomposition}
and~\ref{app:statistics-evaluation}, respectively.

\begin{table*}[!t]
\centering
\scriptsize
\setlength{\tabcolsep}{3pt}
\renewcommand{\arraystretch}{1.0}
\renewcommand{\rqcell}[2]{\shortstack{#1\\[-1pt]{\scriptsize\textcolor{black!65}{(#2)}}}}
\begin{tabular*}{\textwidth}{@{\extracolsep{\fill}}lcccc@{\hspace{12pt}}cccc@{}}
\toprule
& \multicolumn{4}{c}{\textbf{MIMIC-CXR}} & \multicolumn{4}{c}{\textbf{IU-Xray}} \\
\cmidrule(lr){2-5}\cmidrule(l){6-9}
Contrast & Ma-14 & Mi-14 & Ma-5 & Mi-5 & Ma-14 & Mi-14 & Ma-5 & Mi-5 \\
\midrule
\shortstack[l]{Ma--Maj\\[-1pt]{\scriptsize\phantom{(0/0/0)}}} & \rqcell{+0.0644}{13/0/0} & \rqcell{-0.0033}{7/0/6} & \rqcell{+0.0590}{13/0/0} & \rqcell{+0.0238}{11/0/2} & \rqcell{+0.0553}{13/0/0} & \rqcell{-0.1423}{1/0/12} & \rqcell{+0.0767}{12/1/0} & \rqcell{+0.1234}{12/1/0} \\
\shortstack[l]{Mi--Maj\\[-1pt]{\scriptsize\phantom{(0/0/0)}}} & \rqcell{+0.0471}{12/0/1} & \rqcell{+0.0245}{11/0/2} & \rqcell{+0.0563}{12/0/1} & \rqcell{+0.0376}{12/0/1} & \rqcell{+0.0125}{6/4/3} & \rqcell{-0.0008}{4/1/8} & \rqcell{+0.0055}{5/6/2} & \rqcell{+0.0160}{6/6/1} \\
\shortstack[l]{Ma--Mi\\[-1pt]{\scriptsize\phantom{(0/0/0)}}} & \rqcell{+0.0172}{12/1/0} & \rqcell{-0.0278}{1/1/11} & \rqcell{+0.0028}{7/1/5} & \rqcell{-0.0138}{4/1/8} & \rqcell{+0.0428}{13/0/0} & \rqcell{-0.1415}{0/0/13} & \rqcell{+0.0712}{12/1/0} & \rqcell{+0.1074}{12/1/0} \\
\bottomrule
\end{tabular*}
\caption{
Shared-threshold comparisons: mean within-generator
paired differences across 13 configurations per benchmark,
with positive/zero/negative counts in parentheses.
Contrasts denote first minus second.
Ma/Mi denote macro-/micro-F1-14-selected shared thresholds;
Maj denotes binary majority.
Metric columns report macro/micro-F1 over 14/five labels.
}
\label{tab:rq1-thresholds}
\end{table*}

\begin{table*}[!t]
\centering
\scriptsize
\setlength{\tabcolsep}{3pt}
\renewcommand{\arraystretch}{1.0}
\renewcommand{\rqcell}[2]{\shortstack{#1\\[-1pt]{\scriptsize\textcolor{black!65}{(#2)}}}}
\begin{tabular*}{\textwidth}{@{\extracolsep{\fill}}llccccccccc@{}}
\toprule
Data & Contrast & Ma-14 & Mi-14 & Ma-5 & Mi-5 & RG & RL & GREEN & RaTE & B4 \\
\midrule
\multirow{3}{*}{\textbf{MIMIC}} & \shortstack[l]{RG--CX\\[-1pt]{\scriptsize\phantom{(0/0/0)}}} & \rqcell{-0.0223}{0/0/13} & \rqcell{-0.0282}{1/0/12} & \rqcell{-0.0289}{0/0/13} & \rqcell{-0.0332}{1/0/12} & \rqcell{+0.0428}{13/0/0} & \rqcell{+0.0227}{12/0/1} & \rqcell{+0.0069}{11/0/2} & \rqcell{+0.0139}{12/0/1} & \rqcell{+0.0079}{10/0/3} \\
 & \shortstack[l]{RL--CX\\[-1pt]{\scriptsize\phantom{(0/0/0)}}} & \rqcell{-0.0243}{0/0/13} & \rqcell{-0.0355}{1/0/12} & \rqcell{-0.0333}{0/0/13} & \rqcell{-0.0436}{0/0/13} & \rqcell{+0.0183}{13/0/0} & \rqcell{+0.0369}{12/0/1} & \rqcell{+0.0035}{9/0/4} & \rqcell{+0.0086}{11/0/2} & \rqcell{+0.0146}{13/0/0} \\
 & \shortstack[l]{RG--RL\\[-1pt]{\scriptsize\phantom{(0/0/0)}}} & \rqcell{+0.0020}{7/0/6} & \rqcell{+0.0073}{10/0/3} & \rqcell{+0.0044}{10/0/3} & \rqcell{+0.0104}{12/0/1} & \rqcell{+0.0246}{13/0/0} & \rqcell{-0.0142}{0/0/13} & \rqcell{+0.0033}{10/0/3} & \rqcell{+0.0053}{13/0/0} & \rqcell{-0.0067}{0/0/13} \\
\addlinespace[3pt]
\multirow{3}{*}{\textbf{IU}} & \shortstack[l]{RG--CX\\[-1pt]{\scriptsize\phantom{(0/0/0)}}} & \rqcell{+0.0012}{6/3/4} & \rqcell{+0.0108}{5/3/5} & \rqcell{-0.0015}{2/9/2} & \rqcell{+0.0021}{2/9/2} & \rqcell{+0.0715}{13/0/0} & \rqcell{+0.0188}{13/0/0} & \rqcell{+0.0085}{8/0/5} & \rqcell{+0.0337}{13/0/0} & \rqcell{+0.0153}{12/0/1} \\
 & \shortstack[l]{RL--CX\\[-1pt]{\scriptsize\phantom{(0/0/0)}}} & \rqcell{+0.0028}{5/5/3} & \rqcell{+0.0052}{5/5/3} & \rqcell{-0.0008}{1/9/3} & \rqcell{-0.0019}{2/9/2} & \rqcell{+0.0364}{11/0/2} & \rqcell{+0.0316}{13/0/0} & \rqcell{+0.0030}{8/0/5} & \rqcell{+0.0158}{12/0/1} & \rqcell{+0.0187}{13/0/0} \\
 & \shortstack[l]{RG--RL\\[-1pt]{\scriptsize\phantom{(0/0/0)}}} & \rqcell{-0.0017}{4/4/5} & \rqcell{+0.0056}{5/4/4} & \rqcell{-0.0007}{2/9/2} & \rqcell{+0.0040}{3/9/1} & \rqcell{+0.0351}{13/0/0} & \rqcell{-0.0129}{1/0/12} & \rqcell{+0.0055}{7/0/6} & \rqcell{+0.0179}{12/0/1} & \rqcell{-0.0034}{5/0/8} \\
\bottomrule
\end{tabular*}
\caption{
Candidate-utility comparisons across 13 configurations
per benchmark. Each cell reports the mean paired
difference and positive/zero/negative counts.
Contrasts denote first minus second;
CX/RG/RL denote CheXbert-14/RadGraph/ROUGE-L utilities.
Ma/Mi denote macro/micro-F1 over 14/five labels;
RaTE/B4 denote RaTEScore/BLEU-4.
}
\label{tab:rq1-utilities}
\end{table*}

\section{Experimental Setup}
\label{sec:experimental-setup}

\paragraph{Data and generators}
We evaluate MIMIC-CXR \cite{johnson2019mimic} on 2,047 images (615 dev/1,432 test)
with patient-disjoint splits, and IU-Xray \cite{demnerfushman2016preparing} on 638 images
(192 dev/446 test) with study-disjoint splits.
Training and evaluation patients/studies are disjoint.
Data and split details appear in Appendix~\ref{app:data-representations}.
Both datasets use the same 13 generator configurations,
covering instruction-tuned VLMs, supervised and RL variants,
RRG systems, and LLaVA backbones.
IU configurations use IU-trained weights.
Unless stated otherwise, we sample $N=32$ reports at
temperature $0.5$.
All dev-fitted components and configuration choices are
frozen before test evaluation.

\paragraph{Baselines.}
For RQ3, Qwen2.5-VL-7B serves exclusively as the candidate generator, producing 32 sampled reports per image, while the fine-tuned Qwen3-VL-8B serves as the text realization model. The two models have distinct roles, and structured targets are fixed before realization.
RQ3 compares all twelve realization configurations
with greedy decoding, MBR, CCS, and SDR, alongside
our constrained-verbatim internal control.
Development-selected variants are chosen using
development Ma-14 and frozen before test evaluation.
Additional baselines and development-selection records
appear in Appendices~\ref{app:rq3-benchmark}
and~\ref{app:dev-selection}.

\smallskip
\noindent\textbf{Evaluation.}\enspace
We evaluate structured targets and narrative-extracted
findings using macro/micro F1 over fourteen labels
and a fixed five-label subset.
Realization fidelity is measured by retention,
insertion, and Hamming disagreement.
Report metrics include GREEN \cite{ostmeier-etal-2024-green}, RadGraph, RaTEScore \cite{zhao-etal-2024-ratescore},
ROUGE-L, and BLEU-4 \cite{papineni-etal-2002-bleu}.

\begin{table*}[!t]
\centering
\scriptsize
\setlength{\tabcolsep}{3pt}
\renewcommand{\arraystretch}{1.0}
\renewcommand{\rqcell}[2]{\shortstack{#1\\[-1pt]{\scriptsize\textcolor{black!65}{(#2)}}}}
\begin{tabular*}{\textwidth}{@{\extracolsep{\fill}}lcccc@{\hspace{12pt}}cccc@{}}
\toprule
& \multicolumn{4}{c}{\textbf{MIMIC-CXR}} & \multicolumn{4}{c}{\textbf{IU-Xray}} \\
\cmidrule(lr){2-5}\cmidrule(l){6-9}
Effect & Ma-14 & Mi-14 & Ma-5 & Mi-5 & Ma-14 & Mi-14 & Ma-5 & Mi-5 \\
\midrule
\shortstack[l]{$\Delta^{\mathrm{ma,P}}$\\[-1pt]{\scriptsize\phantom{(0/0/0)}}} & \rqcell{+0.0069}{10/0/3} & \rqcell{-0.0322}{0/0/13} & \rqcell{-0.0068}{1/0/12} & \rqcell{-0.0086}{2/0/11} & \rqcell{-0.0010}{9/0/4} & \rqcell{-0.2208}{0/0/13} & \rqcell{-0.0038}{11/0/2} & \rqcell{-0.1023}{2/0/11} \\
\shortstack[l]{$\Delta^{\mathrm{ma,F}}$\\[-1pt]{\scriptsize\phantom{(0/0/0)}}} & \rqcell{+0.0181}{11/0/2} & \rqcell{-0.0269}{0/0/13} & \rqcell{+0.0138}{10/0/3} & \rqcell{+0.0141}{10/0/3} & \rqcell{+0.0439}{11/0/2} & \rqcell{-0.2206}{1/0/12} & \rqcell{+0.0526}{10/0/3} & \rqcell{-0.0411}{5/0/8} \\
\shortstack[l]{$\Delta^{\mathrm{mi,P}}$\\[-1pt]{\scriptsize\phantom{(0/0/0)}}} & \rqcell{+0.0196}{13/0/0} & \rqcell{+0.0002}{6/0/7} & \rqcell{+0.0206}{13/0/0} & \rqcell{+0.0008}{8/0/5} & \rqcell{-0.0085}{0/11/2} & \rqcell{+0.0001}{4/7/2} & \rqcell{-0.0205}{0/11/2} & \rqcell{-0.0009}{2/9/2} \\
\shortstack[l]{$\Delta^{\mathrm{mi,F}}$\\[-1pt]{\scriptsize\phantom{(0/0/0)}}} & \rqcell{+0.0289}{13/0/0} & \rqcell{+0.0075}{9/0/4} & \rqcell{+0.0328}{13/0/0} & \rqcell{+0.0135}{13/0/0} & \rqcell{+0.0254}{11/0/2} & \rqcell{+0.0035}{10/0/3} & \rqcell{+0.0628}{11/0/2} & \rqcell{+0.2085}{12/0/1} \\
\addlinespace[2pt]
\shortstack[l]{$I^{\mathrm{P}}$\\[-1pt]{\scriptsize\phantom{(0/0/0)}}} & \rqcell{-0.0127}{0/0/13} & \rqcell{-0.0323}{0/0/13} & \rqcell{-0.0274}{0/0/13} & \rqcell{-0.0093}{0/0/13} & \rqcell{+0.0076}{10/0/3} & \rqcell{-0.2209}{0/0/13} & \rqcell{+0.0167}{12/0/1} & \rqcell{-0.1015}{2/0/11} \\
\shortstack[l]{$I^{\mathrm{F}}$\\[-1pt]{\scriptsize\phantom{(0/0/0)}}} & \rqcell{-0.0108}{2/0/11} & \rqcell{-0.0345}{0/0/13} & \rqcell{-0.0190}{1/0/12} & \rqcell{+0.0006}{6/0/7} & \rqcell{+0.0185}{9/0/4} & \rqcell{-0.2241}{1/0/12} & \rqcell{-0.0102}{3/0/10} & \rqcell{-0.2496}{0/0/13} \\
\bottomrule
\end{tabular*}
\caption{
RQ2: mean paired differences across 13 configurations,
with positive/zero/negative counts in parentheses.
$\Delta^{o,t}$ denotes Partial (P) or Free (F)
minus Candidate under macro (ma) or micro (mi) scoring;
$I^t=\Delta^{\mathrm{ma},t}-\Delta^{\mathrm{mi},t}$
uses the same metric.
Ma/Mi-14/5 denote macro/micro-F1 over 14/five labels.
Partial replaces five training-defined low-prevalence
labels, distinct from the five-label evaluation subset.
}
\label{tab:rq2}
\end{table*}

\begin{table*}[!t]
\centering

\scriptsize
\setlength{\tabcolsep}{1.5pt}
\renewcommand{\arraystretch}{0.9}
\begin{tabular*}{\textwidth}{@{\extracolsep{\fill}}llrrrrrrrrrrrr@{}}
\toprule
& & \multicolumn{3}{c}{Ma-14}
  & \multicolumn{3}{c}{Mi-14}
  & \multicolumn{3}{c}{Ma-5}
  & \multicolumn{3}{c}{Mi-5} \\
\cmidrule(lr){3-5}\cmidrule(lr){6-8}\cmidrule(lr){9-11}\cmidrule(l){12-14}
Obj. & Contrast
& $\Delta D$ & $A_R$ & $A_G$
& $\Delta D$ & $A_R$ & $A_G$
& $\Delta D$ & $A_R$ & $A_G$
& $\Delta D$ & $A_R$ & $A_G$ \\
\addlinespace[1pt]
\multicolumn{14}{l}{\textbf{MIMIC-CXR}} \\
\addlinespace[1pt]
Macro & P--C
& +.0215 & -.0149 & +.0010
& -.0162 & -.0037 & +.0012
& -.0127 & +.0025 & -.0025
& -.0133 & +.0004 & +.0013 \\

& F--C
& +.0307 & -.0153 & +.0062
& -.0154 & +.0006 & +.0115
& +.0056 & -.0084 & -.0030
& +.0062 & -.0039 & -.0013 \\
\addlinespace[1pt]
Micro & P--C
& +.0220 & -.0079 & -.0052
& -.0010 & -.0023 & +.0007
& +.0160 & -.0018 & -.0048
& -.0024 & +.0002 & +.0012 \\

& F--C
& +.0243 & +.0034 & -.0066
& -.0022 & -.0011 & +.0022
& +.0215 & +.0041 & -.0081
& +.0020 & +.0013 & -.0021 \\
\addlinespace[1pt]
\multicolumn{14}{l}{\textbf{IU-Xray}} \\
\addlinespace[1pt]
Macro & P--C
& -.0106 & -.0193 & -.0466
& -.1975 & -.0897 & -.1177
& +.0055 & +.0212 & -.0265
& -.1414 & +.0164 & +.0263 \\

& F--C
& +.0198 & -.0464 & -.0699
& -.1659 & -.0838 & -.1310
& +.0495 & -.0105 & -.0317
& -.0870 & +.0125 & +.0376 \\
\addlinespace[1pt]
Micro & P--C
& .0000  & .0000  & -.0029
& +.0013 & -.0006 & -.0029
& .0000  & .0000  & .0000
& +.0014 & .0000  & .0000 \\

& F--C
& +.0179 & +.0003 & +.0003
& +.0024 & +.0058 & +.0030
& +.0698 & -.0017 & +.0007
& +.1789 & -.0111 & +.0006 \\
\bottomrule
\end{tabular*}
\caption{
Decision-level contrasts and their changes after
report realization with SFT-Qwen3-VL-8B.
P--C and F--C denote partial-minus-constrained
and free-minus-constrained, respectively.
For each metric, $\Delta D$ is the decision-level
contrast, and $A_r=\Delta M_r-\Delta D$ measures
its change after retrieval ($A_R$) or generation
($A_G$).
The report-level contrast is $\Delta M_r=\Delta D+A_r$.
Ma/Mi denote macro/micro F1.
}
\label{tab:rq3-a}
\end{table*}

\section{Results and Analysis}

\providecommand{\rqcell}[2]{}
\renewcommand{\rqcell}[2]{%
  \shortstack{#1\\[-1pt]{\scriptsize(#2)}}%
}


\subsection{RQ1: Objective Choice under Fixed Evidence}
\label{sec:objective-results}

We compare shared-threshold objectives and candidate
utilities across 13 generator configurations per benchmark.
Tables~\ref{tab:rq1-thresholds} and~\ref{tab:rq1-utilities}
report mean within-generator paired differences and
direction counts; Appendices~\ref{app:rq1-thresholds}
and~\ref{app:rq1-utilities} provide individual results
and available paired confidence intervals.
All four finding metrics evaluate the same predictions
within each condition, without threshold reselection
for the five-label subset.

\paragraph{Shared-threshold objectives.}
With vote counts, rule capacity, and search space fixed,
macro selection improves macro-F1-14 over majority
in all configurations, with mean gains of 0.0644/0.0553
on MIMIC-CXR/IU-Xray.
Micro selection improves mean micro-F1-14 on MIMIC
(+0.0245), but yields mixed effects on IU (-0.0008).
Relative to micro selection, macro selection raises
macro-F1-14 by 0.0172/0.0428 while lowering
micro-F1-14 by 0.0278/0.1415.
The macro-F1 advantage occurs in 12 MIMIC configurations
with one tie and all 13 IU configurations, although
paired intervals exclude zero in only one MIMIC
configuration versus 11 on IU.
These trade-offs depend on the label set:
MIMIC macro-F1-5 effects are small and mixed,
whereas IU macro-F1-5 and micro-F1-5 improve
in 12 configurations with one tie.
Constant controls qualify these gains:
shared macro selection exceeds the always-positive
macro-F1-14 baseline in all MIMIC configurations,
but only 6/13 IU configurations.
\renewcommand{\rqcell}[2]{%
  \shortstack{#1\\[-1pt]{\scriptsize(#2)}}%
}

\paragraph{Candidate utilities.}
With pools, aggregation, and candidate constraints fixed,
RadGraph and ROUGE-L utilities lower all four mean
finding-F1 scores relative to CheXbert on MIMIC,
while improving their corresponding report metrics
(+0.0428 for RadGraph and +0.0369 for ROUGE-L).
On IU, both raise mean 14-label F1, but individual
finding effects are mixed; corresponding report-metric
gains (+0.0715/+0.0316) occur in all configurations.
Direct utility comparisons expose further trade-offs:
on MIMIC, RadGraph utility exceeds ROUGE-L utility
on RadGraph and RaTEScore in every configuration,
but lowers ROUGE-L and BLEU-4 in every configuration.
Mean GREEN gains also conceal decreases for some
generators on both benchmarks.
Report-metric comparisons are point estimates.

Together, these comparisons show that
\textbf{the same sampled evidence yields different
performance trade-offs under different objectives,
with gains that depend on the benchmark, label set,
and generator rather than extending uniformly
across evaluation metrics.}

\begin{table*}[!t]
\centering
\setlength{\abovecaptionskip}{3pt}

\scriptsize
\setlength{\tabcolsep}{2pt}
\renewcommand{\arraystretch}{0.8}
\begin{tabular*}{\textwidth}{@{\extracolsep{\fill}}lrrrrrrrrrrrr@{}}
\toprule
& \multicolumn{6}{c}{MIMIC-CXR}
& \multicolumn{6}{c}{IU-Xray} \\
\cmidrule(lr){2-7}\cmidrule(l){8-13}
Method
& Ma-14 & Mi-14 & Ma-5 & Mi-5 & GREEN & RG
& Ma-14 & Mi-14 & Ma-5 & Mi-5 & GREEN & RG \\
\midrule
Greedy
& .3531 & .5807 & .4712 & .5729 & .4866 & .3146
& .1020 & .5521 & .0467 & .0727 & .5731 & .3122 \\
MBR-1
& .3725 & .5903 & .5111 & .5775 & .4977 & .3382
& .0544 & .5825 & .0000 & .0000 & .5993 & .3960 \\
MBR-2
& .3742 & .5833 & .5145 & .5762 & .4942 & .3347
& .0629 & .5843 & .0000 & .0000 & .5692 & .3585 \\
CCS-Emb
& .3881 & .6092 & .5178 & .6059 & .4952 & .3313
& .0775 & .5842 & .0000 & .0000 & .6022 & .3740 \\
CCS-Dev
& .3789 & .5898 & .5145 & .5824 & .4970 & .3310
& .1022 & .5787 & .0707 & .1565 & .5553 & .2997 \\
SDR-Dev
& .3478 & .5511 & .4740 & .5426 & .4689 & .2914
& .0562 & .5046 & .0157 & .0396 & .6255 & .3543 \\
Constrained verbatim
& .4562 & .6306 & .5799 & .6390 & .4845 & .3115
& .2105 & .5596 & .1529 & .2667 & .5369 & .3027 \\
\addlinespace[1pt]
\multicolumn{13}{l}{\textbf{Ours: macro-oriented targets}} \\
\addlinespace[1pt]
C $\rightarrow$ R
& .4607 & .6353 & .6010 & .6464 & .4375 & .2722
& .1301 & .3491 & .1195 & .1867 & .4867 & .2617 \\
C $\rightarrow$ G
& .4567 & .6370 & .6158 & .6604 & .4890 & .3200
& .1271 & .3783 & .1094 & .1750 & .4963 & .3154 \\
P $\rightarrow$ R
& .4673 & .6154 & .5907 & .6335 & .4274 & .2687
& .1001 & .0619 & .1462 & .0617 & .4367 & .2182 \\
P $\rightarrow$ G
& .4792 & .6220 & .6005 & .6484 & .4817 & .3147
& .0699 & .0630 & .0884 & .0599 & .2920 & .2195 \\
F $\rightarrow$ R
& .4761 & .6205 & .5982 & .6487 & .4243 & .2647
& .1035 & .0994 & .1585 & .1121 & .0774 & .1049 \\
F $\rightarrow$ G
& .4937 & .6332 & .6184 & .6653 & .4813 & .3156
& .0770 & .0813 & .1272 & .1255 & .1981 & .1975 \\
\addlinespace[1pt]
\multicolumn{13}{l}{\textbf{Ours: micro-oriented targets}} \\
\addlinespace[1pt]
C $\rightarrow$ R
& .4239 & .6499 & .5868 & .6583 & .4389 & .2690
& .1207 & .6175 & .0877 & .1774 & .5609 & .2882 \\
C $\rightarrow$ G
& .4248 & .6505 & .5966 & .6718 & .4841 & .3175
& .1212 & .6207 & .0885 & .1789 & .5378 & .3356 \\
P $\rightarrow$ R
& .4381 & .6466 & .6010 & .6561 & .4392 & .2688
& .1207 & .6181 & .0877 & .1789 & .5605 & .2879 \\
P $\rightarrow$ G
& .4416 & .6502 & .6078 & .6706 & .4843 & .3164
& .1183 & .6191 & .0885 & .1803 & .5401 & .3349 \\
F $\rightarrow$ R
& .4517 & .6466 & .6124 & .6616 & .4347 & .2680
& .1389 & .6256 & .1558 & .3452 & .5486 & .2856 \\
F $\rightarrow$ G
& .4424 & .6505 & .6100 & .6717 & .4823 & .3162
& .1394 & .6260 & .1590 & .3584 & .5346 & .3358 \\
\bottomrule
\end{tabular*}
\caption{
Report-level performance with SFT-Qwen3-VL-8B on the
held-out test subset.
C/P/F denote constrained, partial, and free targets;
R/G denote retrieval/generation.
MBR-1/2 use unigram/bigram F1 utilities.
CCS-Emb uses a fixed embedding utility;
CCS-Dev and SDR-Dev select their utility or variant
on development Ma-14.
}
\label{tab:rq3-b}
\end{table*}
\subsection{RQ2: Scoring Objectives and Candidate Constraints}
\label{sec:constraint-results}
\providecommand{\rqcell}[2]{}
\renewcommand{\rqcell}[2]{%
  \shortstack{#1\\[-1pt]{\scriptsize(#2)}}%
}

We compare candidate-constrained (C) and free (F)
finding decisions under macro- and micro-oriented scoring
across 13 generator configurations per benchmark.
All conditions share fitted binary state estimates;
within each objective, scoring coefficients are frozen
across decision conditions.
We additionally evaluate partial replacement (P),
which substitutes free predictions for the five
lowest-prevalence training labels while retaining
candidate predictions elsewhere.
Table~\ref{tab:rq2} reports paired differences and
same-metric interactions; Appendices~\ref{app:rq2-absolute}--\ref{app:rq2-coverage}
provide individual results, paired confidence intervals,
and coverage diagnostics.
For every test input, the scores satisfy
$S(\mathrm{F})\ge S(\mathrm{P})\ge S(\mathrm{C})$,
but this ordering does not guarantee higher held-out F1.

\paragraph{Full and partial relaxation.}
Under macro scoring, Free raises mean Ma-14
by 0.0181/0.0439 on MIMIC-CXR/IU-Xray but lowers
Mi-14 by 0.0269/0.2206.
Mean assertions increase from 2.03 to 2.65
and from 1.13 to 9.82, respectively.
Under micro scoring, Mi-14 instead improves
by 0.0075/0.0035.
Partial does not reliably mitigate these trade-offs:
its macro-oriented Mi-14 loss exceeds Free's
on MIMIC, while its mean Ma-14 contrast is
negative on IU despite gains in 9/13 configurations.

\paragraph{Threshold and candidate-pool diagnostics.}
Threshold diagnostics help explain the IU
macro-oriented assertion expansion.
All thirteen configurations have 1--11 thresholds
below $10^{-8}$, while the precision of positives
added by free inference is only 3.1\%--8.8\%.
Candidate-oracle diagnostics distinguish coverage
limitations from selection headroom.
Zero masked-error candidates exist for
63.8\%/70.5\% of MIMIC/IU inputs on average,
yet macro selection misses an available lower-error
candidate on 51.3\%/47.2\%.
Thus, expanding the pool is not the only possible
source of improvement.
The reference-informed Hamming oracle is diagnostic,
not deployable or an F1 upper bound
(Appendix~\ref{app:rq2-diagnostics}).

\paragraph{Objective-by-constraint interaction.}
Free-minus-constrained Mi-14 interactions average
$-0.0345/-0.2241$ on MIMIC/IU and are negative
in 13/13 and 12/13 configurations.
Uncorrected paired 95\% intervals exclude zero
for 12 and 11 negative interactions, respectively.
Partial Mi-14 interactions are negative throughout
both benchmarks.
Five-label results qualify this pattern:
Free Mi-5 interactions are mixed on MIMIC
but negative throughout IU.
\textbf{Constraint relaxation has objective-dependent
effects: higher scores need not improve held-out F1,
and limited coverage can coexist with selection headroom.}

\subsection{RQ3: Target Construction and Text Realization}
\label{sec:realization-results}


\paragraph{Controlled report realization.}
Table~\ref{tab:rq3-a} compares the same RQ2 targets
before and after retrieval/generation.
On MIMIC-CXR, retrieval attenuates macro-oriented
Ma-14 gains, whereas generation amplifies them.
On IU-Xray, both routes worsen the partial target's
Ma-14 loss and reverse the free target's gain.
Micro-oriented free targets on IU retain positive
contrasts in all four finding metrics.
IU macro-oriented free targets retain
0.3639/0.3469 of requested positives under
retrieval/generation, compared with 0.9177/0.8730
for micro-oriented free targets.
Change decomposition separates correction from
information loss: MIMIC macro free generation
removes 488 false positives and 215 true positives,
yet Mi-14 rises from 0.6311 to 0.6332.
Complete finding scores and change decomposition appear in
Appendices~\ref{app:rq3-absolute}
and~\ref{app:rq3-decomposition}.

\paragraph{Report-level benchmarking.}
Table~\ref{tab:rq3-b} compares all twelve configurations
with greedy, MBR, CCS, and SDR, alongside our
macro-constrained verbatim control.
On MIMIC-CXR, all realization configurations
improve all four finding metrics over greedy.
On IU-Xray, micro free generation leads on
Mi-14, Ma-5, and Mi-5, while constrained verbatim
achieves the highest Ma-14.
Report-quality gains are less consistent:
all realization configurations score below MBR-1
on GREEN and RadGraph on MIMIC, and none exceeds
greedy on GREEN on IU.
\textbf{Realization can correct false positives
while losing true positives; fidelity, finding
accuracy, and report quality need not improve together.}

\section{Conclusion}
Controlled comparisons reveal objective-dependent
trade-offs in sampling-based radiology report inference.
Diagnostics distinguish threshold behavior,
candidate coverage, and selection headroom.
Realization can correct false positives while
discarding true positives, separating fidelity
from reference accuracy.
Structured decisions and final reports therefore
require separate evaluation.

\section*{Limitations}

Our experiments cover two radiology benchmarks and
a fixed sampling budget.
The findings may not generalize to other clinical
domains, institutions, or sampling settings.
Although RQ1 and RQ2 evaluate 13 generator configurations
per benchmark, RQ3 focuses on SFT-Qwen3-VL-8B and
the evaluated retrieval and generation procedures.
Its realization results therefore do not establish
cross-generator generality.

Finding evaluation relies on a fourteen-label
representation and automatic extraction.
This representation omits details such as location,
severity, and temporal change, and extraction errors
can affect both accuracy and fidelity measurements.
Mapping non-positive states to non-assertion also
discards distinctions between absence, uncertainty,
and lack of mention.
Reference-based report metrics provide complementary
signals but do not replace clinical expert assessment.

The fitted state estimates and F-measure-informed
scores do not guarantee calibrated probabilities,
held-out optimality, or clinically consistent label
combinations.
Paired bootstrap intervals condition on fitted
development parameters and do not capture
development-fitting uncertainty; multiple comparisons
are uncorrected.
We do not evaluate clinical deployment, clinician
decision-making, or patient outcomes.

\section*{Ethics Statement}

This work studies inference and evaluation for
radiology report generation using MIMIC-CXR and
IU-Xray.
Access to and use of these datasets must comply
with their applicable agreements and restrictions.
Any release of code or derived artifacts should
respect those restrictions and avoid redistributing
protected clinical records.

Generated reports may omit findings, introduce
unsupported assertions, or express inconsistent
finding combinations.
Higher finding F1 or report-metric scores do not
establish clinical safety.
The evaluated systems are research tools and should
not be used for patient care without clinical
validation and qualified professional oversight.

Medical datasets may underrepresent patient groups
and clinical settings.
Our benchmark-level results do not establish
equitable performance across demographic subgroups.
Any deployment would require assessment of subgroup
performance, privacy protections, and potential
effects on clinical workflows.




\bibliography{custom}

\appendix

\setlength{\floatsep}{6pt plus 2pt minus 2pt}
\setlength{\dblfloatsep}{6pt plus 2pt minus 2pt}
\setlength{\textfloatsep}{10pt plus 2pt minus 2pt}
\setlength{\dbltextfloatsep}{10pt plus 2pt minus 2pt}
\setlength{\abovedisplayskip}{6pt plus 2pt minus 2pt}
\setlength{\belowdisplayskip}{6pt plus 2pt minus 2pt}
\renewcommand{\topfraction}{0.95}
\renewcommand{\dbltopfraction}{0.95}
\renewcommand{\textfraction}{0.05}
\renewcommand{\floatpagefraction}{0.8}
\renewcommand{\dblfloatpagefraction}{0.8}
\setcounter{topnumber}{8}
\setcounter{totalnumber}{10}
\setcounter{dbltopnumber}{8}
\makeatletter
\setlength{\@fptop}{0pt}
\setlength{\@fpsep}{4pt}
\setlength{\@fpbot}{0pt plus 1fil}
\setlength{\@dblfptop}{0pt}
\setlength{\@dblfpsep}{4pt}
\setlength{\@dblfpbot}{0pt plus 1fil}
\makeatother

\section{Derivation and Development Fitting}
\label{app:derivation}

\paragraph{Table notation.}
Generator abbreviations used throughout the appendix tables are Q25-3/Q25-1 (Qwen2.5-VL-7B, 3/1 training epochs), Q35 (Qwen3.5-122B), IV35 (InternVL3.5-241B), DPO-G (DPO-RRG-GREEN), R2CMN (R2GenCMN), L15 (LLaVA-1.5), and LN-M/LN-V (LLaVA-NeXT-M/V). Other generator names are unchanged. In compact headers, Obj., Cond., T/R, and Wdr mean objective, condition, target/route, and withdrawn. Out denotes out-of-pool rate; $H$ and $H_o$ denote nearest Hamming distance overall and conditional on out-of-pool targets. $\Delta S_{PC}=S(P)-S(C)$ and $\Delta S_{FP}=S(F)-S(P)$. Gen. and T denote generator and target; leading zeros are omitted in compact numeric cells. Train prev. means training prevalence; P, R, and Asrt. denote precision, recall, and assertions. In development records, the family column supplies the SDR/CCS prefix; Cham., Haus., and Hung. denote chamfer, hausdorff, and hungarian; Dev and Sel. denote development Ma-14 and selected configuration.

The derivation uses established F-measure optimization principles to motivate our
controlled scoring interventions. It is not a new optimality result, and does not
establish calibrated state estimates or optimal held-out F1.

\paragraph{Shared table notes.}
For RQ1 paired contrasts, cells show the observed difference and the exported, configuration-specific, uncorrected 95\% paired cluster-bootstrap interval; no cross-generator interval is implied. RQ2 contrast intervals are supplied by the RQ2 export. In RQ2 absolute target tables, precision/recall are fourteen-label macro averages; assertions use the original exported per-input statistic, distinct from the unmasked positive rates in threshold diagnostics. For candidate-coverage tables, $H_o$ is conditional on the free target being outside the pool.

\subsection{Local F1 Changes}
\label{app:local-f1}
For one binary finding, let $D=2TP+FP+FN>0$. Then
\begin{equation}
F_1=\frac{2TP}{D},\qquad
\frac{\partial F_1}{\partial TP}=\frac{2(FP+FN)}{D^2}=a,
\end{equation}
\begin{equation}
\frac{\partial F_1}{\partial FP}
=\frac{\partial F_1}{\partial FN}=-\frac{2TP}{D^2}=-b.
\end{equation}
The first-order change is
\begin{equation}
\Delta F_1\approx a\Delta TP-b(\Delta FP+\Delta FN).
\end{equation}
Changing a negative prediction to positive on a reference-positive cell increases
$TP$ and decreases $FN$, giving local gain $a+b$. On a reference-negative cell,
it increases $FP$, giving local cost $b$. These are derivatives at the development
operating point, not exact finite changes in F1.

\subsection{Soft Counts and Macro/Micro Coefficients}
\label{app:soft-counts}
The implementation evaluates the derivative formulas at probability-weighted
counts, not confusion counts computed directly against development references.
For current constrained development selections $z_i$, define
\begin{equation}
\begin{aligned}
T_L&=\sum_{i\in\mathrm{dev}}z_{iL}p_{iL}^{(1)},\\
P_L&=\sum_{i\in\mathrm{dev}}z_{iL}p_{iL}^{(0)},\\
Q_L&=\sum_{i\in\mathrm{dev}}(1-z_{iL})p_{iL}^{(1)}.
\end{aligned}
\end{equation}
These are soft TP, FP, and FN under the fitted probabilities. Reference labels
are used to fit state models; coefficient updates instead use these fitted
probabilities on development inputs. No reference-uncertainty mask is applied
to the sums above.
For $D_L=2T_L+P_L+Q_L$, macro derivatives give
\begin{equation}
a_L^{\mathrm{ma}}=\frac{2(P_L+Q_L)}{D_L^2},\qquad
b_L^{\mathrm{ma}}=\frac{2T_L}{D_L^2}.
\end{equation}
The common $1/K$ factor is omitted. For pooled counts
$T=\sum_LT_L$, $P=\sum_LP_L$, $Q=\sum_LQ_L$, and $D=2T+P+Q$,
\begin{equation}
a_L^{\mathrm{mi}}=\frac{2(P+Q)}{D^2},\qquad
b_L^{\mathrm{mi}}=\frac{2T}{D^2}.
\end{equation}
Macro coefficients vary by label; micro coefficients are shared.
Substituting expected counts into an F1 ratio is not the exact expectation
of that ratio, and does not guarantee held-out optimality.
\subsection{Action Scores and Binary Threshold}
\label{app:action-scores}
With fitted state estimates, define
\begin{equation}
u_L(1)=a_Lp_L^{(1)}-b_Lp_L^{(0)},\qquad
u_L(0)=-b_Lp_L^{(1)}.
\end{equation}
Summing the selected actions gives the score in the main text. It can be written as
\begin{equation}
S(z)=\sum_Lu_L(0)+\sum_Lz_L[u_L(1)-u_L(0)].
\end{equation}
The first sum is constant in $z$, and
\begin{equation}
u_L(1)-u_L(0)=(a_L+b_L)p_L^{(1)}-b_Lp_L^{(0)}.
\end{equation}
Hence unconstrained maximization chooses each label independently.
Our experiments use $p_L^{(0)}=1-p_L^{(1)}$ by construction. For $a_L+2b_L>0$,
\begin{equation}
\hat z_L^{\mathrm{free}}=\mathbf1\left[p_L^{(1)}\ge\frac{b_L}{a_L+2b_L}\right].
\end{equation}
For the unmodified nondegenerate derivative coefficients,
\begin{equation}
\frac{b_L}{a_L+2b_L}=\frac{T_L}{2T_L+P_L+Q_L}
=\frac{F_{1,L}^{\mathrm{soft}}}{2}.
\end{equation}
This is soft-count F1, not empirical development F1 against references.
For micro coefficients, the identity uses pooled soft counts instead.
This algebraic relation does not guarantee held-out optimality. Numerical coefficient floors can change the soft-F1 identity.
The action comparison and $b_L/(a_L+2b_L)$ threshold remain equivalent
for the final floored coefficients.

\subsection{Normalization and Development Fitting}
For $\bar b=K^{-1}\sum_Lb_L>0$, jointly replacing $a,b$ by $a/\bar b,b/\bar b$
scales $S$ by a common positive factor and preserves its maximizers.
Algorithm~\ref{alg:constraints} uses twelve fixed iterations, without a convergence
criterion. State models remain fixed. Common positive rescaling preserves score
maximizers; the subsequent elementwise coefficient floors need not.
The frozen coefficients are those that produced the last development selection,
not the subsequent update. There is no monotonic-F1 or global-optimality guarantee.

\begin{algorithm}[t]
\footnotesize
\caption{Development Score Fitting and Test Decisions}
\label{alg:constraints}
\begin{algorithmic}[1]
\REQUIRE Development data, test inputs, ordered candidate pools,
training-defined replacement labels $R_5$
\ENSURE Constrained, partial, and free test predictions
\STATE Fit and freeze per-label state models on reference-scored development rows
\STATE Compute $p_i^{(1)}$, $p_i^{(0)}=1-p_i^{(1)}$, and ordered candidate vectors $\mathcal Z_i$
\FOR{$o\in\{\mathrm{ma},\mathrm{mi}\}$}
\STATE Initialize $a_L\leftarrow2$, $b_L\leftarrow1$ for all $L$
\FOR{$t=1,\ldots,12$}
\STATE $(a_{\mathrm{used}},b_{\mathrm{used}})\leftarrow(a,b)$
\STATE Select $z_i\leftarrow\arg\max_{z\in\mathcal Z_i}S(z;p_i,a,b)$ for each development input
\STATE Compute probability-weighted development counts $T_L,P_L,Q_L$
\STATE Update $(a,b)$ for objective $o$, flooring F1 denominators at $10^{-9}$
\STATE $c\leftarrow\max(\operatorname{mean}_Lb_L,10^{-12})$
\STATE $(a,b)\leftarrow(a/c,b/c)$
\STATE $a_L\leftarrow\max(a_L,10^{-9})$, $b_L\leftarrow\max(b_L,10^{-9})$
\ENDFOR
\STATE Freeze $(a^o,b^o)\leftarrow(a_{\mathrm{used}},b_{\mathrm{used}})$
\FOR{each test input $i$}
\STATE $\hat z_i^{o,\mathrm{con}}\leftarrow\arg\max_{z\in\mathcal Z_i}S(z;p_i,a^o,b^o)$
\STATE $\hat z_{iL}^{o,\mathrm{free}}\leftarrow\mathbf1[(a_L^o+b_L^o)p_{iL}^{(1)}\ge b_L^op_{iL}^{(0)}]$
\STATE Set $\hat z_{iL}^{o,\mathrm{partial}}=\hat z_{iL}^{o,\mathrm{free}}$ on $R_5$, and $\hat z_{iL}^{o,\mathrm{con}}$ elsewhere
\ENDFOR
\ENDFOR
\STATE \textbf{return} all objective-specific C/P/F test predictions
\end{algorithmic}
\end{algorithm}
Candidate ties use the earliest pool index; partial does not reselect a candidate.
\subsection{Constrained, Partial, and Free Score Ordering}
Let $u_L(z_L)$ denote the score contribution for the chosen label action.
Because $\hat z_L^{\mathrm{free}}$ maximizes this contribution,
\begin{equation}
u_L(\hat z_L^{\mathrm{free}})-u_L(\hat z_L^{\mathrm{con}})\ge0.
\end{equation}
Partial replacement applies the free action only on the training-defined $R_5$:
\begin{equation}
\begin{aligned}
&S(\hat z^{\mathrm{partial}})-S(\hat z^{\mathrm{con}})\\
&\quad=\sum_{L\in R_5}[u_L(\hat z_L^{\mathrm{free}})-u_L(\hat z_L^{\mathrm{con}})]\ge0.
\end{aligned}
\end{equation}
Likewise,
\begin{equation}
\begin{aligned}
&S(\hat z^{\mathrm{free}})-S(\hat z^{\mathrm{partial}})\\
&\quad=\sum_{L\notin R_5}[u_L(\hat z_L^{\mathrm{free}})-u_L(\hat z_L^{\mathrm{con}})]\ge0.
\end{aligned}
\end{equation}
Thus $S(F)\ge S(P)\ge S(C)$. This concerns the frozen additive score, not the
ordering of held-out F1, target fidelity, or report quality. Partial is a hybrid
intervention, not an independently optimized intermediate feasible set.

\section{Additional Results for Objective Choice}
\label{app:rq1}

\subsection{Scope and Evaluation Protocol}
We report results for 13 generator configurations on each benchmark,
using 1,432 MIMIC-CXR and 446 IU-Xray test inputs.
All comparisons are paired within a generator.
Ma-14/Mi-14 and Ma-5/Mi-5 denote unweighted macro/micro-F1
over 14 findings and the five-finding subset
(Atelectasis, Cardiomegaly, Consolidation, Edema, and Pleural Effusion).
All four metrics evaluate the same predictions without threshold
reselection for the subset. Reference-uncertain cells are masked;
predicted uncertainty is treated as non-assertion.
For the constant controls, we use the fixed reference-only mask
rather than the alternative prediction-dependent masking export.
Direction counts describe positive, zero, and negative differences,
not statistical significance.

\subsection{Shared-Threshold Objectives}
\label{app:rq1-thresholds}
The corresponding results are reported in Tables~\ref{tab:app-rq1-threshold-mimic}, \ref{tab:app-rq1-threshold-iu}.

With $N=32$, binary majority uses $\tau=17$.
Macro- and micro-selected thresholds maximize their respective
14-label development F1 within $\{1,\ldots,32\}$ and are frozen for test.
The rules share evidence, search space, and parameter count.
The absolute scores below retain all three rules, allowing both
objective-selected rules to be compared with majority.

\begin{table}[!htbp]\centering\scriptsize
\setlength{\tabcolsep}{1.2pt}\renewcommand{\arraystretch}{0.95}
\begin{tabular}{ccccccccc}\toprule
Generator & Rule & $\tau$ & Ma-14 & Mi-14 & Ma-5 & Mi-5 & P & R \\ \midrule
Qwen2.5-VL-8B & Maj & 17 & .3637 & .6136 & .4877 & .5946 & .5282 & .3335 \\
 & Ma & 3 & .4212 & .5940 & .5529 & .6073 & .3311 & .5893 \\
 & Mi & 10 & .4036 & .6436 & .5527 & .6474 & .4308 & .4262 \\
Q25-1 & Maj & 17 & .2885 & .5441 & .4163 & .5251 & .3941 & .2610 \\
 & Ma & 8 & .3651 & .6023 & .5148 & .5998 & .3943 & .4002 \\
 & Mi & 9 & .3614 & .6018 & .5077 & .5988 & .4208 & .3849 \\
Q35 & Maj & 17 & .3547 & .5184 & .4775 & .4986 & .4087 & .4196 \\
 & Ma & 2 & .3940 & .5043 & .5156 & .5424 & .3364 & .6089 \\
 & Mi & 24 & .3329 & .5119 & .4334 & .4555 & .4504 & .3498 \\
IV35 & Maj & 17 & .3770 & .5847 & .5117 & .5330 & .4005 & .4015 \\
 & Ma & 2 & .4115 & .5392 & .5356 & .5487 & .3600 & .6168 \\
 & Mi & 13 & .3858 & .5814 & .5290 & .5452 & .3894 & .4434 \\
ESC-RL & Maj & 17 & .3682 & .6123 & .4971 & .6013 & .5726 & .3293 \\
 & Ma & 4 & .4442 & .6139 & .5621 & .6191 & .3728 & .5668 \\
 & Mi & 8 & .4322 & .6501 & .5741 & .6568 & .4356 & .4721 \\
MPO & Maj & 17 & .3601 & .6077 & .5102 & .6023 & .5492 & .3279 \\
 & Ma & 3 & .4148 & .5748 & .5229 & .5744 & .3250 & .6035 \\
 & Mi & 12 & .3998 & .6380 & .5602 & .6418 & .4706 & .4006 \\
EditGRPO & Maj & 17 & .3748 & .6219 & .5078 & .6168 & .5398 & .3408 \\
 & Ma & 3 & .4302 & .5848 & .5458 & .5990 & .3341 & .6214 \\
 & Mi & 13 & .4078 & .6442 & .5484 & .6411 & .4854 & .4006 \\
\shortstack[l]{DPO-RRG\\-GREEN} & Maj & 17 & .3452 & .6031 & .4925 & .6049 & .5070 & .3257 \\
 & Ma & 3 & .4312 & .5910 & .5530 & .6205 & .3788 & .5341 \\
 & Mi & 9 & .4116 & .6332 & .5686 & .6436 & .4989 & .4099 \\
R2Gen & Maj & 17 & .3698 & .5941 & .5018 & .5934 & .5503 & .3392 \\
 & Ma & 5 & .4150 & .6151 & .5637 & .6276 & .3982 & .5003 \\
 & Mi & 8 & .4070 & .6239 & .5615 & .6387 & .4070 & .4424 \\
R2CMN & Maj & 17 & .3407 & .5762 & .4506 & .5617 & .4550 & .3112 \\
 & Ma & 3 & .4236 & .5874 & .5500 & .6047 & .3515 & .5538 \\
 & Mi & 8 & .4047 & .6150 & .5371 & .6282 & .5047 & .4310 \\
L15 & Maj & 17 & .4029 & .6391 & .5262 & .6300 & .5625 & .3672 \\
 & Ma & 5 & .4876 & .6429 & .5995 & .6550 & .4323 & .5757 \\
 & Mi & 8 & .4719 & .6614 & .6138 & .6816 & .4887 & .4996 \\
\shortstack[l]{LLaVA-NeXT\\-M} & Maj & 17 & .4274 & .6414 & .5432 & .6246 & .6422 & .3802 \\
 & Ma & 6 & .4934 & .6482 & .6024 & .6584 & .4402 & .5742 \\
 & Mi & 6 & .4934 & .6482 & .6024 & .6584 & .4402 & .5742 \\
\shortstack[l]{LLaVA-NeXT\\-V} & Maj & 17 & .4047 & .6301 & .5164 & .6151 & .5807 & .3632 \\
 & Ma & 5 & .4826 & .6457 & .5883 & .6543 & .4250 & .5767 \\
 & Mi & 6 & .4785 & .6522 & .5814 & .6529 & .4443 & .5413 \\
\bottomrule\end{tabular}
\caption{MIMIC: absolute shared-threshold test results. Maj/Ma/Mi denote majority/macro-selected/micro-selected rules; P/R are 14-label macro precision/recall.}\label{tab:app-rq1-threshold-mimic}
\end{table}

\begin{table}[!htbp]\centering\scriptsize
\setlength{\tabcolsep}{1.2pt}\renewcommand{\arraystretch}{0.95}
\begin{tabular}{ccccccccc}\toprule
Generator & Rule & $\tau$ & Ma-14 & Mi-14 & Ma-5 & Mi-5 & P & R \\ \midrule
Q35 & Maj & 17 & .2106 & .6235 & .1822 & .1833 & .4147 & .2491 \\
 & Ma & 1 & .2479 & .5491 & .3080 & .3841 & .2578 & .4113 \\
 & Mi & 9 & .2102 & .6054 & .1811 & .2222 & .3142 & .2853 \\
IV35 & Maj & 17 & .2072 & .6322 & .2339 & .3421 & .2719 & .2668 \\
 & Ma & 2 & .2332 & .5522 & .2879 & .3692 & .1839 & .4372 \\
 & Mi & 18 & .2025 & .6319 & .2151 & .3087 & .3106 & .2611 \\
Qwen2.5-VL-8B(3 epoch) & Maj & 17 & .0556 & .5948 & .0000 & .0000 & .0458 & .0707 \\
 & Ma & 1 & .1713 & .4163 & .1452 & .2393 & .1214 & .3556 \\
 & Mi & 7 & .0899 & .5975 & .0077 & .0198 & .2003 & .0918 \\
Qwen2.5-VL-8B(1 epoch) & Maj & 17 & .0550 & .5917 & .0000 & .0000 & .0447 & .0714 \\
 & Ma & 1 & .0909 & .3872 & .0000 & .0000 & .0776 & .1647 \\
 & Mi & 4 & .0550 & .5911 & .0000 & .0000 & .0447 & .0714 \\
ESC-RL & Maj & 17 & .0556 & .5946 & .0000 & .0000 & .0460 & .0704 \\
 & Ma & 1 & .1735 & .4052 & .1535 & .2654 & .1225 & .3521 \\
 & Mi & 4 & .0965 & .5805 & .0381 & .1026 & .1040 & .1060 \\
MPO & Maj & 17 & .0552 & .5936 & .0000 & .0000 & .0450 & .0714 \\
 & Ma & 1 & .1612 & .4156 & .1744 & .1957 & .1221 & .3024 \\
 & Mi & 6 & .0837 & .5962 & .0082 & .0204 & .2381 & .0880 \\
R2Gen & Maj & 17 & .0550 & .5917 & .0000 & .0000 & .0447 & .0714 \\
 & Ma & 2 & .0633 & .5539 & .0233 & .0789 & .0525 & .0804 \\
 & Mi & 4 & .0550 & .5911 & .0000 & .0000 & .0447 & .0714 \\
EditGRPO & Maj & 17 & .0526 & .5557 & .0000 & .0000 & .0450 & .0632 \\
 & Ma & 1 & .0929 & .5697 & .0548 & .0820 & .1203 & .1024 \\
 & Mi & 2 & .0788 & .5893 & .0222 & .0200 & .1365 & .0886 \\
\shortstack[l]{DPO-RRG\\-GREEN} & Maj & 17 & .0550 & .5911 & .0000 & .0000 & .0447 & .0714 \\
 & Ma & 1 & .1283 & .5334 & .1062 & .1912 & .1659 & .1490 \\
 & Mi & 2 & .0928 & .5818 & .0154 & .0396 & .1487 & .1077 \\
R2CMN & Maj & 17 & .0550 & .5917 & .0000 & .0000 & .0447 & .0714 \\
 & Ma & 1 & .1209 & .3892 & .0961 & .1584 & .0916 & .2362 \\
 & Mi & 22 & .0552 & .5936 & .0000 & .0000 & .0450 & .0714 \\
L15 & Maj & 17 & .0550 & .5917 & .0000 & .0000 & .0447 & .0714 \\
 & Ma & 1 & .0833 & .3907 & .0175 & .0541 & .0610 & .2090 \\
 & Mi & 29 & .0545 & .5846 & .0000 & .0000 & .0446 & .0699 \\
\shortstack[l]{LLaVA-NeXT\\-M} & Maj & 17 & .0550 & .5917 & .0000 & .0000 & .0447 & .0714 \\
 & Ma & 1 & .0779 & .3864 & .0074 & .0172 & .0678 & .2158 \\
 & Mi & 3 & .0550 & .5917 & .0000 & .0000 & .0447 & .0714 \\
\shortstack[l]{LLaVA-NeXT\\-V} & Maj & 17 & .0550 & .5917 & .0000 & .0000 & .0447 & .0714 \\
 & Ma & 1 & .0954 & .3370 & .0393 & .0945 & .0921 & .2756 \\
 & Mi & 4 & .0550 & .5905 & .0000 & .0000 & .0447 & .0714 \\
\bottomrule\end{tabular}
\caption{IU: absolute shared-threshold test results. Maj/Ma/Mi denote majority/macro-selected/micro-selected rules; P/R are 14-label macro precision/recall.}\label{tab:app-rq1-threshold-iu}
\end{table}

\paragraph{Paired differences.}
The corresponding results are reported in Tables~\ref{tab:app-rq1-ci-mimic-macro_oriented_minus_micro_oriented}, \ref{tab:app-rq1-ci-mimic-macro_oriented_minus_binary_majority}, \ref{tab:app-rq1-ci-iu-macro_oriented_minus_micro_oriented}, \ref{tab:app-rq1-ci-iu-macro_oriented_minus_binary_majority}.

We report the paired finding intervals available in the threshold export
for Macro minus Micro and Macro minus Majority.
Micro minus Majority point differences follow from the absolute scores;
its paired intervals are not present in the result export and are
therefore not reconstructed by subtracting other intervals.

\begin{table*}[!htbp]\centering\scriptsize
\setlength{\tabcolsep}{2pt}\renewcommand{\arraystretch}{0.95}
\begin{tabular}{@{}ccccc@{}}\toprule
Generator & Ma-14 & Mi-14 & Ma-5 & Mi-5 \\ \midrule
Q25-3 & +0.0176\,{\fontsize{7.5}{9}\selectfont [-0.0024, +0.0364]} & -0.0495\,{\fontsize{7.5}{9}\selectfont [-0.0611, -0.0380]} & +0.0002\,{\fontsize{7.5}{9}\selectfont [-0.0352, +0.0301]} & -0.0401\,{\fontsize{7.5}{9}\selectfont [-0.0582, -0.0208]} \\
Q25-1 & +0.0037\,{\fontsize{7.5}{9}\selectfont [-0.0022, +0.0113]} & +0.0004\,{\fontsize{7.5}{9}\selectfont [-0.0037, +0.0048]} & +0.0072\,{\fontsize{7.5}{9}\selectfont [-0.0053, +0.0259]} & +0.0010\,{\fontsize{7.5}{9}\selectfont [-0.0059, +0.0082]} \\
Q35 & +0.0611\,{\fontsize{7.5}{9}\selectfont [+0.0360, +0.0892]} & -0.0076\,{\fontsize{7.5}{9}\selectfont [-0.0253, +0.0106]} & +0.0822\,{\fontsize{7.5}{9}\selectfont [+0.0409, +0.1269]} & +0.0870\,{\fontsize{7.5}{9}\selectfont [+0.0459, +0.1331]} \\
IV35 & +0.0257\,{\fontsize{7.5}{9}\selectfont [-0.0008, +0.0509]} & -0.0422\,{\fontsize{7.5}{9}\selectfont [-0.0552, -0.0293]} & +0.0066\,{\fontsize{7.5}{9}\selectfont [-0.0167, +0.0313]} & +0.0035\,{\fontsize{7.5}{9}\selectfont [-0.0201, +0.0293]} \\
ESC-RL & +0.0120\,{\fontsize{7.5}{9}\selectfont [-0.0072, +0.0311]} & -0.0362\,{\fontsize{7.5}{9}\selectfont [-0.0458, -0.0264]} & -0.0121\,{\fontsize{7.5}{9}\selectfont [-0.0411, +0.0151]} & -0.0377\,{\fontsize{7.5}{9}\selectfont [-0.0518, -0.0242]} \\
MPO & +0.0150\,{\fontsize{7.5}{9}\selectfont [-0.0126, +0.0403]} & -0.0632\,{\fontsize{7.5}{9}\selectfont [-0.0760, -0.0507]} & -0.0373\,{\fontsize{7.5}{9}\selectfont [-0.0806, +0.0004]} & -0.0674\,{\fontsize{7.5}{9}\selectfont [-0.0870, -0.0481]} \\
EditGRPO & +0.0224\,{\fontsize{7.5}{9}\selectfont [-0.0025, +0.0461]} & -0.0594\,{\fontsize{7.5}{9}\selectfont [-0.0741, -0.0455]} & -0.0025\,{\fontsize{7.5}{9}\selectfont [-0.0399, +0.0306]} & -0.0422\,{\fontsize{7.5}{9}\selectfont [-0.0640, -0.0192]} \\
DPO-G & +0.0196\,{\fontsize{7.5}{9}\selectfont [-0.0014, +0.0404]} & -0.0422\,{\fontsize{7.5}{9}\selectfont [-0.0542, -0.0305]} & -0.0157\,{\fontsize{7.5}{9}\selectfont [-0.0435, +0.0114]} & -0.0232\,{\fontsize{7.5}{9}\selectfont [-0.0389, -0.0070]} \\
R2Gen & +0.0079\,{\fontsize{7.5}{9}\selectfont [-0.0056, +0.0232]} & -0.0087\,{\fontsize{7.5}{9}\selectfont [-0.0166, -0.0011]} & +0.0022\,{\fontsize{7.5}{9}\selectfont [-0.0201, +0.0257]} & -0.0111\,{\fontsize{7.5}{9}\selectfont [-0.0230, +0.0012]} \\
R2CMN & +0.0188\,{\fontsize{7.5}{9}\selectfont [-0.0012, +0.0397]} & -0.0276\,{\fontsize{7.5}{9}\selectfont [-0.0380, -0.0170]} & +0.0129\,{\fontsize{7.5}{9}\selectfont [-0.0202, +0.0420]} & -0.0235\,{\fontsize{7.5}{9}\selectfont [-0.0393, -0.0069]} \\
L15 & +0.0158\,{\fontsize{7.5}{9}\selectfont [-0.0041, +0.0364]} & -0.0186\,{\fontsize{7.5}{9}\selectfont [-0.0264, -0.0108]} & -0.0144\,{\fontsize{7.5}{9}\selectfont [-0.0406, +0.0099]} & -0.0266\,{\fontsize{7.5}{9}\selectfont [-0.0393, -0.0150]} \\
LN-M & +0.0000\,{\fontsize{7.5}{9}\selectfont [+0.0000, +0.0000]} & +0.0000\,{\fontsize{7.5}{9}\selectfont [+0.0000, +0.0000]} & +0.0000\,{\fontsize{7.5}{9}\selectfont [+0.0000, +0.0000]} & +0.0000\,{\fontsize{7.5}{9}\selectfont [+0.0000, +0.0000]} \\
LN-V & +0.0041\,{\fontsize{7.5}{9}\selectfont [-0.0070, +0.0167]} & -0.0065\,{\fontsize{7.5}{9}\selectfont [-0.0113, -0.0017]} & +0.0069\,{\fontsize{7.5}{9}\selectfont [-0.0089, +0.0218]} & +0.0013\,{\fontsize{7.5}{9}\selectfont [-0.0066, +0.0094]} \\
\bottomrule\end{tabular}
\caption{MIMIC: Macro minus Micro.}\label{tab:app-rq1-ci-mimic-macro_oriented_minus_micro_oriented}
\end{table*}

\begin{table*}[!htbp]\centering\scriptsize
\setlength{\tabcolsep}{2pt}\renewcommand{\arraystretch}{0.95}
\begin{tabular}{@{}ccccc@{}}\toprule
Generator & Ma-14 & Mi-14 & Ma-5 & Mi-5 \\ \midrule
Q25-3 & +0.0575\,{\fontsize{7.5}{9}\selectfont [+0.0333, +0.0803]} & -0.0196\,{\fontsize{7.5}{9}\selectfont [-0.0364, -0.0036]} & +0.0652\,{\fontsize{7.5}{9}\selectfont [+0.0238, +0.1057]} & +0.0128\,{\fontsize{7.5}{9}\selectfont [-0.0163, +0.0449]} \\
Q25-1 & +0.0766\,{\fontsize{7.5}{9}\selectfont [+0.0558, +0.0986]} & +0.0581\,{\fontsize{7.5}{9}\selectfont [+0.0431, +0.0734]} & +0.0986\,{\fontsize{7.5}{9}\selectfont [+0.0655, +0.1395]} & +0.0748\,{\fontsize{7.5}{9}\selectfont [+0.0465, +0.1092]} \\
Q35 & +0.0393\,{\fontsize{7.5}{9}\selectfont [+0.0183, +0.0610]} & -0.0141\,{\fontsize{7.5}{9}\selectfont [-0.0284, +0.0015]} & +0.0381\,{\fontsize{7.5}{9}\selectfont [+0.0141, +0.0655]} & +0.0438\,{\fontsize{7.5}{9}\selectfont [+0.0135, +0.0795]} \\
IV35 & +0.0345\,{\fontsize{7.5}{9}\selectfont [+0.0070, +0.0624]} & -0.0454\,{\fontsize{7.5}{9}\selectfont [-0.0602, -0.0313]} & +0.0239\,{\fontsize{7.5}{9}\selectfont [-0.0024, +0.0527]} & +0.0157\,{\fontsize{7.5}{9}\selectfont [-0.0105, +0.0449]} \\
ESC-RL & +0.0761\,{\fontsize{7.5}{9}\selectfont [+0.0499, +0.1020]} & +0.0016\,{\fontsize{7.5}{9}\selectfont [-0.0145, +0.0183]} & +0.0650\,{\fontsize{7.5}{9}\selectfont [+0.0268, +0.0997]} & +0.0178\,{\fontsize{7.5}{9}\selectfont [-0.0092, +0.0494]} \\
MPO & +0.0547\,{\fontsize{7.5}{9}\selectfont [+0.0303, +0.0781]} & -0.0329\,{\fontsize{7.5}{9}\selectfont [-0.0482, -0.0170]} & +0.0127\,{\fontsize{7.5}{9}\selectfont [-0.0274, +0.0505]} & -0.0279\,{\fontsize{7.5}{9}\selectfont [-0.0528, +0.0014]} \\
EditGRPO & +0.0554\,{\fontsize{7.5}{9}\selectfont [+0.0321, +0.0785]} & -0.0371\,{\fontsize{7.5}{9}\selectfont [-0.0536, -0.0210]} & +0.0381\,{\fontsize{7.5}{9}\selectfont [+0.0036, +0.0704]} & -0.0179\,{\fontsize{7.5}{9}\selectfont [-0.0442, +0.0125]} \\
DPO-G & +0.0860\,{\fontsize{7.5}{9}\selectfont [+0.0612, +0.1085]} & -0.0121\,{\fontsize{7.5}{9}\selectfont [-0.0283, +0.0032]} & +0.0604\,{\fontsize{7.5}{9}\selectfont [+0.0305, +0.0883]} & +0.0156\,{\fontsize{7.5}{9}\selectfont [-0.0073, +0.0407]} \\
R2Gen & +0.0451\,{\fontsize{7.5}{9}\selectfont [+0.0248, +0.0660]} & +0.0211\,{\fontsize{7.5}{9}\selectfont [+0.0055, +0.0356]} & +0.0619\,{\fontsize{7.5}{9}\selectfont [+0.0290, +0.0950]} & +0.0342\,{\fontsize{7.5}{9}\selectfont [+0.0109, +0.0611]} \\
R2CMN & +0.0829\,{\fontsize{7.5}{9}\selectfont [+0.0610, +0.1060]} & +0.0112\,{\fontsize{7.5}{9}\selectfont [-0.0076, +0.0290]} & +0.0994\,{\fontsize{7.5}{9}\selectfont [+0.0672, +0.1332]} & +0.0430\,{\fontsize{7.5}{9}\selectfont [+0.0118, +0.0784]} \\
L15 & +0.0847\,{\fontsize{7.5}{9}\selectfont [+0.0556, +0.1117]} & +0.0038\,{\fontsize{7.5}{9}\selectfont [-0.0103, +0.0173]} & +0.0733\,{\fontsize{7.5}{9}\selectfont [+0.0363, +0.1074]} & +0.0250\,{\fontsize{7.5}{9}\selectfont [-0.0011, +0.0532]} \\
LN-M & +0.0660\,{\fontsize{7.5}{9}\selectfont [+0.0364, +0.0928]} & +0.0068\,{\fontsize{7.5}{9}\selectfont [-0.0068, +0.0209]} & +0.0592\,{\fontsize{7.5}{9}\selectfont [+0.0236, +0.0937]} & +0.0338\,{\fontsize{7.5}{9}\selectfont [+0.0099, +0.0623]} \\
LN-V & +0.0779\,{\fontsize{7.5}{9}\selectfont [+0.0532, +0.1013]} & +0.0155\,{\fontsize{7.5}{9}\selectfont [+0.0012, +0.0295]} & +0.0719\,{\fontsize{7.5}{9}\selectfont [+0.0368, +0.1090]} & +0.0392\,{\fontsize{7.5}{9}\selectfont [+0.0136, +0.0669]} \\
\bottomrule\end{tabular}
\caption{MIMIC: Macro minus Majority.}\label{tab:app-rq1-ci-mimic-macro_oriented_minus_binary_majority}
\end{table*}

\begin{table*}[!htbp]\centering\scriptsize
\setlength{\tabcolsep}{2pt}\renewcommand{\arraystretch}{0.95}
\begin{tabular}{@{}ccccc@{}}\toprule
Generator & Ma-14 & Mi-14 & Ma-5 & Mi-5 \\ \midrule
Q35 & +0.0376\,{\fontsize{7.5}{9}\selectfont [-0.0048, +0.0829]} & -0.0563\,{\fontsize{7.5}{9}\selectfont [-0.0798, -0.0318]} & +0.1269\,{\fontsize{7.5}{9}\selectfont [+0.0543, +0.2034]} & +0.1618\,{\fontsize{7.5}{9}\selectfont [+0.0710, +0.2584]} \\
IV35 & +0.0307\,{\fontsize{7.5}{9}\selectfont [-0.0124, +0.0770]} & -0.0797\,{\fontsize{7.5}{9}\selectfont [-0.1077, -0.0515]} & +0.0728\,{\fontsize{7.5}{9}\selectfont [-0.0135, +0.1632]} & +0.0605\,{\fontsize{7.5}{9}\selectfont [-0.0307, +0.1541]} \\
Q25-3 & +0.0815\,{\fontsize{7.5}{9}\selectfont [+0.0455, +0.1144]} & -0.1812\,{\fontsize{7.5}{9}\selectfont [-0.2218, -0.1405]} & +0.1375\,{\fontsize{7.5}{9}\selectfont [+0.0976, +0.1750]} & +0.2195\,{\fontsize{7.5}{9}\selectfont [+0.1486, +0.2843]} \\
Q25-1 & +0.0359\,{\fontsize{7.5}{9}\selectfont [+0.0187, +0.0565]} & -0.2039\,{\fontsize{7.5}{9}\selectfont [-0.2356, -0.1722]} & +0.0000\,{\fontsize{7.5}{9}\selectfont [+0.0000, +0.0000]} & +0.0000\,{\fontsize{7.5}{9}\selectfont [+0.0000, +0.0000]} \\
ESC-RL & +0.0769\,{\fontsize{7.5}{9}\selectfont [+0.0474, +0.1080]} & -0.1753\,{\fontsize{7.5}{9}\selectfont [-0.2099, -0.1408]} & +0.1154\,{\fontsize{7.5}{9}\selectfont [+0.0748, +0.1570]} & +0.1628\,{\fontsize{7.5}{9}\selectfont [+0.0812, +0.2420]} \\
MPO & +0.0775\,{\fontsize{7.5}{9}\selectfont [+0.0415, +0.1138]} & -0.1806\,{\fontsize{7.5}{9}\selectfont [-0.2195, -0.1418]} & +0.1663\,{\fontsize{7.5}{9}\selectfont [+0.0953, +0.2376]} & +0.1753\,{\fontsize{7.5}{9}\selectfont [+0.1040, +0.2410]} \\
R2Gen & +0.0083\,{\fontsize{7.5}{9}\selectfont [+0.0028, +0.0147]} & -0.0372\,{\fontsize{7.5}{9}\selectfont [-0.0502, -0.0244]} & +0.0233\,{\fontsize{7.5}{9}\selectfont [+0.0078, +0.0410]} & +0.0789\,{\fontsize{7.5}{9}\selectfont [+0.0265, +0.1392]} \\
EditGRPO & +0.0141\,{\fontsize{7.5}{9}\selectfont [+0.0001, +0.0287]} & -0.0196\,{\fontsize{7.5}{9}\selectfont [-0.0329, -0.0066]} & +0.0326\,{\fontsize{7.5}{9}\selectfont [+0.0005, +0.0714]} & +0.0620\,{\fontsize{7.5}{9}\selectfont [+0.0139, +0.1201]} \\
DPO-G & +0.0355\,{\fontsize{7.5}{9}\selectfont [+0.0123, +0.0587]} & -0.0484\,{\fontsize{7.5}{9}\selectfont [-0.0649, -0.0318]} & +0.0908\,{\fontsize{7.5}{9}\selectfont [+0.0358, +0.1429]} & +0.1516\,{\fontsize{7.5}{9}\selectfont [+0.0742, +0.2286]} \\
R2CMN & +0.0657\,{\fontsize{7.5}{9}\selectfont [+0.0443, +0.0911]} & -0.2044\,{\fontsize{7.5}{9}\selectfont [-0.2452, -0.1633]} & +0.0961\,{\fontsize{7.5}{9}\selectfont [+0.0624, +0.1318]} & +0.1584\,{\fontsize{7.5}{9}\selectfont [+0.1084, +0.2095]} \\
L15 & +0.0289\,{\fontsize{7.5}{9}\selectfont [+0.0218, +0.0359]} & -0.1939\,{\fontsize{7.5}{9}\selectfont [-0.2325, -0.1551]} & +0.0175\,{\fontsize{7.5}{9}\selectfont [+0.0037, +0.0327]} & +0.0541\,{\fontsize{7.5}{9}\selectfont [+0.0112, +0.1034]} \\
LN-M & +0.0229\,{\fontsize{7.5}{9}\selectfont [+0.0168, +0.0300]} & -0.2053\,{\fontsize{7.5}{9}\selectfont [-0.2435, -0.1671]} & +0.0074\,{\fontsize{7.5}{9}\selectfont [+0.0000, +0.0235]} & +0.0172\,{\fontsize{7.5}{9}\selectfont [+0.0000, +0.0556]} \\
LN-V & +0.0404\,{\fontsize{7.5}{9}\selectfont [+0.0296, +0.0510]} & -0.2535\,{\fontsize{7.5}{9}\selectfont [-0.2966, -0.2118]} & +0.0393\,{\fontsize{7.5}{9}\selectfont [+0.0136, +0.0667]} & +0.0945\,{\fontsize{7.5}{9}\selectfont [+0.0308, +0.1678]} \\
\bottomrule\end{tabular}
\caption{IU: Macro minus Micro.}\label{tab:app-rq1-ci-iu-macro_oriented_minus_micro_oriented}
\end{table*}

\begin{table*}[!htbp]\centering\scriptsize
\setlength{\tabcolsep}{2pt}\renewcommand{\arraystretch}{0.95}
\begin{tabular}{@{}ccccc@{}}\toprule
Generator & Ma-14 & Mi-14 & Ma-5 & Mi-5 \\ \midrule
Q35 & +0.0373\,{\fontsize{7.5}{9}\selectfont [-0.0209, +0.1073]} & -0.0744\,{\fontsize{7.5}{9}\selectfont [-0.1027, -0.0453]} & +0.1258\,{\fontsize{7.5}{9}\selectfont [+0.0258, +0.2337]} & +0.2007\,{\fontsize{7.5}{9}\selectfont [+0.0939, +0.3109]} \\
IV35 & +0.0260\,{\fontsize{7.5}{9}\selectfont [-0.0161, +0.0710]} & -0.0800\,{\fontsize{7.5}{9}\selectfont [-0.1076, -0.0531]} & +0.0540\,{\fontsize{7.5}{9}\selectfont [-0.0292, +0.1414]} & +0.0271\,{\fontsize{7.5}{9}\selectfont [-0.0570, +0.1136]} \\
Q25-3 & +0.1157\,{\fontsize{7.5}{9}\selectfont [+0.0888, +0.1411]} & -0.1785\,{\fontsize{7.5}{9}\selectfont [-0.2205, -0.1368]} & +0.1452\,{\fontsize{7.5}{9}\selectfont [+0.1076, +0.1805]} & +0.2393\,{\fontsize{7.5}{9}\selectfont [+0.1794, +0.2965]} \\
Q25-1 & +0.0359\,{\fontsize{7.5}{9}\selectfont [+0.0187, +0.0565]} & -0.2046\,{\fontsize{7.5}{9}\selectfont [-0.2365, -0.1730]} & +0.0000\,{\fontsize{7.5}{9}\selectfont [+0.0000, +0.0000]} & +0.0000\,{\fontsize{7.5}{9}\selectfont [+0.0000, +0.0000]} \\
ESC-RL & +0.1178\,{\fontsize{7.5}{9}\selectfont [+0.0883, +0.1456]} & -0.1894\,{\fontsize{7.5}{9}\selectfont [-0.2304, -0.1489]} & +0.1535\,{\fontsize{7.5}{9}\selectfont [+0.1150, +0.1913]} & +0.2654\,{\fontsize{7.5}{9}\selectfont [+0.2089, +0.3202]} \\
MPO & +0.1060\,{\fontsize{7.5}{9}\selectfont [+0.0728, +0.1387]} & -0.1780\,{\fontsize{7.5}{9}\selectfont [-0.2188, -0.1375]} & +0.1744\,{\fontsize{7.5}{9}\selectfont [+0.1045, +0.2439]} & +0.1957\,{\fontsize{7.5}{9}\selectfont [+0.1371, +0.2540]} \\
R2Gen & +0.0083\,{\fontsize{7.5}{9}\selectfont [+0.0028, +0.0147]} & -0.0378\,{\fontsize{7.5}{9}\selectfont [-0.0509, -0.0249]} & +0.0233\,{\fontsize{7.5}{9}\selectfont [+0.0078, +0.0410]} & +0.0789\,{\fontsize{7.5}{9}\selectfont [+0.0265, +0.1392]} \\
EditGRPO & +0.0403\,{\fontsize{7.5}{9}\selectfont [+0.0132, +0.0683]} & +0.0141\,{\fontsize{7.5}{9}\selectfont [-0.0067, +0.0358]} & +0.0548\,{\fontsize{7.5}{9}\selectfont [+0.0066, +0.1100]} & +0.0820\,{\fontsize{7.5}{9}\selectfont [+0.0160, +0.1649]} \\
DPO-G & +0.0733\,{\fontsize{7.5}{9}\selectfont [+0.0380, +0.1077]} & -0.0577\,{\fontsize{7.5}{9}\selectfont [-0.0794, -0.0354]} & +0.1062\,{\fontsize{7.5}{9}\selectfont [+0.0439, +0.1668]} & +0.1912\,{\fontsize{7.5}{9}\selectfont [+0.0992, +0.2832]} \\
R2CMN & +0.0659\,{\fontsize{7.5}{9}\selectfont [+0.0445, +0.0913]} & -0.2025\,{\fontsize{7.5}{9}\selectfont [-0.2436, -0.1614]} & +0.0961\,{\fontsize{7.5}{9}\selectfont [+0.0624, +0.1318]} & +0.1584\,{\fontsize{7.5}{9}\selectfont [+0.1084, +0.2095]} \\
L15 & +0.0284\,{\fontsize{7.5}{9}\selectfont [+0.0213, +0.0355]} & -0.2011\,{\fontsize{7.5}{9}\selectfont [-0.2398, -0.1622]} & +0.0175\,{\fontsize{7.5}{9}\selectfont [+0.0037, +0.0327]} & +0.0541\,{\fontsize{7.5}{9}\selectfont [+0.0112, +0.1034]} \\
LN-M & +0.0229\,{\fontsize{7.5}{9}\selectfont [+0.0168, +0.0300]} & -0.2053\,{\fontsize{7.5}{9}\selectfont [-0.2435, -0.1671]} & +0.0074\,{\fontsize{7.5}{9}\selectfont [+0.0000, +0.0235]} & +0.0172\,{\fontsize{7.5}{9}\selectfont [+0.0000, +0.0556]} \\
LN-V & +0.0404\,{\fontsize{7.5}{9}\selectfont [+0.0296, +0.0510]} & -0.2548\,{\fontsize{7.5}{9}\selectfont [-0.2979, -0.2128]} & +0.0393\,{\fontsize{7.5}{9}\selectfont [+0.0136, +0.0667]} & +0.0945\,{\fontsize{7.5}{9}\selectfont [+0.0308, +0.1678]} \\
\bottomrule\end{tabular}
\caption{IU: Macro minus Majority.}\label{tab:app-rq1-ci-iu-macro_oriented_minus_binary_majority}
\end{table*}

\subsection{Capacity and Constant-Prediction Controls}
\label{app:rq1-controls}
The corresponding results are reported in Tables~\ref{tab:app-rq1-controls-mimic}, \ref{tab:app-rq1-controls-iu}.

Label-specific macro selection is a higher-capacity comparison with
14 threshold parameters, rather than a controlled change of objective
within the shared one-parameter family.
Its fixed-mask macro-F1 matches the corresponding label-specific result
in the shared-threshold export for every configuration.
We additionally report always-positive, always-negative, and the exported
constant predictor. The constant predictor coincides with always-positive
in every evaluated configuration. Always-negative macro-F1 is zero.
We retain the exported nonconstant control (NC) and the number of labels
flagged as degenerate; NC is reported as a supplementary diagnostic.

Shared macro selection exceeds always-positive in all MIMIC configurations
but only six IU configurations. Label-specific calibration also exceeds
that constant baseline in 13/13 MIMIC and 6/13 IU configurations.
The exported nonconstant control equals unrestricted label-specific
calibration on MIMIC; on IU its mean macro-F1 is 0.1318 versus 0.1381.
These controls qualify the gains over majority without changing the
same-capacity comparison between shared objectives.

\begin{table*}[!htbp]\centering\scriptsize
\setlength{\tabcolsep}{3pt}\renewcommand{\arraystretch}{0.95}
\begin{tabular}{@{}cccccccc@{}}\toprule
Generator & Maj & Shared Ma & Shared Mi & Label Ma & NC & Const/AP & Deg. \\ \midrule
Q25-3 & 0.3637 & 0.4212 & 0.4036 & 0.4206 & 0.4206 & 0.2172 & 0 \\
Q25-1 & 0.2885 & 0.3651 & 0.3614 & 0.4002 & 0.4002 & 0.2172 & 0 \\
Q35 & 0.3547 & 0.3940 & 0.3329 & 0.4310 & 0.4310 & 0.2172 & 0 \\
IV35 & 0.3770 & 0.4115 & 0.3858 & 0.4269 & 0.4269 & 0.2172 & 0 \\
ESC-RL & 0.3682 & 0.4442 & 0.4322 & 0.4612 & 0.4612 & 0.2172 & 0 \\
MPO & 0.3601 & 0.4148 & 0.3998 & 0.4444 & 0.4444 & 0.2172 & 0 \\
EditGRPO & 0.3748 & 0.4302 & 0.4078 & 0.4659 & 0.4659 & 0.2172 & 0 \\
DPO-G & 0.3452 & 0.4312 & 0.4116 & 0.4451 & 0.4451 & 0.2172 & 0 \\
R2Gen & 0.3698 & 0.4150 & 0.4070 & 0.4450 & 0.4450 & 0.2172 & 0 \\
R2CMN & 0.3407 & 0.4236 & 0.4047 & 0.4365 & 0.4365 & 0.2172 & 0 \\
L15 & 0.4029 & 0.4876 & 0.4719 & 0.4719 & 0.4719 & 0.2172 & 0 \\
LN-M & 0.4274 & 0.4934 & 0.4934 & 0.4969 & 0.4969 & 0.2172 & 0 \\
LN-V & 0.4047 & 0.4826 & 0.4785 & 0.4753 & 0.4753 & 0.2172 & 0 \\
\bottomrule\end{tabular}
\caption{MIMIC: 14-label macro-F1 under the common reference-only mask. Label Ma is label-specific calibration; NC is the exported nonconstant control. Const/AP denotes the coincident constant/always-positive baseline; Deg. is the exported degenerate-label count. Always-negative scores are zero.}\label{tab:app-rq1-controls-mimic}
\end{table*}

\begin{table*}[!htbp]\centering\scriptsize
\setlength{\tabcolsep}{3pt}\renewcommand{\arraystretch}{0.95}
\begin{tabular}{@{}cccccccc@{}}\toprule
Generator & Maj & Shared Ma & Shared Mi & Label Ma & NC & Const/AP & Deg. \\ \midrule
Q35 & 0.2106 & 0.2479 & 0.2102 & 0.2692 & 0.2454 & 0.1210 & 1 \\
IV35 & 0.2072 & 0.2332 & 0.2025 & 0.2693 & 0.2525 & 0.1210 & 1 \\
Q25-3 & 0.0556 & 0.1713 & 0.0899 & 0.1884 & 0.1884 & 0.1210 & 0 \\
Q25-1 & 0.0550 & 0.0909 & 0.0550 & 0.0909 & 0.0884 & 0.1210 & 5 \\
ESC-RL & 0.0556 & 0.1735 & 0.0965 & 0.1665 & 0.1665 & 0.1210 & 0 \\
MPO & 0.0552 & 0.1612 & 0.0837 & 0.1516 & 0.1516 & 0.1210 & 1 \\
R2Gen & 0.0550 & 0.0633 & 0.0550 & 0.0733 & 0.0687 & 0.1210 & 9 \\
EditGRPO & 0.0526 & 0.0929 & 0.0788 & 0.0928 & 0.0803 & 0.1210 & 4 \\
DPO-G & 0.0550 & 0.1283 & 0.0928 & 0.1144 & 0.1144 & 0.1210 & 1 \\
R2CMN & 0.0550 & 0.1209 & 0.0552 & 0.1230 & 0.1230 & 0.1210 & 1 \\
L15 & 0.0550 & 0.0833 & 0.0545 & 0.0828 & 0.0766 & 0.1210 & 10 \\
LN-M & 0.0550 & 0.0779 & 0.0550 & 0.0779 & 0.0648 & 0.1210 & 7 \\
LN-V & 0.0550 & 0.0954 & 0.0550 & 0.0950 & 0.0929 & 0.1210 & 8 \\
\bottomrule\end{tabular}
\caption{IU: 14-label macro-F1 under the common reference-only mask. Label Ma is label-specific calibration; NC is the exported nonconstant control. Const/AP denotes the coincident constant/always-positive baseline; Deg. is the exported degenerate-label count. Always-negative scores are zero.}\label{tab:app-rq1-controls-iu}
\end{table*}

\subsection{Candidate-Utility Comparisons}
\label{app:rq1-utilities}
We hold candidate pools, leave-one-out consensus aggregation, and
candidate constraints fixed, and compare CheXbert-14 (CX), RadGraph (RG),
and ROUGE-L (RL) utilities. Finding metrics are re-extracted by CheXbert
from the selected narrative, using the same reference-only mask as the
aggregate utility matrix. They do not evaluate a separate structured
field attached to the selected report.

Finding contrasts and paired intervals appear in Tables~\ref{tab:app-rq1-utility-ci-mimic-rg-cx}--\ref{tab:app-rq1-utility-ci-mimic-rg-rl} and~\ref{tab:app-rq1-utility-ci-iu-rg-cx}--\ref{tab:app-rq1-utility-ci-iu-rg-rl}.
Report-metric point differences appear in Tables~\ref{tab:app-rq1-utility-report-mimic-rg-cx}--\ref{tab:app-rq1-utility-report-mimic-rg-rl} and~\ref{tab:app-rq1-utility-report-iu-rg-cx}--\ref{tab:app-rq1-utility-report-iu-rg-rl}.

\paragraph{Statistical protocol.}

For utility finding metrics, paired cluster bootstrap uses patients
(subject IDs) on MIMIC and studies on IU. Whole clusters are sampled
with replacement to the original cluster count, with 10,000 draws and
seed 12345. A shared multiplicity matrix within each generator pairs
all utility conditions. The 95\% intervals are the 2.5th and 97.5th
percentiles of score differences. The exported two-sided bootstrap
proportion is $\min\{1,2\min[\Pr(\Delta\le0),\Pr(\Delta\ge0)]\}$.
The three contrasts are specified in advance, but intervals and
$p$-values are uncorrected for multiple comparisons.
We show intervals rather than rounding an exported zero proportion
into a claim of $p=0$.
Report-metric differences are point estimates: the result export
contains corpus summaries without resampling intervals.

\begin{table*}[!htbp]\centering\scriptsize
\setlength{\tabcolsep}{2pt}\renewcommand{\arraystretch}{0.95}
\begin{tabular}{@{}ccccc@{}}\toprule
Generator & Ma-14 & Mi-14 & Ma-5 & Mi-5 \\ \midrule
Q25-3 & -0.0212\,{\fontsize{7.5}{9}\selectfont [-0.0359, -0.0069]} & -0.0374\,{\fontsize{7.5}{9}\selectfont [-0.0471, -0.0277]} & -0.0183\,{\fontsize{7.5}{9}\selectfont [-0.0412, +0.0083]} & -0.0420\,{\fontsize{7.5}{9}\selectfont [-0.0594, -0.0259]} \\
Q25-1 & -0.0198\,{\fontsize{7.5}{9}\selectfont [-0.0316, -0.0077]} & -0.0390\,{\fontsize{7.5}{9}\selectfont [-0.0505, -0.0273]} & -0.0187\,{\fontsize{7.5}{9}\selectfont [-0.0364, +0.0011]} & -0.0385\,{\fontsize{7.5}{9}\selectfont [-0.0574, -0.0195]} \\
Q35 & -0.0104\,{\fontsize{7.5}{9}\selectfont [-0.0290, +0.0085]} & +0.0260\,{\fontsize{7.5}{9}\selectfont [+0.0135, +0.0397]} & -0.0224\,{\fontsize{7.5}{9}\selectfont [-0.0510, +0.0044]} & +0.0011\,{\fontsize{7.5}{9}\selectfont [-0.0190, +0.0202]} \\
IV35 & -0.0285\,{\fontsize{7.5}{9}\selectfont [-0.0459, -0.0126]} & -0.0286\,{\fontsize{7.5}{9}\selectfont [-0.0432, -0.0149]} & -0.0386\,{\fontsize{7.5}{9}\selectfont [-0.0638, -0.0130]} & -0.0324\,{\fontsize{7.5}{9}\selectfont [-0.0504, -0.0155]} \\
ESC-RL & -0.0253\,{\fontsize{7.5}{9}\selectfont [-0.0382, -0.0137]} & -0.0377\,{\fontsize{7.5}{9}\selectfont [-0.0469, -0.0284]} & -0.0341\,{\fontsize{7.5}{9}\selectfont [-0.0542, -0.0147]} & -0.0431\,{\fontsize{7.5}{9}\selectfont [-0.0592, -0.0276]} \\
MPO & -0.0320\,{\fontsize{7.5}{9}\selectfont [-0.0445, -0.0211]} & -0.0385\,{\fontsize{7.5}{9}\selectfont [-0.0483, -0.0288]} & -0.0507\,{\fontsize{7.5}{9}\selectfont [-0.0702, -0.0329]} & -0.0457\,{\fontsize{7.5}{9}\selectfont [-0.0608, -0.0314]} \\
EditGRPO & -0.0231\,{\fontsize{7.5}{9}\selectfont [-0.0382, -0.0084]} & -0.0327\,{\fontsize{7.5}{9}\selectfont [-0.0426, -0.0228]} & -0.0226\,{\fontsize{7.5}{9}\selectfont [-0.0455, +0.0003]} & -0.0371\,{\fontsize{7.5}{9}\selectfont [-0.0520, -0.0231]} \\
\bottomrule\end{tabular}
\caption{MIMIC: RG-CX finding contrasts.}\label{tab:app-rq1-utility-ci-mimic-rg-cx}
\end{table*}

\begin{table*}[!htbp]\centering\scriptsize
\setlength{\tabcolsep}{2pt}\renewcommand{\arraystretch}{0.95}
\begin{tabular}{@{}ccccc@{}}\toprule
Generator & Ma-14 & Mi-14 & Ma-5 & Mi-5 \\ \midrule
Q25-3 & -0.0194\,{\fontsize{7.5}{9}\selectfont [-0.0382, -0.0016]} & -0.0446\,{\fontsize{7.5}{9}\selectfont [-0.0568, -0.0327]} & -0.0159\,{\fontsize{7.5}{9}\selectfont [-0.0450, +0.0110]} & -0.0491\,{\fontsize{7.5}{9}\selectfont [-0.0726, -0.0290]} \\
Q25-1 & -0.0260\,{\fontsize{7.5}{9}\selectfont [-0.0357, -0.0157]} & -0.0456\,{\fontsize{7.5}{9}\selectfont [-0.0564, -0.0348]} & -0.0251\,{\fontsize{7.5}{9}\selectfont [-0.0442, -0.0042]} & -0.0450\,{\fontsize{7.5}{9}\selectfont [-0.0645, -0.0262]} \\
Q35 & -0.0017\,{\fontsize{7.5}{9}\selectfont [-0.0208, +0.0167]} & +0.0278\,{\fontsize{7.5}{9}\selectfont [+0.0148, +0.0415]} & -0.0115\,{\fontsize{7.5}{9}\selectfont [-0.0417, +0.0150]} & -0.0051\,{\fontsize{7.5}{9}\selectfont [-0.0244, +0.0131]} \\
IV35 & -0.0346\,{\fontsize{7.5}{9}\selectfont [-0.0509, -0.0192]} & -0.0395\,{\fontsize{7.5}{9}\selectfont [-0.0536, -0.0265]} & -0.0568\,{\fontsize{7.5}{9}\selectfont [-0.0854, -0.0291]} & -0.0518\,{\fontsize{7.5}{9}\selectfont [-0.0715, -0.0336]} \\
ESC-RL & -0.0315\,{\fontsize{7.5}{9}\selectfont [-0.0468, -0.0167]} & -0.0460\,{\fontsize{7.5}{9}\selectfont [-0.0584, -0.0337]} & -0.0426\,{\fontsize{7.5}{9}\selectfont [-0.0704, -0.0165]} & -0.0566\,{\fontsize{7.5}{9}\selectfont [-0.0782, -0.0364]} \\
MPO & -0.0213\,{\fontsize{7.5}{9}\selectfont [-0.0358, -0.0069]} & -0.0434\,{\fontsize{7.5}{9}\selectfont [-0.0539, -0.0331]} & -0.0442\,{\fontsize{7.5}{9}\selectfont [-0.0639, -0.0254]} & -0.0507\,{\fontsize{7.5}{9}\selectfont [-0.0683, -0.0344]} \\
EditGRPO & -0.0183\,{\fontsize{7.5}{9}\selectfont [-0.0336, -0.0017]} & -0.0454\,{\fontsize{7.5}{9}\selectfont [-0.0567, -0.0338]} & -0.0268\,{\fontsize{7.5}{9}\selectfont [-0.0491, -0.0022]} & -0.0527\,{\fontsize{7.5}{9}\selectfont [-0.0718, -0.0340]} \\
\bottomrule\end{tabular}
\caption{MIMIC: RL-CX finding contrasts.}\label{tab:app-rq1-utility-ci-mimic-rl-cx}
\end{table*}

\begin{table*}[!htbp]\centering\scriptsize
\setlength{\tabcolsep}{2pt}\renewcommand{\arraystretch}{0.95}
\begin{tabular}{@{}ccccc@{}}\toprule
Generator & Ma-14 & Mi-14 & Ma-5 & Mi-5 \\ \midrule
Q25-3 & -0.0018\,{\fontsize{7.5}{9}\selectfont [-0.0188, +0.0151]} & +0.0072\,{\fontsize{7.5}{9}\selectfont [-0.0035, +0.0178]} & -0.0024\,{\fontsize{7.5}{9}\selectfont [-0.0280, +0.0250]} & +0.0071\,{\fontsize{7.5}{9}\selectfont [-0.0115, +0.0267]} \\
Q25-1 & +0.0062\,{\fontsize{7.5}{9}\selectfont [-0.0069, +0.0198]} & +0.0066\,{\fontsize{7.5}{9}\selectfont [-0.0030, +0.0164]} & +0.0064\,{\fontsize{7.5}{9}\selectfont [-0.0156, +0.0286]} & +0.0065\,{\fontsize{7.5}{9}\selectfont [-0.0100, +0.0233]} \\
Q35 & -0.0087\,{\fontsize{7.5}{9}\selectfont [-0.0208, +0.0027]} & -0.0018\,{\fontsize{7.5}{9}\selectfont [-0.0093, +0.0058]} & -0.0108\,{\fontsize{7.5}{9}\selectfont [-0.0320, +0.0085]} & +0.0062\,{\fontsize{7.5}{9}\selectfont [-0.0092, +0.0217]} \\
IV35 & +0.0061\,{\fontsize{7.5}{9}\selectfont [-0.0058, +0.0183]} & +0.0109\,{\fontsize{7.5}{9}\selectfont [-0.0014, +0.0234]} & +0.0182\,{\fontsize{7.5}{9}\selectfont [-0.0066, +0.0438]} & +0.0195\,{\fontsize{7.5}{9}\selectfont [+0.0024, +0.0371]} \\
ESC-RL & +0.0062\,{\fontsize{7.5}{9}\selectfont [-0.0072, +0.0194]} & +0.0083\,{\fontsize{7.5}{9}\selectfont [-0.0018, +0.0184]} & +0.0085\,{\fontsize{7.5}{9}\selectfont [-0.0126, +0.0292]} & +0.0135\,{\fontsize{7.5}{9}\selectfont [-0.0036, +0.0309]} \\
MPO & -0.0108\,{\fontsize{7.5}{9}\selectfont [-0.0276, +0.0046]} & +0.0050\,{\fontsize{7.5}{9}\selectfont [-0.0038, +0.0135]} & -0.0065\,{\fontsize{7.5}{9}\selectfont [-0.0287, +0.0150]} & +0.0050\,{\fontsize{7.5}{9}\selectfont [-0.0106, +0.0207]} \\
EditGRPO & -0.0048\,{\fontsize{7.5}{9}\selectfont [-0.0241, +0.0120]} & +0.0127\,{\fontsize{7.5}{9}\selectfont [+0.0025, +0.0224]} & +0.0042\,{\fontsize{7.5}{9}\selectfont [-0.0234, +0.0303]} & +0.0155\,{\fontsize{7.5}{9}\selectfont [-0.0019, +0.0332]} \\
\bottomrule\end{tabular}
\caption{MIMIC: RG-RL finding contrasts.}\label{tab:app-rq1-utility-ci-mimic-rg-rl}
\end{table*}

\begin{table*}[!tp]
\centering
\begin{minipage}[t]{0.48\textwidth}
\centering\scriptsize
\setlength{\tabcolsep}{1.5pt}\renewcommand{\arraystretch}{0.95}
\begin{tabular}{@{}cccccc@{}}\toprule
Generator & RG & RL & GREEN & RaTE & B4 \\ \midrule
Q25-3 & +.0423 & +.0336 & +.0056 & +.0048 & +.0073 \\
Q25-1 & +.0401 & +.0328 & +.0082 & +.0062 & +.0150 \\
Q35 & +.0340 & +.0129 & -.0105 & +.0175 & +.0059 \\
IV35 & +.0093 & -.0039 & -.0077 & -.0017 & +.0049 \\
ESC-RL & +.0370 & +.0320 & +.0074 & +.0080 & +.0060 \\
MPO & +.0425 & +.0300 & +.0011 & +.0051 & +.0056 \\
EditGRPO & +.0445 & +.0394 & +.0081 & +.0105 & +.0119 \\
DPO-G & +.0561 & +.0428 & +.0100 & +.0076 & +.0476 \\
R2Gen & +.0415 & +.0141 & +.0214 & +.0152 & +.0041 \\
R2CMN & +.0391 & +.0145 & +.0252 & +.0174 & +.0024 \\
L15 & +.0628 & +.0191 & +.0077 & +.0403 & -.0009 \\
LN-M & +.0468 & +.0101 & +.0066 & +.0113 & -.0050 \\
LN-V & +.0606 & +.0173 & +.0063 & +.0379 & -.0020 \\
\bottomrule\end{tabular}
\caption{MIMIC: RG-CX report-metric differences. RaTE/B4 denote RaTEScore/BLEU-4. All columns evaluate the narrative selected by the row utility; no report-metric intervals are available.}\label{tab:app-rq1-utility-report-mimic-rg-cx}

\end{minipage}
\hfill
\begin{minipage}[t]{0.48\textwidth}
\centering\scriptsize
\setlength{\tabcolsep}{1.5pt}\renewcommand{\arraystretch}{0.95}
\begin{tabular}{@{}cccccc@{}}\toprule
Generator & RG & RL & GREEN & RaTE & B4 \\ \midrule
Q25-3 & +.0166 & +.0480 & +.0052 & +.0020 & +.0137 \\
Q25-1 & +.0104 & +.0525 & +.0057 & +.0028 & +.0210 \\
Q35 & +.0237 & +.0138 & -.0082 & +.0141 & +.0059 \\
IV35 & +.0000 & -.0032 & -.0094 & -.0033 & +.0062 \\
ESC-RL & +.0118 & +.0517 & +.0044 & +.0035 & +.0163 \\
MPO & +.0170 & +.0449 & +.0034 & -.0025 & +.0120 \\
EditGRPO & +.0167 & +.0585 & +.0028 & +.0054 & +.0219 \\
DPO-G & +.0229 & +.0626 & +.0081 & +.0005 & +.0627 \\
R2Gen & +.0164 & +.0298 & +.0115 & +.0037 & +.0085 \\
R2CMN & +.0160 & +.0286 & +.0195 & +.0109 & +.0074 \\
L15 & +.0388 & +.0329 & -.0020 & +.0368 & +.0048 \\
LN-M & +.0135 & +.0253 & +.0067 & +.0030 & +.0050 \\
LN-V & +.0336 & +.0345 & -.0016 & +.0345 & +.0047 \\
\bottomrule\end{tabular}
\caption{MIMIC: RL-CX report-metric differences. RaTE/B4 denote RaTEScore/BLEU-4. All columns evaluate the narrative selected by the row utility; no report-metric intervals are available.}\label{tab:app-rq1-utility-report-mimic-rl-cx}

\end{minipage}
\end{table*}

\begin{table}[!htbp]\centering\scriptsize
\setlength{\tabcolsep}{1.5pt}\renewcommand{\arraystretch}{0.95}
\begin{tabular}{@{}cccccc@{}}\toprule
Generator & RG & RL & GREEN & RaTE & B4 \\ \midrule
Q25-3 & +.0257 & -.0144 & +.0005 & +.0028 & -.0064 \\
Q25-1 & +.0298 & -.0197 & +.0025 & +.0034 & -.0061 \\
Q35 & +.0103 & -.0009 & -.0023 & +.0034 & -.0000 \\
IV35 & +.0092 & -.0007 & +.0017 & +.0016 & -.0014 \\
ESC-RL & +.0252 & -.0197 & +.0030 & +.0046 & -.0103 \\
MPO & +.0255 & -.0150 & -.0023 & +.0076 & -.0064 \\
EditGRPO & +.0278 & -.0191 & +.0053 & +.0051 & -.0100 \\
DPO-G & +.0332 & -.0198 & +.0019 & +.0071 & -.0151 \\
R2Gen & +.0251 & -.0157 & +.0099 & +.0116 & -.0044 \\
R2CMN & +.0231 & -.0141 & +.0058 & +.0065 & -.0050 \\
L15 & +.0240 & -.0137 & +.0097 & +.0036 & -.0058 \\
LN-M & +.0333 & -.0151 & -.0001 & +.0083 & -.0099 \\
LN-V & +.0270 & -.0172 & +.0079 & +.0034 & -.0066 \\
\bottomrule\end{tabular}
\caption{MIMIC: RG-RL report-metric differences. RaTE/B4 denote RaTEScore/BLEU-4. All columns evaluate the narrative selected by the row utility; no report-metric intervals are available.}\label{tab:app-rq1-utility-report-mimic-rg-rl}
\end{table}

\begin{table*}[!htbp]\centering\scriptsize
\setlength{\tabcolsep}{2pt}\renewcommand{\arraystretch}{0.95}
\begin{tabular}{@{}ccccc@{}}\toprule
Generator & Ma-14 & Mi-14 & Ma-5 & Mi-5 \\ \midrule
Q35 & +0.0128\,{\fontsize{7.5}{9}\selectfont [-0.0181, +0.0441]} & -0.0131\,{\fontsize{7.5}{9}\selectfont [-0.0399, +0.0135]} & +0.0049\,{\fontsize{7.5}{9}\selectfont [-0.0598, +0.0755]} & -0.0308\,{\fontsize{7.5}{9}\selectfont [-0.1000, +0.0377]} \\
IV35 & +0.0072\,{\fontsize{7.5}{9}\selectfont [-0.0362, +0.0447]} & +0.1695\,{\fontsize{7.5}{9}\selectfont [+0.1272, +0.2106]} & -0.0239\,{\fontsize{7.5}{9}\selectfont [-0.0860, +0.0331]} & +0.0589\,{\fontsize{7.5}{9}\selectfont [-0.0077, +0.1221]} \\
Q25-3 & -0.0095\,{\fontsize{7.5}{9}\selectfont [-0.0219, -0.0002]} & -0.0038\,{\fontsize{7.5}{9}\selectfont [-0.0127, +0.0050]} & -0.0056\,{\fontsize{7.5}{9}\selectfont [-0.0179, +0.0000]} & -0.0134\,{\fontsize{7.5}{9}\selectfont [-0.0438, +0.0000]} \\
Q25-1 & +0.0000\,{\fontsize{7.5}{9}\selectfont [+0.0000, +0.0000]} & +0.0000\,{\fontsize{7.5}{9}\selectfont [+0.0000, +0.0000]} & +0.0000\,{\fontsize{7.5}{9}\selectfont [+0.0000, +0.0000]} & +0.0000\,{\fontsize{7.5}{9}\selectfont [+0.0000, +0.0000]} \\
ESC-RL & -0.0029\,{\fontsize{7.5}{9}\selectfont [-0.0123, +0.0044]} & -0.0089\,{\fontsize{7.5}{9}\selectfont [-0.0182, -0.0004]} & +0.0051\,{\fontsize{7.5}{9}\selectfont [+0.0000, +0.0164]} & +0.0128\,{\fontsize{7.5}{9}\selectfont [+0.0000, +0.0428]} \\
MPO & +0.0081\,{\fontsize{7.5}{9}\selectfont [+0.0002, +0.0231]} & +0.0131\,{\fontsize{7.5}{9}\selectfont [+0.0050, +0.0224]} & +0.0000\,{\fontsize{7.5}{9}\selectfont [+0.0000, +0.0000]} & +0.0000\,{\fontsize{7.5}{9}\selectfont [+0.0000, +0.0000]} \\
R2Gen & -0.0008\,{\fontsize{7.5}{9}\selectfont [-0.0014, -0.0002]} & -0.0158\,{\fontsize{7.5}{9}\selectfont [-0.0258, -0.0069]} & +0.0000\,{\fontsize{7.5}{9}\selectfont [+0.0000, +0.0000]} & +0.0000\,{\fontsize{7.5}{9}\selectfont [+0.0000, +0.0000]} \\
\bottomrule\end{tabular}
\caption{IU: RG-CX finding contrasts.}\label{tab:app-rq1-utility-ci-iu-rg-cx}
\end{table*}

\begin{table*}[!htbp]\centering\scriptsize
\setlength{\tabcolsep}{2pt}\renewcommand{\arraystretch}{0.95}
\begin{tabular}{@{}ccccc@{}}\toprule
Generator & Ma-14 & Mi-14 & Ma-5 & Mi-5 \\ \midrule
Q35 & +0.0370\,{\fontsize{7.5}{9}\selectfont [-0.0301, +0.0879]} & -0.0522\,{\fontsize{7.5}{9}\selectfont [-0.0820, -0.0217]} & -0.0037\,{\fontsize{7.5}{9}\selectfont [-0.0747, +0.0803]} & -0.0434\,{\fontsize{7.5}{9}\selectfont [-0.1083, +0.0235]} \\
IV35 & +0.0094\,{\fontsize{7.5}{9}\selectfont [-0.0292, +0.0489]} & +0.1165\,{\fontsize{7.5}{9}\selectfont [+0.0750, +0.1578]} & -0.0061\,{\fontsize{7.5}{9}\selectfont [-0.0577, +0.0440]} & +0.0194\,{\fontsize{7.5}{9}\selectfont [-0.0474, +0.0857]} \\
Q25-3 (3 epoch) & -0.0081\,{\fontsize{7.5}{9}\selectfont [-0.0208, +0.0019]} & -0.0060\,{\fontsize{7.5}{9}\selectfont [-0.0159, +0.0036]} & -0.0056\,{\fontsize{7.5}{9}\selectfont [-0.0179, +0.0000]} & -0.0134\,{\fontsize{7.5}{9}\selectfont [-0.0438, +0.0000]} \\
Q25-1 & +0.0000\,{\fontsize{7.5}{9}\selectfont [+0.0000, +0.0000]} & +0.0000\,{\fontsize{7.5}{9}\selectfont [+0.0000, +0.0000]} & +0.0000\,{\fontsize{7.5}{9}\selectfont [+0.0000, +0.0000]} & +0.0000\,{\fontsize{7.5}{9}\selectfont [+0.0000, +0.0000]} \\
ESC-RL & -0.0041\,{\fontsize{7.5}{9}\selectfont [-0.0129, +0.0006]} & +0.0004\,{\fontsize{7.5}{9}\selectfont [-0.0054, +0.0054]} & +0.0000\,{\fontsize{7.5}{9}\selectfont [+0.0000, +0.0000]} & +0.0000\,{\fontsize{7.5}{9}\selectfont [+0.0000, +0.0000]} \\
MPO & +0.0006\,{\fontsize{7.5}{9}\selectfont [+0.0001, +0.0012]} & +0.0120\,{\fontsize{7.5}{9}\selectfont [+0.0044, +0.0210]} & +0.0000\,{\fontsize{7.5}{9}\selectfont [+0.0000, +0.0000]} & +0.0000\,{\fontsize{7.5}{9}\selectfont [+0.0000, +0.0000]} \\
R2Gen & +0.0018\,{\fontsize{7.5}{9}\selectfont [-0.0006, +0.0065]} & -0.0059\,{\fontsize{7.5}{9}\selectfont [-0.0129, +0.0000]} & +0.0056\,{\fontsize{7.5}{9}\selectfont [+0.0000, +0.0185]} & +0.0133\,{\fontsize{7.5}{9}\selectfont [+0.0000, +0.0447]} \\
\bottomrule\end{tabular}
\caption{IU: RL-CX finding contrasts.}\label{tab:app-rq1-utility-ci-iu-rl-cx}
\end{table*}

\begin{table*}[!htbp]\centering\scriptsize
\setlength{\tabcolsep}{2pt}\renewcommand{\arraystretch}{0.95}
\begin{tabular}{@{}ccccc@{}}\toprule
Generator & Ma-14 & Mi-14 & Ma-5 & Mi-5 \\ \midrule
Q35 & -0.0242\,{\fontsize{7.5}{9}\selectfont [-0.0731, +0.0351]} & +0.0391\,{\fontsize{7.5}{9}\selectfont [+0.0125, +0.0667]} & +0.0086\,{\fontsize{7.5}{9}\selectfont [-0.0541, +0.0724]} & +0.0126\,{\fontsize{7.5}{9}\selectfont [-0.0378, +0.0654]} \\
IV35 & -0.0022\,{\fontsize{7.5}{9}\selectfont [-0.0593, +0.0496]} & +0.0530\,{\fontsize{7.5}{9}\selectfont [+0.0199, +0.0864]} & -0.0178\,{\fontsize{7.5}{9}\selectfont [-0.0873, +0.0439]} & +0.0396\,{\fontsize{7.5}{9}\selectfont [-0.0164, +0.0989]} \\
Q25-3 & -0.0013\,{\fontsize{7.5}{9}\selectfont [-0.0050, +0.0008]} & +0.0021\,{\fontsize{7.5}{9}\selectfont [-0.0072, +0.0113]} & +0.0000\,{\fontsize{7.5}{9}\selectfont [+0.0000, +0.0000]} & +0.0000\,{\fontsize{7.5}{9}\selectfont [+0.0000, +0.0000]} \\
Q25-1 & +0.0000\,{\fontsize{7.5}{9}\selectfont [+0.0000, +0.0000]} & +0.0000\,{\fontsize{7.5}{9}\selectfont [+0.0000, +0.0000]} & +0.0000\,{\fontsize{7.5}{9}\selectfont [+0.0000, +0.0000]} & +0.0000\,{\fontsize{7.5}{9}\selectfont [+0.0000, +0.0000]} \\
ESC-RL & +0.0012\,{\fontsize{7.5}{9}\selectfont [-0.0012, +0.0055]} & -0.0093\,{\fontsize{7.5}{9}\selectfont [-0.0189, -0.0002]} & +0.0051\,{\fontsize{7.5}{9}\selectfont [+0.0000, +0.0164]} & +0.0128\,{\fontsize{7.5}{9}\selectfont [+0.0000, +0.0428]} \\
MPO & +0.0075\,{\fontsize{7.5}{9}\selectfont [-0.0002, +0.0226]} & +0.0011\,{\fontsize{7.5}{9}\selectfont [-0.0012, +0.0044]} & +0.0000\,{\fontsize{7.5}{9}\selectfont [+0.0000, +0.0000]} & +0.0000\,{\fontsize{7.5}{9}\selectfont [+0.0000, +0.0000]} \\
R2Gen & -0.0026\,{\fontsize{7.5}{9}\selectfont [-0.0073, +0.0000]} & -0.0099\,{\fontsize{7.5}{9}\selectfont [-0.0211, +0.0008]} & -0.0056\,{\fontsize{7.5}{9}\selectfont [-0.0185, +0.0000]} & -0.0133\,{\fontsize{7.5}{9}\selectfont [-0.0447, +0.0000]} \\
\bottomrule\end{tabular}
\caption{IU: RG-RL finding contrasts.}\label{tab:app-rq1-utility-ci-iu-rg-rl}
\end{table*}

\begin{table*}[!tp]
\centering
\begin{minipage}[t]{0.48\textwidth}
\centering\scriptsize
\setlength{\tabcolsep}{1.5pt}\renewcommand{\arraystretch}{0.95}
\begin{tabular}{@{}cccccc@{}}\toprule
Generator & RG & RL & GREEN & RaTE & B4 \\ \midrule
Q35 & +.0298 & +.0007 & +.0017 & +.0088 & +.0027 \\
IV35 & +.0296 & +.0190 & +.0028 & +.0036 & +.0069 \\
Q25-3 & +.1072 & +.0435 & +.0237 & +.0362 & +.0244 \\
Q25-1 & +.0040 & +.0021 & -.0238 & +.0059 & +.0049 \\
ESC-RL & +.1006 & +.0330 & +.0072 & +.0203 & +.0252 \\
MPO & +.1200 & +.0507 & -.0219 & +.0475 & +.0310 \\
R2Gen & +.0501 & +.0072 & +.0495 & +.0273 & +.0092 \\
EditGRPO & +.0319 & +.0048 & -.0159 & +.0118 & +.0032 \\
DPO-G & +.0929 & +.0029 & +.0385 & +.0429 & -.0028 \\
R2CMN & +.0549 & +.0084 & +.0111 & +.0263 & +.0058 \\
L15 & +.0651 & +.0276 & -.0120 & +.0358 & +.0299 \\
LN-M & +.1588 & +.0285 & +.0573 & +.1135 & +.0175 \\
LN-V & +.0843 & +.0155 & -.0072 & +.0583 & +.0406 \\
\bottomrule\end{tabular}
\caption{IU: RG-CX report-metric differences. RaTE/B4 denote RaTEScore/BLEU-4. All columns evaluate the narrative selected by the row utility; no report-metric intervals are available.}\label{tab:app-rq1-utility-report-iu-rg-cx}

\end{minipage}
\hfill
\begin{minipage}[t]{0.48\textwidth}
\centering\scriptsize
\setlength{\tabcolsep}{1.5pt}\renewcommand{\arraystretch}{0.95}
\begin{tabular}{@{}cccccc@{}}\toprule
Generator & RG & RL & GREEN & RaTE & B4 \\ \midrule
Q35 & +.0203 & +.0065 & -.0036 & +.0034 & +.0030 \\
IV35 & +.0115 & +.0195 & +.0048 & +.0069 & +.0074 \\
Q25-3 & +.0410 & +.0551 & +.0061 & +.0004 & +.0283 \\
Q25-1 & -.0065 & +.0076 & -.0024 & +.0001 & +.0200 \\
ESC-RL & +.0506 & +.0536 & +.0058 & -.0102 & +.0400 \\
MPO & +.0508 & +.0501 & +.0048 & +.0168 & +.0228 \\
R2Gen & +.0206 & +.0191 & +.0316 & +.0165 & +.0057 \\
EditGRPO & +.0137 & +.0065 & -.0120 & +.0036 & +.0010 \\
DPO-G & +.0698 & +.0131 & +.0665 & +.0317 & +.0076 \\
R2CMN & +.0348 & +.0151 & +.0182 & +.0143 & +.0088 \\
L15 & -.0022 & +.0569 & -.0557 & +.0230 & +.0298 \\
LN-M & +.1052 & +.0548 & +.0370 & +.0512 & +.0352 \\
LN-V & +.0630 & +.0533 & -.0616 & +.0479 & +.0331 \\
\bottomrule\end{tabular}
\caption{IU: RL-CX report-metric differences. RaTE/B4 denote RaTEScore/BLEU-4. All columns evaluate the narrative selected by the row utility; no report-metric intervals are available.}\label{tab:app-rq1-utility-report-iu-rl-cx}

\end{minipage}
\end{table*}

\begin{table}[!htbp]\centering\scriptsize
\setlength{\tabcolsep}{1.5pt}\renewcommand{\arraystretch}{0.95}
\begin{tabular}{@{}cccccc@{}}\toprule
Generator & RG & RL & GREEN & RaTE & B4 \\ \midrule
Q35 & +.0094 & -.0058 & +.0052 & +.0054 & -.0003 \\
IV35 & +.0181 & -.0005 & -.0020 & -.0033 & -.0005 \\
Q25-3 & +.0662 & -.0116 & +.0176 & +.0357 & -.0039 \\
Q25-1 & +.0105 & -.0055 & -.0214 & +.0058 & -.0151 \\
ESC-RL & +.0500 & -.0206 & +.0014 & +.0305 & -.0148 \\
MPO & +.0693 & +.0006 & -.0267 & +.0307 & +.0082 \\
R2Gen & +.0295 & -.0118 & +.0179 & +.0108 & +.0035 \\
EditGRPO & +.0182 & -.0018 & -.0039 & +.0082 & +.0022 \\
DPO-G & +.0232 & -.0101 & -.0279 & +.0112 & -.0104 \\
R2CMN & +.0201 & -.0067 & -.0071 & +.0121 & -.0030 \\
L15 & +.0673 & -.0293 & +.0437 & +.0128 & +.0001 \\
LN-M & +.0536 & -.0264 & +.0203 & +.0624 & -.0177 \\
LN-V & +.0214 & -.0378 & +.0544 & +.0105 & +.0075 \\
\bottomrule\end{tabular}
\caption{IU: RG-RL report-metric differences. RaTE/B4 denote RaTEScore/BLEU-4. All columns evaluate the narrative selected by the row utility; no report-metric intervals are available.}\label{tab:app-rq1-utility-report-iu-rg-rl}
\end{table}

\subsection{Extended Utility Matrix and Prediction Differences}
The corresponding results are reported in Tables~\ref{tab:app-rq1-matrix-mimic}, \ref{tab:app-rq1-matrix-iu}, \ref{tab:app-rq1-disagreement}.

The extended matrix evaluates nine utilities under the same protocol.
It provides a descriptive comparison rather than a test-selected
recommendation of a deployable utility.
Selection disagreement is measured separately for candidate indices,
report text, and narrative-extracted finding vectors. Different selected
reports need not imply different finding vectors, especially on IU.

\begin{table*}[!htbp]\centering\scriptsize
\setlength{\tabcolsep}{3pt}\renewcommand{\arraystretch}{0.95}
\begin{tabular}{@{}cccccccccc@{}}\toprule
Utility & Ma-14 & Mi-14 & Ma-5 & Mi-5 & RG & RL & GREEN & RaTE & B4 \\ \midrule
chexbert14 & 0.3814 & 0.5928 & 0.4955 & 0.5854 & 0.2942 & 0.3133 & 0.4988 & 0.6197 & 0.0864 \\
chexbert5 & 0.3710 & 0.5747 & 0.4937 & 0.5827 & 0.2927 & 0.3156 & 0.4964 & 0.6186 & 0.0873 \\
rougeL & 0.3571 & 0.5573 & 0.4622 & 0.5417 & 0.3124 & 0.3502 & 0.5023 & 0.6282 & 0.1011 \\
bleu4 & 0.3617 & 0.5601 & 0.4659 & 0.5437 & 0.3109 & 0.3471 & 0.5031 & 0.6284 & 0.1044 \\
embed & 0.3735 & 0.5716 & 0.4846 & 0.5648 & 0.3095 & 0.3323 & 0.5059 & 0.6293 & 0.0979 \\
bertscore & 0.3627 & 0.5640 & 0.4665 & 0.5458 & 0.3152 & 0.3429 & 0.5041 & 0.6291 & 0.0988 \\
radgraph & 0.3591 & 0.5646 & 0.4666 & 0.5521 & 0.3370 & 0.3359 & 0.5057 & 0.6335 & 0.0943 \\
ratescore & 0.3753 & 0.5705 & 0.4894 & 0.5657 & 0.3165 & 0.3348 & 0.5074 & 0.6489 & 0.0960 \\
radeval\_bert & 0.3505 & 0.5428 & 0.4392 & 0.5151 & 0.2910 & 0.3169 & 0.4906 & 0.6150 & 0.0898 \\
\bottomrule\end{tabular}
\caption{MIMIC: extended utility matrix. Entries are arithmetic means over 13 configurations. Every metric scores the narrative selected by the corresponding utility.}\label{tab:app-rq1-matrix-mimic}
\end{table*}

\begin{table*}[!htbp]\centering\scriptsize
\setlength{\tabcolsep}{3pt}\renewcommand{\arraystretch}{0.95}
\begin{tabular}{@{}cccccccccc@{}}\toprule
Utility & Ma-14 & Mi-14 & Ma-5 & Mi-5 & RG & RL & GREEN & RaTE & B4 \\ \midrule
chexbert14 & 0.0732 & 0.5468 & 0.0348 & 0.0366 & 0.3117 & 0.3099 & 0.6007 & 0.6220 & 0.0561 \\
chexbert5 & 0.0759 & 0.5317 & 0.0343 & 0.0331 & 0.3079 & 0.3089 & 0.5970 & 0.6197 & 0.0561 \\
rougeL & 0.0760 & 0.5519 & 0.0340 & 0.0347 & 0.3481 & 0.3415 & 0.6037 & 0.6378 & 0.0747 \\
bleu4 & 0.0748 & 0.5540 & 0.0340 & 0.0355 & 0.3424 & 0.3382 & 0.6020 & 0.6328 & 0.0788 \\
embed & 0.0743 & 0.5549 & 0.0309 & 0.0387 & 0.3547 & 0.3282 & 0.6163 & 0.6416 & 0.0706 \\
bertscore & 0.0756 & 0.5544 & 0.0341 & 0.0395 & 0.3553 & 0.3311 & 0.5921 & 0.6420 & 0.0658 \\
radgraph & 0.0743 & 0.5575 & 0.0333 & 0.0387 & 0.3832 & 0.3287 & 0.6092 & 0.6557 & 0.0713 \\
ratescore & 0.0736 & 0.5416 & 0.0305 & 0.0395 & 0.3618 & 0.3291 & 0.6186 & 0.6665 & 0.0719 \\
radeval\_bert & 0.0785 & 0.5331 & 0.0421 & 0.0561 & 0.3105 & 0.3100 & 0.6009 & 0.6216 & 0.0571 \\
\bottomrule\end{tabular}
\caption{IU: extended utility matrix. Entries are arithmetic means over 13 configurations. Every metric scores the narrative selected by the corresponding utility.}\label{tab:app-rq1-matrix-iu}
\end{table*}

\begin{table}[!htbp]\centering\scriptsize
\setlength{\tabcolsep}{1.5pt}\renewcommand{\arraystretch}{0.95}
\begin{tabular}{@{}cccc@{}}\toprule
Dataset / contrast & Index & Text & Findings \\ \midrule
MIMIC / CX vs RG & 0.9492 & 0.9492 & 0.3572 \\
MIMIC / CX vs RL & 0.9678 & 0.9678 & 0.3868 \\
MIMIC / RG vs RL & 0.8259 & 0.8259 & 0.3062 \\
IU / CX vs RG & 0.9784 & 0.9784 & 0.0844 \\
IU / CX vs RL & 0.9737 & 0.9737 & 0.0846 \\
IU / RG vs RL & 0.8837 & 0.8837 & 0.0631 \\
\bottomrule\end{tabular}
\caption{Mean utility-selection disagreement rates over 13 configurations. Index/text disagreements are distinct from finding-vector disagreements.}\label{tab:app-rq1-disagreement}
\end{table}

\paragraph{Reproducibility boundary.}
The result exports contain development curves and per-label control
statistics, but not per-input predictions or all per-generator absolute
utility scores. We therefore report directly exported paired utility
differences rather than reconstructing unavailable absolute scores.
The tables support numerical auditing of the reported comparisons;
independent recomputation of bootstrap intervals requires the original
predictions and cluster identifiers.

\section{Implementation and Evaluation Details}
\label{app:implementation}

\subsection{Data and Representations}
\label{app:data-representations}
The evaluation inputs comprise 2,047 MIMIC-CXR images (615 development,
1,432 test) and 638 IU-Xray images (192 development, 446 test).
The statistical clustering units are MIMIC patients and IU studies.
The IU development masks checked for four generator tags are identical.
The reported set contains thirteen configurations per benchmark; intermediate
ESC-RL and MPO checkpoints and the auxiliary IU GRPO-full configuration are excluded.
Qwen2.5-VL-7B (3 epoch) is the supervised-fine-tuned candidate generator named SFT-full
in the raw exports. It is distinct from the realization model.

Present findings map to one; absent, uncertain, and not-mentioned states map to zero.
Reference-based F1 and change decomposition exclude reference-uncertain cells.
In the scoring implementation, missing reference states are encoded as $-2$
and counted as negative, rather than excluded.
The fixed five-label evaluation subset is Atelectasis, Cardiomegaly,
Consolidation, Edema, and Pleural Effusion. It is evaluated on the same predictions
without threshold or coefficient reselection and differs from $R_5$.

\subsection{Image Signals and State Estimation}
\label{app:state-estimation}
For MIMIC, the image-side signal is the mean of the logits from the stage-0
classifier and CLIP-SFT probability arrays. For IU it is the logit of the
IU DenseNet probability array.
For each finding, the eight input features are
\begin{equation}
\begin{aligned}
f_L=[&\ell_L,\ell_L,r_L,h_L,\pi_L,\\
     &\ell_Lr_L,\ell_Lh_L,r_L-h_L].
\end{aligned}
\end{equation}
Here $r_L$ is the positive rate among 32 candidate decision labels,
$h_L$ is the positive rate among their narrative/think labels, and $\pi_L$ is
training prevalence (fallback $10^{-4}$ when missing).
The image logit $\ell_L$ is deliberately reported twice because the implementation
contains duplicate columns. With L2 regularization, removing the duplicate need
not preserve the fit; the reported experiments retain it.
This fused state estimate is not a pure-image baseline.

Each label uses logistic regression with $C=1$ and at most 3,000 iterations,
fitted on development rows with non-uncertain references. If fewer than forty
valid rows exist or only one class occurs, predictions fall back to the
valid-development positive rate (0.5 if no valid rows exist).
Predictions are clipped to $[\epsilon,1-\epsilon]$ and $p^{(0)}=1-p^{(1)}$.
Complementarity does not establish calibration.

\subsection{Coefficient Fitting}
\label{app:coefficient-fitting}
Algorithm~\ref{alg:constraints} uses twelve fixed iterations, initialized at
$a_L=2$, $b_L=1$. Each iteration selects constrained development candidates,
computes soft counts from fitted probabilities, and updates the coefficients.
Macro/micro F1 denominators are floored at $10^{-9}$; normalization divides both
$a$ and $b$ by $\max(\operatorname{mean} b,10^{-12})$. Individual coefficients
are then floored at $10^{-9}$. Pure macro and micro scoring correspond to
mixing weights one and zero, respectively; no interpolation is tuned.
The returned coefficients are the saved values used for the last selection,
not the final post-selection update. Candidate-score ties use the earliest index.
The same state model is shared across objectives; within an objective, the same
frozen coefficients are used for C/P/F. Target reconstruction is checked against
the stored selections, with discrepancies rejected (numerical tolerance $10^{-9}$).

\subsection{Retrieval and Generation}
\label{app:realization-implementation}
\paragraph{Retrieval.}
The index contains training finding texts with empty entries removed and no
deduplication. Before filtering, it contains 118,068 MIMIC and 2,536 IU entries.
Training label vectors are binarized as positive versus all other states.
Retrieval minimizes fourteen-label Hamming distance to the target; among tied
entries, it maximizes the dot product of L2-normalized image features (cosine
similarity), then chooses the earliest index. It returns the training findings
text from the think field, with an IU fallback to the gr field.

\paragraph{Generation.}
The main realization model is a fully fine-tuned Qwen3-VL-8B.
A LayerNorm--Linear projector maps CLIP-SFT image features into eight soft
prefix tokens, placed before text embeddings; the language model does not use
an intrinsic visual encoder. Training uses two epochs, batch size four,
learning rate $10^{-5}$ for the language model and $10^{-4}$ for the projector,
weight decay 0.05, maximum training sequence length 384, and DeepSpeed ZeRO-2.
The input lists all fourteen labels in CheXpert order; the renderer maps
$1/0/-1$ to yes/no/uncertain. RQ2 test targets are binary.
The system prompt is:
\begin{quote}\small
You are a radiologist. Write the findings section of a chest X-ray report that
is consistent with the given findings list and the image.
\end{quote}
Main decoding is greedy, with one report per target, at most 160 new tokens,
and left-padded batches. Output postprocessing strips surrounding whitespace.
The structured answer field is copied from the target, rather than generated.
The narrative-extracted finding representation is distinct from the copied
structured answer field.
The alternate-scale tables use the 1.5B image-label realization checkpoint
identified in the implementation notes and are excluded from configuration selection.

\subsection{Statistics, Fidelity, and Performance Transfer}
\label{app:statistics-evaluation}
RQ2 paired cluster bootstrap resamples MIMIC patients and IU studies with
replacement to the original cluster count, using 10,000 draws and seed 12345.
All conditions within a generator share resampling multiplicities. Intervals
are the 2.5th and 97.5th percentiles of paired differences. Fitted parameters
remain frozen. Intervals capture test sampling uncertainty conditional on the
fits, not development-fitting uncertainty, and are uncorrected for multiple
comparisons. No confidence interval for the across-generator mean is implied.
RQ3 is descriptive; absent bootstrap intervals do not imply nonsignificance.
CheXbert and GREEN are accessed through the local labeling/scoring services.

Fidelity compares requested targets $z$ and narrative-extracted findings
$\tilde z$ independently of the reference, over all fourteen labels:
\begin{equation}
\begin{aligned}
\mathrm{Retention}&=\frac{\sum_{i,L}z_{iL}\tilde z_{iL}}
 {\max(\sum_{i,L}z_{iL},1)},\\
\mathrm{Insertion}&=\frac{\sum_{i,L}(1-z_{iL})\tilde z_{iL}}
 {\max(\sum_{i,L}\tilde z_{iL},1)},\\
H&=\frac{1}{nK}\sum_{i,L}\mathbf1[z_{iL}\ne\tilde z_{iL}].
\end{aligned}
\end{equation}
These are pooled test-set statistics, not per-input averages, and do not apply
reference-uncertainty masks. Empty positive denominators yield zero.
By contrast, finding F1 and reference-aware change decomposition apply the
reference-only mask. Hamming provides a total-cell disagreement denominator;
it does not introduce a third kind of binary change beyond deletion/insertion.
For target $t$ and the same-objective constrained baseline $C$,
\begin{align}
L_r(t)&=M_r(t)-M_{\mathrm{dec}}(t),\\
\Delta D(t)&=M_{\mathrm{dec}}(t)-M_{\mathrm{dec}}(C),\\
A_r(t)&=[M_r(t)-M_r(C)]-\Delta D(t).
\end{align}
All performance contrasts use the same reference-based metric and test subset.
Positive $A_r$ makes a contrast more positive; for negative $\Delta D$, this
can mitigate a loss rather than amplify a gain.

\section{Additional Results for Candidate Constraints}\label{app:rq2}
The corresponding results are reported in Table~\ref{tab:app-rq2-r5}.

We report all thirteen configurations per benchmark. C/P/F denote constrained,
partial and free decisions. Partial replaces five labels chosen by training prevalence;
it is a hybrid intervention, not an independently optimized intermediate feasible set.
The label sets below differ from the five-label evaluation subset.

\begin{table}[!htbp]\centering\scriptsize

\setlength{\tabcolsep}{1.5pt}\renewcommand{\arraystretch}{1.0}

\begin{tabular}{@{}ccc@{}}\toprule

\textbf{Dataset} & \textbf{Replacement label} & \textbf{Train prev.}\\ \midrule

MIMIC & Pleural Other & 0.0169 \\

MIMIC & Pneumothorax & 0.0334 \\

MIMIC & Consolidation & 0.0383 \\

MIMIC & Fracture & 0.0387 \\

MIMIC & Lung Lesion & 0.0395 \\

IU & Pneumothorax & 0.0043 \\

IU & Consolidation & 0.0091 \\

IU & Edema & 0.0126 \\

IU & Pleural Other & 0.0138 \\

IU & Pneumonia & 0.0170 \\

\bottomrule\end{tabular}

\caption{Training-defined labels used in partial replacement.}\label{tab:app-rq2-r5}\end{table}

\subsection{Absolute Finding Results}
\label{app:rq2-absolute}
The absolute finding results for constrained, partial, and free targets are reported in Tables~\ref{tab:app-rq2-abs-mimic-macro}, \ref{tab:app-rq2-abs-mimic-micro}, \ref{tab:app-rq2-abs-iu-macro}, \ref{tab:app-rq2-abs-iu-micro}.

\begin{table*}[!tp]
\centering
\begin{minipage}[t]{0.49\textwidth}

\centering\fontsize{7.5}{9.5}\selectfont
\setlength{\tabcolsep}{2pt}\renewcommand{\arraystretch}{1.05}
\begin{tabular}{@{}c|c|cccc|ccc@{}}
\toprule
\textbf{Gen.} & \textbf{T} & \textbf{Ma-14} & \textbf{Mi-14} & \textbf{Ma-5} & \textbf{Mi-5} & \textbf{P} & \textbf{R} & \textbf{Asrt.}\\
\midrule
\addlinespace[1pt]
Q25-3 & C & .4722 & .6465 & .6158 & .6602 & .4811 & .4895 & 2.0077 \\
Q25-3 & P & .4937 & .6303 & .6030 & .6469 & .4799 & .5707 & 2.1872 \\
Q25-3 & F & .5030 & .6311 & .6214 & .6664 & .4667 & .6228 & 2.5007 \\
\addlinespace[1pt]
Q25-1 & C & .4607 & .6318 & .5938 & .6443 & .4791 & .4575 & 1.8904 \\
Q25-1 & P & .4667 & .5816 & .5859 & .6291 & .4540 & .5703 & 2.3429 \\
Q25-1 & F & .4908 & .5911 & .6192 & .6654 & .4425 & .6642 & 2.8575 \\
\addlinespace[1pt]
Q35 & C & .4586 & .5997 & .5829 & .6214 & .4569 & .5305 & 2.4413 \\
Q35 & P & .4561 & .4760 & .5813 & .6243 & .4219 & .6700 & 3.7172 \\
Q35 & F & .4772 & .4960 & .6095 & .6621 & .4404 & .6986 & 3.7682 \\
\addlinespace[1pt]
IV35 & C & .4668 & .6306 & .5933 & .6262 & .4683 & .5168 & 2.1494 \\
IV35 & P & .4857 & .5896 & .5990 & .6330 & .4560 & .6237 & 2.5447 \\
IV35 & F & .4926 & .5957 & .6234 & .6674 & .4454 & .6662 & 2.8122 \\
\addlinespace[1pt]
ESC-RL & C & .4838 & .6436 & .6128 & .6573 & .4982 & .4980 & 2.0091 \\
ESC-RL & P & .4969 & .6313 & .6045 & .6478 & .4918 & .5593 & 2.1453 \\
ESC-RL & F & .5019 & .6327 & .6237 & .6675 & .4760 & .6017 & 2.4323 \\
\addlinespace[1pt]
MPO & C & .4771 & .6399 & .6080 & .6526 & .4796 & .4963 & 2.0161 \\
MPO & P & .4867 & .6228 & .5966 & .6401 & .4794 & .5619 & 2.1878 \\
MPO & F & .4979 & .6246 & .6133 & .6572 & .4705 & .6115 & 2.4804 \\
\addlinespace[1pt]
EditGRPO & C & .4849 & .6474 & .6112 & .6619 & .5094 & .5029 & 2.0412 \\
EditGRPO & P & .4915 & .6333 & .6041 & .6508 & .4975 & .5598 & 2.1816 \\
EditGRPO & F & .5008 & .6352 & .6223 & .6690 & .4872 & .6051 & 2.4399 \\
\addlinespace[1pt]
DPO-G & C & .4657 & .6288 & .5791 & .6347 & .4960 & .4717 & 1.9057 \\
DPO-G & P & .4732 & .5891 & .5784 & .6192 & .4719 & .5645 & 2.2640 \\
DPO-G & F & .4940 & .6066 & .6107 & .6544 & .4603 & .6379 & 2.6927 \\
\addlinespace[1pt]
R2Gen & C & .4626 & .6305 & .5947 & .6416 & .4777 & .4693 & 1.9378 \\
R2Gen & P & .4703 & .5744 & .5913 & .6329 & .4651 & .5902 & 2.4309 \\
R2Gen & F & .4893 & .5857 & .6267 & .6704 & .4522 & .6689 & 2.8715 \\
\addlinespace[1pt]
R2CMN & C & .4735 & .6344 & .6119 & .6553 & .4863 & .4811 & 1.9190 \\
R2CMN & P & .4803 & .6100 & .5982 & .6412 & .4677 & .5571 & 2.1501 \\
R2CMN & F & .4925 & .6101 & .6197 & .6627 & .4519 & .6304 & 2.6439 \\
\addlinespace[1pt]
L15 & C & .5126 & .6547 & .6290 & .6715 & .5180 & .5361 & 2.0454 \\
L15 & P & .5066 & .6437 & .6178 & .6627 & .4983 & .5686 & 2.1432 \\
L15 & F & .5122 & .6459 & .6277 & .6710 & .4868 & .6025 & 2.3506 \\
\addlinespace[1pt]
LN-M & C & .5201 & .6607 & .6347 & .6738 & .5239 & .5436 & 2.0356 \\
LN-M & P & .5235 & .6567 & .6266 & .6681 & .5178 & .5677 & 2.0824 \\
LN-M & F & .5234 & .6546 & .6314 & .6727 & .5030 & .5936 & 2.2654 \\
\addlinespace[1pt]
LN-V & C & .5190 & .6592 & .6295 & .6739 & .5257 & .5366 & 2.0189 \\
LN-V & P & .5160 & .6510 & .6211 & .6672 & .5086 & .5653 & 2.0950 \\
LN-V & F & .5172 & .6482 & .6272 & .6710 & .4930 & .5960 & 2.3184 \\
\bottomrule
\end{tabular}
\caption{MIMIC: macro-oriented absolute target results.}\label{tab:app-rq2-abs-mimic-macro}

\end{minipage}
\hfill
\begin{minipage}[t]{0.49\textwidth}

\centering\fontsize{7.5}{9.5}\selectfont
\setlength{\tabcolsep}{2pt}\renewcommand{\arraystretch}{1.05}
\begin{tabular}{@{}c|c|cccc|ccc@{}}
\toprule
\textbf{Gen.} & \textbf{T} & \textbf{Ma-14} & \textbf{Mi-14} & \textbf{Ma-5} & \textbf{Mi-5} & \textbf{P} & \textbf{R} & \textbf{Asrt.}\\
\midrule
\addlinespace[1pt]
Q25-3 & C & .4382 & .6683 & .5992 & .6752 & .5248 & .4445 & 1.9232 \\
Q25-3 & P & .4602 & .6673 & .6152 & .6729 & .5496 & .4664 & 1.9518 \\
Q25-3 & F & .4625 & .6661 & .6207 & .6772 & .5527 & .4804 & 2.0468 \\
\addlinespace[1pt]
Q25-1 & C & .4073 & .6516 & .5808 & .6620 & .5008 & .4050 & 1.8317 \\
Q25-1 & P & .4391 & .6514 & .6031 & .6603 & .4978 & .4365 & 1.8694 \\
Q25-1 & F & .4659 & .6638 & .6212 & .6772 & .5610 & .4806 & 2.0517 \\
\addlinespace[1pt]
Q35 & C & .4338 & .6108 & .5759 & .6273 & .4459 & .4732 & 2.2297 \\
Q35 & P & .4475 & .6121 & .5826 & .6339 & .4677 & .4767 & 2.2186 \\
Q35 & F & .4709 & .6595 & .6148 & .6735 & .5186 & .4970 & 2.1739 \\
\addlinespace[1pt]
IV35 & C & .4269 & .6473 & .6016 & .6466 & .5092 & .4442 & 1.9574 \\
IV35 & P & .4653 & .6518 & .6076 & .6537 & .5725 & .4607 & 1.9462 \\
IV35 & F & .4691 & .6619 & .6280 & .6796 & .5648 & .4877 & 2.0943 \\
\addlinespace[1pt]
ESC-RL & C & .4520 & .6684 & .6122 & .6780 & .5706 & .4513 & 1.9225 \\
ESC-RL & P & .4676 & .6670 & .6233 & .6761 & .5706 & .4696 & 1.9490 \\
ESC-RL & F & .4741 & .6681 & .6274 & .6790 & .5637 & .4871 & 2.0538 \\
\addlinespace[1pt]
MPO & C & .4333 & .6597 & .5814 & .6612 & .5070 & .4371 & 1.9057 \\
MPO & P & .4573 & .6598 & .6097 & .6618 & .5262 & .4607 & 1.9309 \\
MPO & F & .4652 & .6633 & .6204 & .6728 & .5528 & .4804 & 2.0440 \\
\addlinespace[1pt]
EditGRPO & C & .4478 & .6677 & .5978 & .6779 & .5718 & .4481 & 1.9330 \\
EditGRPO & P & .4698 & .6680 & .6227 & .6777 & .5807 & .4707 & 1.9560 \\
EditGRPO & F & .4754 & .6674 & .6269 & .6784 & .5898 & .4870 & 2.0454 \\
\addlinespace[1pt]
DPO-G & C & .4390 & .6594 & .5759 & .6586 & .6351 & .4251 & 1.8122 \\
DPO-G & P & .4614 & .6586 & .6095 & .6585 & .5826 & .4511 & 1.8464 \\
DPO-G & F & .4735 & .6674 & .6279 & .6810 & .5826 & .4882 & 2.0684 \\
\addlinespace[1pt]
R2Gen & C & .4300 & .6563 & .5808 & .6623 & .6487 & .4248 & 1.8589 \\
R2Gen & P & .4492 & .6562 & .6114 & .6629 & .5759 & .4468 & 1.8855 \\
R2Gen & F & .4556 & .6587 & .6263 & .6786 & .5562 & .4709 & 2.0538 \\
\addlinespace[1pt]
R2CMN & C & .4267 & .6576 & .5875 & .6661 & .5606 & .4249 & 1.8436 \\
R2CMN & P & .4524 & .6581 & .6170 & .6661 & .5515 & .4487 & 1.8680 \\
R2CMN & F & .4668 & .6656 & .6309 & .6798 & .5600 & .4815 & 2.0412 \\
\addlinespace[1pt]
L15 & C & .4586 & .6698 & .6066 & .6783 & .5448 & .4589 & 1.9211 \\
L15 & P & .4633 & .6685 & .6244 & .6785 & .5513 & .4654 & 1.9344 \\
L15 & F & .4718 & .6728 & .6323 & .6832 & .5694 & .4811 & 2.0042 \\
\addlinespace[1pt]
LN-M & C & .4697 & .6706 & .6008 & .6725 & .5922 & .4632 & 1.9106 \\
LN-M & P & .4797 & .6709 & .6235 & .6730 & .5940 & .4761 & 1.9232 \\
LN-M & F & .4813 & .6704 & .6307 & .6784 & .5881 & .4854 & 1.9888 \\
\addlinespace[1pt]
LN-V & C & .4661 & .6702 & .6013 & .6743 & .5953 & .4607 & 1.9043 \\
LN-V & P & .4722 & .6701 & .6196 & .6751 & .5870 & .4678 & 1.9134 \\
LN-V & F & .4730 & .6707 & .6206 & .6767 & .5792 & .4759 & 1.9784 \\
\bottomrule
\end{tabular}
\caption{MIMIC: micro-oriented absolute target results.}\label{tab:app-rq2-abs-mimic-micro}

\end{minipage}
\end{table*}

\begin{table*}[!tp]
\centering
\begin{minipage}[t]{0.49\textwidth}

\centering\fontsize{7.5}{9.5}\selectfont
\setlength{\tabcolsep}{2pt}\renewcommand{\arraystretch}{1.05}
\begin{tabular}{@{}c|c|cccc|ccc@{}}
\toprule
\textbf{Gen.} & \textbf{T} & \textbf{Ma-14} & \textbf{Mi-14} & \textbf{Ma-5} & \textbf{Mi-5} & \textbf{P} & \textbf{R} & \textbf{Asrt.}\\
\midrule
\addlinespace[1pt]
Q35 & C & .2530 & .6110 & .2581 & .3779 & .2835 & .2837 & 1.3812 \\
Q35 & P & .2537 & .4153 & .2625 & .1411 & .2754 & .5105 & 2.6861 \\
Q35 & F & .2174 & .2822 & .2271 & .1585 & .1610 & .7127 & 5.5269 \\
\addlinespace[1pt]
IV35 & C & .2656 & .6114 & .3044 & .4071 & .3254 & .3645 & 1.3722 \\
IV35 & P & .2248 & .4338 & .2446 & .1546 & .2672 & .4520 & 2.4417 \\
IV35 & F & .2172 & .2892 & .2485 & .1600 & .1817 & .6428 & 4.9013 \\
\addlinespace[1pt]
Q25-3 & C & .1459 & .3446 & .1144 & .1806 & .1654 & .1673 & 1.2018 \\
Q25-3 & P & .1353 & .1471 & .1200 & .0393 & .1544 & .4977 & 4.8296 \\
Q25-3 & F & .1657 & .1787 & .1639 & .0936 & .1205 & .8867 & 9.6009 \\
\addlinespace[1pt]
Q25-1 & C & .0582 & .1807 & .0000 & .0000 & .1004 & .0839 & 1.0695 \\
Q25-1 & P & .0740 & .0874 & .0123 & .0198 & .1090 & .4143 & 4.2242 \\
Q25-1 & F & .1513 & .1746 & .1191 & .1173 & .1047 & .9125 & 10.3901 \\
\addlinespace[1pt]
ESC-RL & C & .1221 & .3057 & .1058 & .1600 & .1554 & .1423 & 1.1592 \\
ESC-RL & P & .1206 & .1588 & .1141 & .0522 & .1499 & .4190 & 3.5740 \\
ESC-RL & F & .1736 & .2037 & .1690 & .1181 & .1272 & .8271 & 8.2422 \\
\addlinespace[1pt]
MPO & C & .1603 & .4537 & .2011 & .2454 & .1803 & .1627 & 1.1883 \\
MPO & P & .1354 & .1711 & .1403 & .0541 & .1562 & .4823 & 5.4103 \\
MPO & F & .1613 & .1736 & .1583 & .1092 & .1160 & .8875 & 9.9686 \\
\addlinespace[1pt]
R2Gen & C & .0828 & .5370 & .0547 & .1576 & .1072 & .0918 & 1.0067 \\
R2Gen & P & .0889 & .1748 & .0601 & .0370 & .1103 & .4489 & 5.8901 \\
R2Gen & F & .1467 & .1590 & .1482 & .0933 & .1050 & .9153 & 11.6031 \\
\addlinespace[1pt]
EditGRPO & C & .0877 & .5837 & .0496 & .0702 & .1281 & .0875 & .9215 \\
EditGRPO & P & .0961 & .2006 & .0550 & .0202 & .1324 & .4179 & 5.1682 \\
EditGRPO & F & .1420 & .1614 & .1252 & .0899 & .0988 & .9275 & 11.6906 \\
\addlinespace[1pt]
DPO-G & C & .1081 & .5397 & .0754 & .1186 & .1904 & .1026 & 1.0045 \\
DPO-G & P & .1169 & .2159 & .0828 & .0374 & .1949 & .4330 & 4.5157 \\
DPO-G & F & .1570 & .1840 & .1575 & .1177 & .1104 & .8878 & 9.6816 \\
\addlinespace[1pt]
R2CMN & C & .1019 & .3375 & .1039 & .1771 & .1188 & .1053 & 1.2377 \\
R2CMN & P & .1082 & .1309 & .1100 & .0541 & .1220 & .4625 & 5.6009 \\
R2CMN & F & .1562 & .1632 & .1491 & .1106 & .1134 & .9140 & 10.7534 \\
\addlinespace[1pt]
L15 & C & .0521 & .2236 & .0000 & .0000 & .0745 & .0779 & 1.0112 \\
L15 & P & .0582 & .0795 & .0054 & .0122 & .0776 & .4350 & 5.9327 \\
L15 & F & .1349 & .1469 & .0877 & .0883 & .0960 & .9666 & 12.7422 \\
\addlinespace[1pt]
LN-M & C & .0849 & .5200 & .0077 & .0182 & .0896 & .1034 & 1.1883 \\
LN-M & P & .0920 & .1895 & .0131 & .0142 & .0932 & .4605 & 5.6525 \\
LN-M & F & .1413 & .1632 & .1002 & .0933 & .1003 & .8891 & 10.9507 \\
\addlinespace[1pt]
LN-V & C & .0227 & .0553 & .0291 & .0741 & .1175 & .0771 & .9955 \\
LN-V & P & .0288 & .0287 & .0345 & .0203 & .1206 & .4342 & 5.9170 \\
LN-V & F & .1517 & .1562 & .1340 & .1026 & .1076 & .9369 & 11.6592 \\
\bottomrule
\end{tabular}
\caption{IU: macro-oriented absolute target results.}\label{tab:app-rq2-abs-iu-macro}

\end{minipage}
\hfill
\begin{minipage}[t]{0.49\textwidth}

\centering\fontsize{7.5}{9.5}\selectfont
\setlength{\tabcolsep}{2pt}\renewcommand{\arraystretch}{1.05}
\begin{tabular}{@{}c|c|cccc|ccc@{}}
\toprule
\textbf{Gen.} & \textbf{T} & \textbf{Ma-14} & \textbf{Mi-14} & \textbf{Ma-5} & \textbf{Mi-5} & \textbf{P} & \textbf{R} & \textbf{Asrt.}\\
\midrule
\addlinespace[1pt]
Q35 & C & .2499 & .6386 & .3146 & .3151 & .4004 & .2590 & 1.1188 \\
Q35 & P & .1864 & .6365 & .1813 & .3056 & .2933 & .1787 & 1.1121 \\
Q35 & F & .1711 & .6427 & .1597 & .3494 & .1629 & .1841 & 1.1951 \\
\addlinespace[1pt]
IV35 & C & .2479 & .6567 & .3641 & .4217 & .3114 & .2714 & 1.1256 \\
IV35 & P & .2003 & .6566 & .2307 & .4172 & .2757 & .2000 & 1.1188 \\
IV35 & F & .1731 & .6534 & .1859 & .3902 & .1834 & .1765 & 1.1368 \\
\addlinespace[1pt]
Q25-3 & C & .1207 & .6168 & .0877 & .1774 & .2022 & .1145 & 1.0448 \\
Q25-3 & P & .1207 & .6181 & .0877 & .1789 & .2022 & .1145 & 1.0404 \\
Q25-3 & F & .1386 & .6192 & .1575 & .3563 & .1229 & .1604 & 1.2466 \\
\addlinespace[1pt]
Q25-1 & C & .0665 & .6019 & .0000 & .0000 & .0710 & .0765 & .9641 \\
Q25-1 & P & .0665 & .6019 & .0000 & .0000 & .0710 & .0765 & .9641 \\
Q25-1 & F & .1369 & .6135 & .1520 & .3464 & .1177 & .1647 & 1.3049 \\
\addlinespace[1pt]
ESC-RL & C & .1255 & .6245 & .1308 & .2615 & .1656 & .1204 & 1.0471 \\
ESC-RL & P & .1255 & .6251 & .1308 & .2615 & .1656 & .1204 & 1.0448 \\
ESC-RL & F & .1392 & .6242 & .1566 & .3543 & .1235 & .1609 & 1.2422 \\
\addlinespace[1pt]
MPO & C & .1199 & .6203 & .0778 & .1296 & .2660 & .1109 & 1.0112 \\
MPO & P & .1199 & .6209 & .0778 & .1308 & .2660 & .1109 & 1.0090 \\
MPO & F & .1372 & .6177 & .1556 & .3523 & .1203 & .1609 & 1.2668 \\
\addlinespace[1pt]
R2Gen & C & .0773 & .6034 & .0529 & .1525 & .1501 & .0846 & .9888 \\
R2Gen & P & .0773 & .6034 & .0529 & .1525 & .1501 & .0846 & .9888 \\
R2Gen & F & .1369 & .6135 & .1520 & .3464 & .1177 & .1647 & 1.3049 \\
\addlinespace[1pt]
EditGRPO & C & .0815 & .6056 & .0572 & .0769 & .1826 & .0843 & .9664 \\
EditGRPO & P & .0815 & .6056 & .0572 & .0769 & .1826 & .0843 & .9664 \\
EditGRPO & F & .1364 & .6113 & .1511 & .3444 & .1169 & .1647 & 1.3139 \\
\addlinespace[1pt]
DPO-G & C & .0799 & .6108 & .0400 & .1091 & .1239 & .0842 & .9709 \\
DPO-G & P & .0799 & .6108 & .0400 & .1091 & .1239 & .0842 & .9709 \\
DPO-G & F & .1366 & .6138 & .1547 & .3503 & .1185 & .1622 & 1.2892 \\
\addlinespace[1pt]
R2CMN & C & .1078 & .6152 & .0867 & .1983 & .1409 & .1063 & 1.0359 \\
R2CMN & P & .1078 & .6152 & .0867 & .1983 & .1409 & .1063 & 1.0359 \\
R2CMN & F & .1374 & .6161 & .1556 & .3523 & .1198 & .1622 & 1.2803 \\
\addlinespace[1pt]
L15 & C & .0808 & .6108 & .0000 & .0000 & .0748 & .0911 & 1.0000 \\
L15 & P & .0808 & .6108 & .0000 & .0000 & .0748 & .0911 & 1.0000 \\
L15 & F & .1366 & .6138 & .1547 & .3503 & .1185 & .1622 & 1.2892 \\
\addlinespace[1pt]
LN-M & C & .0788 & .6044 & .0000 & .0000 & .0798 & .0870 & 1.0224 \\
LN-M & P & .0788 & .6050 & .0000 & .0000 & .0798 & .0870 & 1.0202 \\
LN-M & F & .1368 & .6140 & .1533 & .3483 & .1182 & .1634 & 1.2960 \\
\addlinespace[1pt]
LN-V & C & .0866 & .6129 & .0157 & .0400 & .1231 & .0938 & 1.0000 \\
LN-V & P & .0866 & .6129 & .0157 & .0400 & .1231 & .0938 & 1.0000 \\
LN-V & F & .1369 & .6144 & .1556 & .3523 & .1191 & .1622 & 1.2870 \\
\bottomrule
\end{tabular}
\caption{IU: micro-oriented absolute target results.}\label{tab:app-rq2-abs-iu-micro}

\end{minipage}
\end{table*}

\subsection{Paired Contrasts and Interactions}
\label{app:rq2-contrasts}
The corresponding results are reported in Tables~\ref{tab:app-rq2-diff-mimic-macro-partial}, \ref{tab:app-rq2-diff-mimic-macro-free}, \ref{tab:app-rq2-diff-mimic-micro-partial}, \ref{tab:app-rq2-diff-mimic-micro-free}, \ref{tab:app-rq2-interaction-mimic-partial}, \ref{tab:app-rq2-interaction-mimic-free}, \ref{tab:app-rq2-diff-iu-macro-partial}, \ref{tab:app-rq2-diff-iu-macro-free}, \ref{tab:app-rq2-diff-iu-micro-partial}, \ref{tab:app-rq2-diff-iu-micro-free}, \ref{tab:app-rq2-interaction-iu-partial}, \ref{tab:app-rq2-interaction-iu-free}.

The contrasts and interactions below are computed within each generator.
Cells contain the observed difference and its uncorrected 95\% paired cluster-bootstrap
interval. An interaction subtracts the micro-oriented relaxation effect from the
macro-oriented effect on the same evaluation metric. Intervals are taken directly
from the RQ2 export and are not formed by subtracting marginal interval endpoints.

\begin{table*}[!htbp]\centering\scriptsize

\setlength{\tabcolsep}{3pt}\renewcommand{\arraystretch}{1.0}

\begin{tabular}{@{}ccccc@{}}\toprule

\textbf{Generator} & \textbf{Ma-14} & \textbf{Mi-14} & \textbf{Ma-5} & \textbf{Mi-5}\\ \midrule

Q25-3 & +.0215\,{\fontsize{7.5}{9}\selectfont [+.0046, +.0384]} & -.0162\,{\fontsize{7.5}{9}\selectfont [-.0217, -.0101]} & -.0127\,{\fontsize{7.5}{9}\selectfont [-.0291, +.0043]} & -.0133\,{\fontsize{7.5}{9}\selectfont [-.0200, -.0064]} \\

Q25-1 & +.0060\,{\fontsize{7.5}{9}\selectfont [-.0126, +.0253]} & -.0502\,{\fontsize{7.5}{9}\selectfont [-.0585, -.0419]} & -.0079\,{\fontsize{7.5}{9}\selectfont [-.0278, +.0116]} & -.0152\,{\fontsize{7.5}{9}\selectfont [-.0229, -.0078]} \\

Q35 & -.0025\,{\fontsize{7.5}{9}\selectfont [-.0184, +.0130]} & -.1237\,{\fontsize{7.5}{9}\selectfont [-.1328, -.1150]} & -.0016\,{\fontsize{7.5}{9}\selectfont [-.0114, +.0068]} & +.0029\,{\fontsize{7.5}{9}\selectfont [-.0004, +.0062]} \\

IV35 & +.0188\,{\fontsize{7.5}{9}\selectfont [+.0004, +.0392]} & -.0410\,{\fontsize{7.5}{9}\selectfont [-.0492, -.0331]} & +.0058\,{\fontsize{7.5}{9}\selectfont [-.0052, +.0160]} & +.0068\,{\fontsize{7.5}{9}\selectfont [+.0011, +.0124]} \\

ESC-RL & +.0131\,{\fontsize{7.5}{9}\selectfont [+.0002, +.0263]} & -.0123\,{\fontsize{7.5}{9}\selectfont [-.0165, -.0078]} & -.0084\,{\fontsize{7.5}{9}\selectfont [-.0234, +.0068]} & -.0095\,{\fontsize{7.5}{9}\selectfont [-.0150, -.0040]} \\

MPO & +.0097\,{\fontsize{7.5}{9}\selectfont [-.0028, +.0228]} & -.0171\,{\fontsize{7.5}{9}\selectfont [-.0216, -.0125]} & -.0114\,{\fontsize{7.5}{9}\selectfont [-.0246, +.0021]} & -.0125\,{\fontsize{7.5}{9}\selectfont [-.0185, -.0067]} \\

EditGRPO & +.0067\,{\fontsize{7.5}{9}\selectfont [-.0062, +.0195]} & -.0141\,{\fontsize{7.5}{9}\selectfont [-.0184, -.0094]} & -.0071\,{\fontsize{7.5}{9}\selectfont [-.0215, +.0079]} & -.0111\,{\fontsize{7.5}{9}\selectfont [-.0177, -.0049]} \\
\bottomrule\end{tabular}

\caption{MIMIC: macro-oriented partial-minus-constrained target contrasts.}\label{tab:app-rq2-diff-mimic-macro-partial}\end{table*}

\begin{table*}[!htbp]\centering\scriptsize

\setlength{\tabcolsep}{3pt}\renewcommand{\arraystretch}{1.0}

\begin{tabular}{@{}ccccc@{}}\toprule

\textbf{Generator} & \textbf{Ma-14} & \textbf{Mi-14} & \textbf{Ma-5} & \textbf{Mi-5}\\ \midrule

Q25-3 & +.0307\,{\fontsize{7.5}{9}\selectfont [+.0120, +.0504]} & -.0154\,{\fontsize{7.5}{9}\selectfont [-.0255, -.0051]} & +.0056\,{\fontsize{7.5}{9}\selectfont [-.0144, +.0257]} & +.0062\,{\fontsize{7.5}{9}\selectfont [-.0068, +.0193]} \\

Q25-1 & +.0301\,{\fontsize{7.5}{9}\selectfont [+.0087, +.0519]} & -.0407\,{\fontsize{7.5}{9}\selectfont [-.0554, -.0269]} & +.0254\,{\fontsize{7.5}{9}\selectfont [+.0005, +.0512]} & +.0211\,{\fontsize{7.5}{9}\selectfont [+.0027, +.0406]} \\

Q35 & +.0186\,{\fontsize{7.5}{9}\selectfont [+.0004, +.0370]} & -.1037\,{\fontsize{7.5}{9}\selectfont [-.1172, -.0902]} & +.0266\,{\fontsize{7.5}{9}\selectfont [+.0123, +.0420]} & +.0408\,{\fontsize{7.5}{9}\selectfont [+.0239, +.0602]} \\

IV35 & +.0257\,{\fontsize{7.5}{9}\selectfont [+.0056, +.0490]} & -.0350\,{\fontsize{7.5}{9}\selectfont [-.0474, -.0227]} & +.0301\,{\fontsize{7.5}{9}\selectfont [+.0147, +.0473]} & +.0413\,{\fontsize{7.5}{9}\selectfont [+.0261, +.0587]} \\

ESC-RL & +.0181\,{\fontsize{7.5}{9}\selectfont [+.0032, +.0340]} & -.0109\,{\fontsize{7.5}{9}\selectfont [-.0198, -.0017]} & +.0109\,{\fontsize{7.5}{9}\selectfont [-.0090, +.0309]} & +.0102\,{\fontsize{7.5}{9}\selectfont [-.0029, +.0240]} \\

MPO & +.0209\,{\fontsize{7.5}{9}\selectfont [+.0044, +.0377]} & -.0153\,{\fontsize{7.5}{9}\selectfont [-.0248, -.0058]} & +.0053\,{\fontsize{7.5}{9}\selectfont [-.0124, +.0243]} & +.0046\,{\fontsize{7.5}{9}\selectfont [-.0090, +.0180]} \\

EditGRPO & +.0160\,{\fontsize{7.5}{9}\selectfont [+.0016, +.0312]} & -.0122\,{\fontsize{7.5}{9}\selectfont [-.0193, -.0046]} & +.0111\,{\fontsize{7.5}{9}\selectfont [-.0090, +.0306]} & +.0072\,{\fontsize{7.5}{9}\selectfont [-.0066, +.0203]} \\
\bottomrule\end{tabular}

\caption{MIMIC: macro-oriented free-minus-constrained target contrasts.}\label{tab:app-rq2-diff-mimic-macro-free}\end{table*}

\begin{table*}[!htbp]\centering\scriptsize

\setlength{\tabcolsep}{3pt}\renewcommand{\arraystretch}{1.0}

\begin{tabular}{@{}ccccc@{}}\toprule

\textbf{Generator} & \textbf{Ma-14} & \textbf{Mi-14} & \textbf{Ma-5} & \textbf{Mi-5}\\ \midrule

Q25-3 & +.0220\,{\fontsize{7.5}{9}\selectfont [+.0064, +.0396]} & -.0010\,{\fontsize{7.5}{9}\selectfont [-.0031, +.0013]} & +.0160\,{\fontsize{7.5}{9}\selectfont [-.0083, +.0429]} & -.0024\,{\fontsize{7.5}{9}\selectfont [-.0065, +.0017]} \\

Q25-1 & +.0317\,{\fontsize{7.5}{9}\selectfont [+.0102, +.0520]} & -.0002\,{\fontsize{7.5}{9}\selectfont [-.0027, +.0023]} & +.0222\,{\fontsize{7.5}{9}\selectfont [-.0068, +.0476]} & -.0017\,{\fontsize{7.5}{9}\selectfont [-.0058, +.0019]} \\

Q35 & +.0137\,{\fontsize{7.5}{9}\selectfont [-.0079, +.0331]} & +.0014\,{\fontsize{7.5}{9}\selectfont [-.0017, +.0039]} & +.0067\,{\fontsize{7.5}{9}\selectfont [-.0098, +.0199]} & +.0066\,{\fontsize{7.5}{9}\selectfont [+.0024, +.0102]} \\

IV35 & +.0383\,{\fontsize{7.5}{9}\selectfont [+.0155, +.0589]} & +.0045\,{\fontsize{7.5}{9}\selectfont [+.0018, +.0074]} & +.0060\,{\fontsize{7.5}{9}\selectfont [-.0155, +.0234]} & +.0070\,{\fontsize{7.5}{9}\selectfont [+.0019, +.0118]} \\

ESC-RL & +.0156\,{\fontsize{7.5}{9}\selectfont [-.0029, +.0327]} & -.0014\,{\fontsize{7.5}{9}\selectfont [-.0036, +.0007]} & +.0111\,{\fontsize{7.5}{9}\selectfont [-.0142, +.0361]} & -.0018\,{\fontsize{7.5}{9}\selectfont [-.0054, +.0016]} \\

MPO & +.0240\,{\fontsize{7.5}{9}\selectfont [+.0095, +.0389]} & .0000\,{\fontsize{7.5}{9}\selectfont [-.0018, +.0019]} & +.0283\,{\fontsize{7.5}{9}\selectfont [+.0085, +.0475]} & +.0006\,{\fontsize{7.5}{9}\selectfont [-.0024, +.0035]} \\

EditGRPO & +.0220\,{\fontsize{7.5}{9}\selectfont [+.0056, +.0388]} & +.0004\,{\fontsize{7.5}{9}\selectfont [-.0018, +.0025]} & +.0248\,{\fontsize{7.5}{9}\selectfont [-.0042, +.0543]} & -.0001\,{\fontsize{7.5}{9}\selectfont [-.0046, +.0040]} \\
\bottomrule\end{tabular}

\caption{MIMIC: micro-oriented partial-minus-constrained target contrasts.}\label{tab:app-rq2-diff-mimic-micro-partial}\end{table*}

\begin{table*}[!htbp]\centering\scriptsize

\setlength{\tabcolsep}{3pt}\renewcommand{\arraystretch}{1.0}

\begin{tabular}{@{}ccccc@{}}\toprule

\textbf{Generator} & \textbf{Ma-14} & \textbf{Mi-14} & \textbf{Ma-5} & \textbf{Mi-5}\\ \midrule

Q25-3 & +.0243\,{\fontsize{7.5}{9}\selectfont [+.0065, +.0439]} & -.0022\,{\fontsize{7.5}{9}\selectfont [-.0079, +.0034]} & +.0215\,{\fontsize{7.5}{9}\selectfont [-.0069, +.0504]} & +.0020\,{\fontsize{7.5}{9}\selectfont [-.0075, +.0112]} \\

Q25-1 & +.0586\,{\fontsize{7.5}{9}\selectfont [+.0335, +.0818]} & +.0121\,{\fontsize{7.5}{9}\selectfont [+.0045, +.0199]} & +.0404\,{\fontsize{7.5}{9}\selectfont [+.0086, +.0685]} & +.0153\,{\fontsize{7.5}{9}\selectfont [+.0028, +.0284]} \\

Q35 & +.0371\,{\fontsize{7.5}{9}\selectfont [+.0114, +.0598]} & +.0487\,{\fontsize{7.5}{9}\selectfont [+.0372, +.0603]} & +.0388\,{\fontsize{7.5}{9}\selectfont [+.0191, +.0586]} & +.0461\,{\fontsize{7.5}{9}\selectfont [+.0297, +.0646]} \\

IV35 & +.0422\,{\fontsize{7.5}{9}\selectfont [+.0188, +.0646]} & +.0146\,{\fontsize{7.5}{9}\selectfont [+.0068, +.0223]} & +.0263\,{\fontsize{7.5}{9}\selectfont [+.0020, +.0482]} & +.0330\,{\fontsize{7.5}{9}\selectfont [+.0184, +.0477]} \\

ESC-RL & +.0222\,{\fontsize{7.5}{9}\selectfont [+.0019, +.0414]} & -.0003\,{\fontsize{7.5}{9}\selectfont [-.0054, +.0050]} & +.0152\,{\fontsize{7.5}{9}\selectfont [-.0114, +.0415]} & +.0010\,{\fontsize{7.5}{9}\selectfont [-.0077, +.0093]} \\

MPO & +.0319\,{\fontsize{7.5}{9}\selectfont [+.0146, +.0485]} & +.0035\,{\fontsize{7.5}{9}\selectfont [-.0029, +.0102]} & +.0390\,{\fontsize{7.5}{9}\selectfont [+.0155, +.0621]} & +.0116\,{\fontsize{7.5}{9}\selectfont [+.0023, +.0216]} \\

EditGRPO & +.0276\,{\fontsize{7.5}{9}\selectfont [+.0074, +.0477]} & -.0003\,{\fontsize{7.5}{9}\selectfont [-.0069, +.0064]} & +.0291\,{\fontsize{7.5}{9}\selectfont [-.0058, +.0647]} & +.0006\,{\fontsize{7.5}{9}\selectfont [-.0104, +.0112]} \\
\bottomrule\end{tabular}

\caption{MIMIC: micro-oriented free-minus-constrained target contrasts.}\label{tab:app-rq2-diff-mimic-micro-free}\end{table*}

\begin{table*}[!htbp]\centering\scriptsize

\setlength{\tabcolsep}{3pt}\renewcommand{\arraystretch}{1.0}

\begin{tabular}{@{}ccccc@{}}\toprule

\textbf{Generator} & \textbf{Ma-14} & \textbf{Mi-14} & \textbf{Ma-5} & \textbf{Mi-5}\\ \midrule

Q25-3 & -.0005\,{\fontsize{7.5}{9}\selectfont [-.0279, +.0241]} & -.0152\,{\fontsize{7.5}{9}\selectfont [-.0209, -.0091]} & -.0287\,{\fontsize{7.5}{9}\selectfont [-.0634, +.0033]} & -.0109\,{\fontsize{7.5}{9}\selectfont [-.0185, -.0029]} \\

Q25-1 & -.0258\,{\fontsize{7.5}{9}\selectfont [-.0573, +.0072]} & -.0500\,{\fontsize{7.5}{9}\selectfont [-.0592, -.0409]} & -.0301\,{\fontsize{7.5}{9}\selectfont [-.0533, -.0039]} & -.0135\,{\fontsize{7.5}{9}\selectfont [-.0206, -.0063]} \\

Q35 & -.0161\,{\fontsize{7.5}{9}\selectfont [-.0379, +.0078]} & -.1250\,{\fontsize{7.5}{9}\selectfont [-.1339, -.1165]} & -.0082\,{\fontsize{7.5}{9}\selectfont [-.0209, +.0075]} & -.0036\,{\fontsize{7.5}{9}\selectfont [-.0073, +.0007]} \\

IV35 & -.0195\,{\fontsize{7.5}{9}\selectfont [-.0433, +.0073]} & -.0456\,{\fontsize{7.5}{9}\selectfont [-.0543, -.0369]} & -.0002\,{\fontsize{7.5}{9}\selectfont [-.0197, +.0220]} & -.0002\,{\fontsize{7.5}{9}\selectfont [-.0069, +.0066]} \\

ESC-RL & -.0026\,{\fontsize{7.5}{9}\selectfont [-.0211, +.0180]} & -.0109\,{\fontsize{7.5}{9}\selectfont [-.0153, -.0062]} & -.0195\,{\fontsize{7.5}{9}\selectfont [-.0466, +.0097]} & -.0077\,{\fontsize{7.5}{9}\selectfont [-.0136, -.0010]} \\

MPO & -.0143\,{\fontsize{7.5}{9}\selectfont [-.0338, +.0050]} & -.0171\,{\fontsize{7.5}{9}\selectfont [-.0216, -.0126]} & -.0397\,{\fontsize{7.5}{9}\selectfont [-.0608, -.0192]} & -.0131\,{\fontsize{7.5}{9}\selectfont [-.0189, -.0073]} \\

EditGRPO & -.0153\,{\fontsize{7.5}{9}\selectfont [-.0354, +.0041]} & -.0144\,{\fontsize{7.5}{9}\selectfont [-.0192, -.0096]} & -.0319\,{\fontsize{7.5}{9}\selectfont [-.0638, -.0002]} & -.0110\,{\fontsize{7.5}{9}\selectfont [-.0183, -.0038]} \\
\bottomrule\end{tabular}

\caption{MIMIC: same-metric objective-by-partial-relaxation interactions.}\label{tab:app-rq2-interaction-mimic-partial}\end{table*}

\begin{table*}[!htbp]\centering\scriptsize

\setlength{\tabcolsep}{3pt}\renewcommand{\arraystretch}{1.0}

\begin{tabular}{@{}ccccc@{}}\toprule

\textbf{Generator} & \textbf{Ma-14} & \textbf{Mi-14} & \textbf{Ma-5} & \textbf{Mi-5}\\ \midrule

Q25-3 & +.0065\,{\fontsize{7.5}{9}\selectfont [-.0223, +.0331]} & -.0131\,{\fontsize{7.5}{9}\selectfont [-.0231, -.0030]} & -.0159\,{\fontsize{7.5}{9}\selectfont [-.0519, +.0182]} & +.0043\,{\fontsize{7.5}{9}\selectfont [-.0080, +.0177]} \\

Q25-1 & -.0285\,{\fontsize{7.5}{9}\selectfont [-.0650, +.0100]} & -.0528\,{\fontsize{7.5}{9}\selectfont [-.0679, -.0387]} & -.0149\,{\fontsize{7.5}{9}\selectfont [-.0403, +.0141]} & +.0058\,{\fontsize{7.5}{9}\selectfont [-.0088, +.0218]} \\

Q35 & -.0185\,{\fontsize{7.5}{9}\selectfont [-.0430, +.0093]} & -.1524\,{\fontsize{7.5}{9}\selectfont [-.1634, -.1417]} & -.0122\,{\fontsize{7.5}{9}\selectfont [-.0287, +.0057]} & -.0054\,{\fontsize{7.5}{9}\selectfont [-.0167, +.0062]} \\

IV35 & -.0164\,{\fontsize{7.5}{9}\selectfont [-.0426, +.0130]} & -.0495\,{\fontsize{7.5}{9}\selectfont [-.0617, -.0377]} & +.0038\,{\fontsize{7.5}{9}\selectfont [-.0186, +.0293]} & +.0083\,{\fontsize{7.5}{9}\selectfont [-.0038, +.0212]} \\

ESC-RL & -.0041\,{\fontsize{7.5}{9}\selectfont [-.0244, +.0188]} & -.0106\,{\fontsize{7.5}{9}\selectfont [-.0192, -.0018]} & -.0044\,{\fontsize{7.5}{9}\selectfont [-.0309, +.0258]} & +.0092\,{\fontsize{7.5}{9}\selectfont [-.0035, +.0227]} \\

MPO & -.0110\,{\fontsize{7.5}{9}\selectfont [-.0340, +.0132]} & -.0188\,{\fontsize{7.5}{9}\selectfont [-.0279, -.0099]} & -.0336\,{\fontsize{7.5}{9}\selectfont [-.0586, -.0083]} & -.0070\,{\fontsize{7.5}{9}\selectfont [-.0205, +.0062]} \\

EditGRPO & -.0117\,{\fontsize{7.5}{9}\selectfont [-.0344, +.0125]} & -.0119\,{\fontsize{7.5}{9}\selectfont [-.0201, -.0031]} & -.0180\,{\fontsize{7.5}{9}\selectfont [-.0536, +.0184]} & +.0066\,{\fontsize{7.5}{9}\selectfont [-.0069, +.0210]} \\
\bottomrule\end{tabular}

\caption{MIMIC: same-metric objective-by-free-relaxation interactions.}\label{tab:app-rq2-interaction-mimic-free}\end{table*}

\begin{table*}[!htbp]\centering\scriptsize

\setlength{\tabcolsep}{3pt}\renewcommand{\arraystretch}{1.0}

\begin{tabular}{@{}ccccc@{}}\toprule

\textbf{Generator} & \textbf{Ma-14} & \textbf{Mi-14} & \textbf{Ma-5} & \textbf{Mi-5}\\ \midrule

Q35 & +.0007\,{\fontsize{7.5}{9}\selectfont [-.0312, +.0332]} & -.1956\,{\fontsize{7.5}{9}\selectfont [-.2169, -.1750]} & +.0045\,{\fontsize{7.5}{9}\selectfont [-.0604, +.0651]} & -.2368\,{\fontsize{7.5}{9}\selectfont [-.2921, -.1815]} \\

IV35 & -.0408\,{\fontsize{7.5}{9}\selectfont [-.0810, +.0004]} & -.1776\,{\fontsize{7.5}{9}\selectfont [-.1942, -.1612]} & -.0598\,{\fontsize{7.5}{9}\selectfont [-.1202, +.0029]} & -.2525\,{\fontsize{7.5}{9}\selectfont [-.3035, -.1984]} \\

Q25-3 & -.0106\,{\fontsize{7.5}{9}\selectfont [-.0369, +.0097]} & -.1975\,{\fontsize{7.5}{9}\selectfont [-.2243, -.1705]} & +.0055\,{\fontsize{7.5}{9}\selectfont [+.0018, +.0101]} & -.1414\,{\fontsize{7.5}{9}\selectfont [-.2178, -.0669]} \\

Q25-1 & +.0158\,{\fontsize{7.5}{9}\selectfont [+.0057, +.0276]} & -.0933\,{\fontsize{7.5}{9}\selectfont [-.1158, -.0716]} & +.0123\,{\fontsize{7.5}{9}\selectfont [+.0018, +.0294]} & +.0198\,{\fontsize{7.5}{9}\selectfont [+.0066, +.0364]} \\

ESC-RL & -.0015\,{\fontsize{7.5}{9}\selectfont [-.0304, +.0222]} & -.1469\,{\fontsize{7.5}{9}\selectfont [-.1683, -.1252]} & +.0083\,{\fontsize{7.5}{9}\selectfont [+.0018, +.0175]} & -.1078\,{\fontsize{7.5}{9}\selectfont [-.1674, -.0470]} \\

MPO & -.0249\,{\fontsize{7.5}{9}\selectfont [-.0533, +.0058]} & -.2826\,{\fontsize{7.5}{9}\selectfont [-.3138, -.2516]} & -.0608\,{\fontsize{7.5}{9}\selectfont [-.1275, +.0059]} & -.1913\,{\fontsize{7.5}{9}\selectfont [-.2647, -.1139]} \\

R2Gen & +.0061\,{\fontsize{7.5}{9}\selectfont [+.0035, +.0089]} & -.3622\,{\fontsize{7.5}{9}\selectfont [-.3988, -.3275]} & +.0054\,{\fontsize{7.5}{9}\selectfont [+.0018, +.0099]} & -.1205\,{\fontsize{7.5}{9}\selectfont [-.1801, -.0626]} \\
\bottomrule\end{tabular}

\caption{IU: macro-oriented partial-minus-constrained target contrasts.}\label{tab:app-rq2-diff-iu-macro-partial}\end{table*}

\begin{table*}[!htbp]\centering\scriptsize

\setlength{\tabcolsep}{3pt}\renewcommand{\arraystretch}{1.0}

\begin{tabular}{@{}ccccc@{}}\toprule

\textbf{Generator} & \textbf{Ma-14} & \textbf{Mi-14} & \textbf{Ma-5} & \textbf{Mi-5}\\ \midrule

Q35 & -.0357\,{\fontsize{7.5}{9}\selectfont [-.0795, +.0117]} & -.3287\,{\fontsize{7.5}{9}\selectfont [-.3676, -.2918]} & -.0310\,{\fontsize{7.5}{9}\selectfont [-.1101, +.0451]} & -.2194\,{\fontsize{7.5}{9}\selectfont [-.2887, -.1504]} \\

IV35 & -.0484\,{\fontsize{7.5}{9}\selectfont [-.0942, +.0007]} & -.3221\,{\fontsize{7.5}{9}\selectfont [-.3537, -.2911]} & -.0558\,{\fontsize{7.5}{9}\selectfont [-.1221, +.0167]} & -.2471\,{\fontsize{7.5}{9}\selectfont [-.3013, -.1887]} \\

Q25-3 & +.0198\,{\fontsize{7.5}{9}\selectfont [-.0158, +.0558]} & -.1659\,{\fontsize{7.5}{9}\selectfont [-.2052, -.1266]} & +.0495\,{\fontsize{7.5}{9}\selectfont [-.0024, +.1010]} & -.0870\,{\fontsize{7.5}{9}\selectfont [-.1742, -.0038]} \\

Q25-1 & +.0931\,{\fontsize{7.5}{9}\selectfont [+.0652, +.1195]} & -.0061\,{\fontsize{7.5}{9}\selectfont [-.0423, +.0287]} & +.1191\,{\fontsize{7.5}{9}\selectfont [+.0850, +.1515]} & +.1173\,{\fontsize{7.5}{9}\selectfont [+.0924, +.1431]} \\

ESC-RL & +.0515\,{\fontsize{7.5}{9}\selectfont [+.0176, +.0841]} & -.1020\,{\fontsize{7.5}{9}\selectfont [-.1404, -.0642]} & +.0632\,{\fontsize{7.5}{9}\selectfont [+.0152, +.1112]} & -.0419\,{\fontsize{7.5}{9}\selectfont [-.1157, +.0312]} \\

MPO & +.0009\,{\fontsize{7.5}{9}\selectfont [-.0355, +.0405]} & -.2802\,{\fontsize{7.5}{9}\selectfont [-.3236, -.2370]} & -.0428\,{\fontsize{7.5}{9}\selectfont [-.1230, +.0475]} & -.1362\,{\fontsize{7.5}{9}\selectfont [-.2215, -.0487]} \\

R2Gen & +.0639\,{\fontsize{7.5}{9}\selectfont [+.0484, +.0792]} & -.3780\,{\fontsize{7.5}{9}\selectfont [-.4273, -.3306]} & +.0935\,{\fontsize{7.5}{9}\selectfont [+.0644, +.1224]} & -.0642\,{\fontsize{7.5}{9}\selectfont [-.1354, +.0038]} \\
\bottomrule\end{tabular}

\caption{IU: macro-oriented free-minus-constrained target contrasts.}\label{tab:app-rq2-diff-iu-macro-free}\end{table*}

\begin{table*}[!htbp]\centering\scriptsize

\setlength{\tabcolsep}{3pt}\renewcommand{\arraystretch}{1.0}

\begin{tabular}{@{}ccccc@{}}\toprule

\textbf{Generator} & \textbf{Ma-14} & \textbf{Mi-14} & \textbf{Ma-5} & \textbf{Mi-5}\\ \midrule

Q35 & -.0635\,{\fontsize{7.5}{9}\selectfont [-.1071, .0000]} & -.0021\,{\fontsize{7.5}{9}\selectfont [-.0082, +.0019]} & -.1333\,{\fontsize{7.5}{9}\selectfont [-.2000, .0000]} & -.0095\,{\fontsize{7.5}{9}\selectfont [-.0363, +.0060]} \\

IV35 & -.0476\,{\fontsize{7.5}{9}\selectfont [-.0714, .0000]} & .0000\,{\fontsize{7.5}{9}\selectfont [-.0034, +.0028]} & -.1333\,{\fontsize{7.5}{9}\selectfont [-.2000, .0000]} & -.0045\,{\fontsize{7.5}{9}\selectfont [-.0279, +.0105]} \\

Q25-3 & .0000\,{\fontsize{7.5}{9}\selectfont [.0000, .0000]} & +.0013\,{\fontsize{7.5}{9}\selectfont [.0000, +.0033]} & .0000\,{\fontsize{7.5}{9}\selectfont [.0000, .0000]} & +.0014\,{\fontsize{7.5}{9}\selectfont [.0000, +.0051]} \\

Q25-1 & .0000\,{\fontsize{7.5}{9}\selectfont [.0000, .0000]} & .0000\,{\fontsize{7.5}{9}\selectfont [.0000, .0000]} & .0000\,{\fontsize{7.5}{9}\selectfont [.0000, .0000]} & .0000\,{\fontsize{7.5}{9}\selectfont [.0000, .0000]} \\

ESC-RL & .0000\,{\fontsize{7.5}{9}\selectfont [.0000, .0000]} & +.0006\,{\fontsize{7.5}{9}\selectfont [.0000, +.0021]} & .0000\,{\fontsize{7.5}{9}\selectfont [.0000, .0000]} & .0000\,{\fontsize{7.5}{9}\selectfont [.0000, .0000]} \\

MPO & .0000\,{\fontsize{7.5}{9}\selectfont [.0000, .0000]} & +.0007\,{\fontsize{7.5}{9}\selectfont [.0000, +.0021]} & .0000\,{\fontsize{7.5}{9}\selectfont [.0000, .0000]} & +.0012\,{\fontsize{7.5}{9}\selectfont [.0000, +.0045]} \\

R2Gen & .0000\,{\fontsize{7.5}{9}\selectfont [.0000, .0000]} & .0000\,{\fontsize{7.5}{9}\selectfont [.0000, .0000]} & .0000\,{\fontsize{7.5}{9}\selectfont [.0000, .0000]} & .0000\,{\fontsize{7.5}{9}\selectfont [.0000, .0000]} \\
\bottomrule\end{tabular}

\caption{IU: micro-oriented partial-minus-constrained target contrasts.}\label{tab:app-rq2-diff-iu-micro-partial}\end{table*}

\begin{table*}[!htbp]\centering\scriptsize

\setlength{\tabcolsep}{3pt}\renewcommand{\arraystretch}{1.0}

\begin{tabular}{@{}ccccc@{}}\toprule

\textbf{Generator} & \textbf{Ma-14} & \textbf{Mi-14} & \textbf{Ma-5} & \textbf{Mi-5}\\ \midrule

Q35 & -.0788\,{\fontsize{7.5}{9}\selectfont [-.1350, +.0113]} & +.0042\,{\fontsize{7.5}{9}\selectfont [-.0136, +.0216]} & -.1549\,{\fontsize{7.5}{9}\selectfont [-.2564, +.0192]} & +.0343\,{\fontsize{7.5}{9}\selectfont [-.0548, +.1214]} \\

IV35 & -.0748\,{\fontsize{7.5}{9}\selectfont [-.1161, -.0078]} & -.0033\,{\fontsize{7.5}{9}\selectfont [-.0179, +.0121]} & -.1781\,{\fontsize{7.5}{9}\selectfont [-.2805, -.0057]} & -.0314\,{\fontsize{7.5}{9}\selectfont [-.1010, +.0331]} \\

Q25-3 & +.0179\,{\fontsize{7.5}{9}\selectfont [-.0156, +.0467]} & +.0024\,{\fontsize{7.5}{9}\selectfont [-.0153, +.0201]} & +.0698\,{\fontsize{7.5}{9}\selectfont [+.0203, +.1206]} & +.1789\,{\fontsize{7.5}{9}\selectfont [+.0861, +.2733]} \\

Q25-1 & +.0705\,{\fontsize{7.5}{9}\selectfont [+.0523, +.0872]} & +.0116\,{\fontsize{7.5}{9}\selectfont [-.0118, +.0356]} & +.1520\,{\fontsize{7.5}{9}\selectfont [+.1084, +.1914]} & +.3464\,{\fontsize{7.5}{9}\selectfont [+.2584, +.4246]} \\

ESC-RL & +.0137\,{\fontsize{7.5}{9}\selectfont [-.0045, +.0336]} & -.0003\,{\fontsize{7.5}{9}\selectfont [-.0178, +.0172]} & +.0257\,{\fontsize{7.5}{9}\selectfont [-.0193, +.0758]} & +.0927\,{\fontsize{7.5}{9}\selectfont [+.0162, +.1741]} \\

MPO & +.0173\,{\fontsize{7.5}{9}\selectfont [-.0167, +.0509]} & -.0026\,{\fontsize{7.5}{9}\selectfont [-.0218, +.0169]} & +.0778\,{\fontsize{7.5}{9}\selectfont [+.0172, +.1388]} & +.2226\,{\fontsize{7.5}{9}\selectfont [+.1126, +.3282]} \\

R2Gen & +.0596\,{\fontsize{7.5}{9}\selectfont [+.0413, +.0776]} & +.0101\,{\fontsize{7.5}{9}\selectfont [-.0129, +.0341]} & +.0990\,{\fontsize{7.5}{9}\selectfont [+.0511, +.1464]} & +.1938\,{\fontsize{7.5}{9}\selectfont [+.0972, +.2908]} \\
\bottomrule\end{tabular}

\caption{IU: micro-oriented free-minus-constrained target contrasts.}\label{tab:app-rq2-diff-iu-micro-free}\end{table*}

\begin{table*}[!htbp]\centering\scriptsize

\setlength{\tabcolsep}{3pt}\renewcommand{\arraystretch}{1.0}

\begin{tabular}{@{}ccccc@{}}\toprule

\textbf{Generator} & \textbf{Ma-14} & \textbf{Mi-14} & \textbf{Ma-5} & \textbf{Mi-5}\\ \midrule

Q35 & +.0642\,{\fontsize{7.5}{9}\selectfont [-.0304, +.1348]} & -.1935\,{\fontsize{7.5}{9}\selectfont [-.2167, -.1698]} & +.1378\,{\fontsize{7.5}{9}\selectfont [-.0595, +.2558]} & -.2273\,{\fontsize{7.5}{9}\selectfont [-.2907, -.1571]} \\

IV35 & +.0068\,{\fontsize{7.5}{9}\selectfont [-.0456, +.0458]} & -.1775\,{\fontsize{7.5}{9}\selectfont [-.1937, -.1614]} & +.0736\,{\fontsize{7.5}{9}\selectfont [-.0961, +.1855]} & -.2480\,{\fontsize{7.5}{9}\selectfont [-.3010, -.1907]} \\

Q25-3 & -.0106\,{\fontsize{7.5}{9}\selectfont [-.0369, +.0097]} & -.1988\,{\fontsize{7.5}{9}\selectfont [-.2258, -.1719]} & +.0055\,{\fontsize{7.5}{9}\selectfont [+.0018, +.0101]} & -.1428\,{\fontsize{7.5}{9}\selectfont [-.2195, -.0684]} \\

Q25-1 & +.0158\,{\fontsize{7.5}{9}\selectfont [+.0057, +.0276]} & -.0933\,{\fontsize{7.5}{9}\selectfont [-.1158, -.0716]} & +.0123\,{\fontsize{7.5}{9}\selectfont [+.0018, +.0294]} & +.0198\,{\fontsize{7.5}{9}\selectfont [+.0066, +.0364]} \\

ESC-RL & -.0015\,{\fontsize{7.5}{9}\selectfont [-.0304, +.0222]} & -.1476\,{\fontsize{7.5}{9}\selectfont [-.1690, -.1257]} & +.0083\,{\fontsize{7.5}{9}\selectfont [+.0018, +.0175]} & -.1078\,{\fontsize{7.5}{9}\selectfont [-.1674, -.0470]} \\

MPO & -.0249\,{\fontsize{7.5}{9}\selectfont [-.0533, +.0058]} & -.2833\,{\fontsize{7.5}{9}\selectfont [-.3145, -.2521]} & -.0608\,{\fontsize{7.5}{9}\selectfont [-.1275, +.0059]} & -.1926\,{\fontsize{7.5}{9}\selectfont [-.2664, -.1150]} \\

R2Gen & +.0061\,{\fontsize{7.5}{9}\selectfont [+.0035, +.0089]} & -.3622\,{\fontsize{7.5}{9}\selectfont [-.3988, -.3275]} & +.0054\,{\fontsize{7.5}{9}\selectfont [+.0018, +.0099]} & -.1205\,{\fontsize{7.5}{9}\selectfont [-.1801, -.0626]} \\
\bottomrule\end{tabular}

\caption{IU: same-metric objective-by-partial-relaxation interactions.}\label{tab:app-rq2-interaction-iu-partial}\end{table*}

\begin{table*}[!htbp]\centering\scriptsize

\setlength{\tabcolsep}{3pt}\renewcommand{\arraystretch}{1.0}

\begin{tabular}{@{}ccccc@{}}\toprule

\textbf{Generator} & \textbf{Ma-14} & \textbf{Mi-14} & \textbf{Ma-5} & \textbf{Mi-5}\\ \midrule

Q35 & +.0431\,{\fontsize{7.5}{9}\selectfont [-.0648, +.1237]} & -.3329\,{\fontsize{7.5}{9}\selectfont [-.3720, -.2943]} & +.1239\,{\fontsize{7.5}{9}\selectfont [-.0850, +.2545]} & -.2537\,{\fontsize{7.5}{9}\selectfont [-.3556, -.1518]} \\

IV35 & +.0263\,{\fontsize{7.5}{9}\selectfont [-.0396, +.0773]} & -.3189\,{\fontsize{7.5}{9}\selectfont [-.3540, -.2843]} & +.1223\,{\fontsize{7.5}{9}\selectfont [-.0680, +.2460]} & -.2156\,{\fontsize{7.5}{9}\selectfont [-.2991, -.1271]} \\

Q25-3 & +.0019\,{\fontsize{7.5}{9}\selectfont [-.0360, +.0401]} & -.1683\,{\fontsize{7.5}{9}\selectfont [-.2090, -.1270]} & -.0204\,{\fontsize{7.5}{9}\selectfont [-.0891, +.0426]} & -.2659\,{\fontsize{7.5}{9}\selectfont [-.3959, -.1389]} \\

Q25-1 & +.0226\,{\fontsize{7.5}{9}\selectfont [-.0019, +.0479]} & -.0177\,{\fontsize{7.5}{9}\selectfont [-.0558, +.0188]} & -.0329\,{\fontsize{7.5}{9}\selectfont [-.0674, +.0040]} & -.2290\,{\fontsize{7.5}{9}\selectfont [-.3040, -.1488]} \\

ESC-RL & +.0378\,{\fontsize{7.5}{9}\selectfont [+.0030, +.0706]} & -.1017\,{\fontsize{7.5}{9}\selectfont [-.1417, -.0622]} & +.0374\,{\fontsize{7.5}{9}\selectfont [-.0112, +.0841]} & -.1346\,{\fontsize{7.5}{9}\selectfont [-.2326, -.0382]} \\

MPO & -.0164\,{\fontsize{7.5}{9}\selectfont [-.0579, +.0270]} & -.2776\,{\fontsize{7.5}{9}\selectfont [-.3205, -.2350]} & -.1206\,{\fontsize{7.5}{9}\selectfont [-.2157, -.0115]} & -.3589\,{\fontsize{7.5}{9}\selectfont [-.4876, -.2293]} \\

R2Gen & +.0043\,{\fontsize{7.5}{9}\selectfont [-.0130, +.0219]} & -.3881\,{\fontsize{7.5}{9}\selectfont [-.4314, -.3461]} & -.0056\,{\fontsize{7.5}{9}\selectfont [-.0438, +.0314]} & -.2581\,{\fontsize{7.5}{9}\selectfont [-.3461, -.1712]} \\
\bottomrule\end{tabular}

\caption{IU: same-metric objective-by-free-relaxation interactions.}\label{tab:app-rq2-interaction-iu-free}\end{table*}

\subsection{Candidate Coverage and Score Ordering}
\label{app:rq2-coverage}
The corresponding results are reported in Tables~\ref{tab:app-rq2-coverage-mimic-macro}, \ref{tab:app-rq2-coverage-mimic-micro}, \ref{tab:app-rq2-coverage-iu-macro}, \ref{tab:app-rq2-coverage-iu-micro}.

Unique-vector counts concern structured candidate findings. Out-of-pool rates
measure whether the free vector appears in that pool; nearest Hamming distances
are numbers of differing labels, not normalized fractions. Score increments are
within-objective diagnostics and should not be compared across differently scaled objectives.
The export records no negative score-ordering violations in the reported conditions.

\begin{table*}[!tp]
\centering
\begin{minipage}[t]{0.48\textwidth}
\centering\scriptsize

\setlength{\tabcolsep}{1.5pt}\renewcommand{\arraystretch}{1.0}

\begin{tabular}{@{}ccccccc@{}}\toprule

\textbf{Generator} & \textbf{Unique} & \textbf{Out} & \textbf{$H$} & \textbf{$H_o$} & \textbf{$\Delta S_{PC}$} & \textbf{$\Delta S_{FP}$}\\ \midrule

Q25-3 & 8.4525 & .4253 & .6257 & 1.4713 & .1426 & .0648 \\

Q25-1 & 8.6711 & .6124 & 1.1313 & 1.8472 & .3213 & .1377 \\

Q35 & 4.7968 & .9839 & 2.2668 & 2.3038 & .8428 & .1246 \\

IV35 & 6.0810 & .5740 & 1.0105 & 1.7603 & .3411 & .0770 \\

ESC-RL & 8.9246 & .4057 & .5552 & 1.3683 & .0926 & .0608 \\

MPO & 9.3149 & .4385 & .6229 & 1.4204 & .1286 & .0609 \\

EditGRPO & 9.4260 & .3911 & .5370 & 1.3732 & .0843 & .0565 \\

DPO-G & 6.7989 & .5552 & .9351 & 1.6843 & .2027 & .1036 \\

R2Gen & 7.3296 & .5978 & 1.1013 & 1.8423 & .4022 & .1237 \\

R2CMN & 7.6683 & .5293 & .8862 & 1.6741 & .1603 & .1076 \\

L15 & 8.7570 & .3177 & .3973 & 1.2505 & .0689 & .0363 \\

LN-M & 9.4637 & .2737 & .3331 & 1.2168 & .0298 & .0298 \\

LN-V & 8.6913 & .3073 & .3834 & 1.2477 & .0468 & .0368 \\

\bottomrule\end{tabular}

\caption{MIMIC: macro-oriented coverage and mean score increments.}\label{tab:app-rq2-coverage-mimic-macro}
\end{minipage}
\hfill
\begin{minipage}[t]{0.48\textwidth}
\centering\scriptsize

\setlength{\tabcolsep}{1.5pt}\renewcommand{\arraystretch}{1.0}

\begin{tabular}{@{}ccccccc@{}}\toprule

\textbf{Generator} & \textbf{Unique} & \textbf{Out} & \textbf{$H$} & \textbf{$H_o$} & \textbf{$\Delta S_{PC}$} & \textbf{$\Delta S_{FP}$}\\ \midrule

Q25-3 & 8.4525 & .1725 & .1913 & 1.1093 & .0090 & .0359 \\

Q25-1 & 8.6711 & .2772 & .3261 & 1.1763 & .0138 & .0888 \\

Q35 & 4.7968 & .4923 & .7528 & 1.5291 & .0209 & .3939 \\

IV35 & 6.0810 & .3310 & .4078 & 1.2321 & .0220 & .1217 \\

ESC-RL & 8.9246 & .1795 & .1962 & 1.0934 & .0089 & .0402 \\

MPO & 9.3149 & .1878 & .2067 & 1.1004 & .0086 & .0456 \\

EditGRPO & 9.4260 & .1774 & .1934 & 1.0906 & .0077 & .0401 \\

DPO-G & 6.7989 & .2821 & .3415 & 1.2104 & .0123 & .1099 \\

R2Gen & 7.3296 & .2444 & .2758 & 1.1286 & .0096 & .0724 \\

R2CMN & 7.6683 & .2758 & .3038 & 1.1013 & .0133 & .0821 \\

L15 & 8.7570 & .1145 & .1180 & 1.0305 & .0067 & .0209 \\

LN-M & 9.4637 & .1278 & .1390 & 1.0874 & .0073 & .0222 \\

LN-V & 8.6913 & .1194 & .1250 & 1.0468 & .0067 & .0239 \\

\bottomrule\end{tabular}

\caption{MIMIC: micro-oriented coverage and mean score increments.}\label{tab:app-rq2-coverage-mimic-micro}
\end{minipage}
\end{table*}

\begin{table*}[!tp]
\centering
\begin{minipage}[t]{0.48\textwidth}
\centering\scriptsize

\setlength{\tabcolsep}{1.5pt}\renewcommand{\arraystretch}{1.0}

\begin{tabular}{@{}ccccccc@{}}\toprule

\textbf{Generator} & \textbf{Unique} & \textbf{Out} & \textbf{$H$} & \textbf{$H_o$} & \textbf{$\Delta S_{PC}$} & \textbf{$\Delta S_{FP}$}\\ \midrule

Q35 & 2.0673 & 1.0000 & 4.3004 & 4.3004 & .3093 & 1.5921 \\

IV35 & 3.2601 & 1.0000 & 3.6390 & 3.6390 & .1345 & 1.3682 \\

Q25-3 & 3.9776 & 1.0000 & 8.5045 & 8.5045 & 5.6389 & 7.5089 \\

Q25-1 & 2.9641 & 1.0000 & 9.4933 & 9.4933 & 13.0188 & 2.7424 \\

ESC-RL & 4.1143 & 1.0000 & 7.1368 & 7.1368 & 2.4039 & 6.3558 \\

MPO & 3.5000 & 1.0000 & 8.8946 & 8.8946 & 5.9429 & 4.8077 \\

R2Gen & 2.0538 & 1.0000 & 1.8453 & 1.8453 & 42.4451 & 5.5198 \\

EditGRPO & 2.1502 & 1.0000 & 1.9776 & 1.9776 & 18.3463 & 17.0957 \\

DPO-G & 2.2892 & 1.0000 & 8.8700 & 8.8700 & 6.3721 & 9.8748 \\

R2CMN & 3.5628 & 1.0000 & 9.6861 & 9.6861 & 1.9369 & 1.1149 \\

L15 & 2.9798 & 1.0000 & 11.8767 & 11.8767 & 63.6476 & 75.4937 \\

LN-M & 3.7108 & 1.0000 & 1.0000 & 1.0000 & 26.0993 & 29.3715 \\

LN-V & 3.9305 & 1.0000 & 1.8386 & 1.8386 & 4.5805 & 46.9476 \\

\bottomrule\end{tabular}

\caption{IU: macro-oriented coverage and mean score increments.}\label{tab:app-rq2-coverage-iu-macro}
\end{minipage}
\hfill
\begin{minipage}[t]{0.48\textwidth}
\centering\scriptsize

\setlength{\tabcolsep}{1.5pt}\renewcommand{\arraystretch}{1.0}

\begin{tabular}{@{}ccccccc@{}}\toprule

\textbf{Generator} & \textbf{Unique} & \textbf{Out} & \textbf{$H$} & \textbf{$H_o$} & \textbf{$\Delta S_{PC}$} & \textbf{$\Delta S_{FP}$}\\ \midrule

Q35 & 2.0673 & .1390 & .1906 & 1.3710 & .0063 & .0593 \\

IV35 & 3.2601 & .1009 & .1166 & 1.1556 & .0065 & .0263 \\

Q25-3 & 3.9776 & .1570 & .2242 & 1.4286 & .0043 & .0728 \\

Q25-1 & 2.9641 & .1996 & .3632 & 1.8202 & .0000 & .1672 \\

ESC-RL & 4.1143 & .1570 & .2085 & 1.3286 & .0022 & .0699 \\

MPO & 3.5000 & .1771 & .2825 & 1.5949 & .0021 & .1051 \\

R2Gen & 2.0538 & .1996 & .3610 & 1.8090 & .0000 & .1511 \\

EditGRPO & 2.1502 & .1973 & .3587 & 1.8182 & .0000 & .1646 \\

DPO-G & 2.2892 & .1928 & .3363 & 1.7442 & .0000 & .1389 \\

R2CMN & 3.5628 & .1839 & .2825 & 1.5366 & .0000 & .1065 \\

L15 & 2.9798 & .1883 & .2960 & 1.5714 & .0000 & .1198 \\

LN-M & 3.7108 & .1973 & .3430 & 1.7386 & .0011 & .1464 \\

LN-V & 3.9305 & .1883 & .2937 & 1.5595 & .0000 & .1184 \\

\bottomrule\end{tabular}

\caption{IU: micro-oriented coverage and mean score increments.}\label{tab:app-rq2-coverage-iu-micro}
\end{minipage}
\end{table*}

\subsection{Assertion-Change Decomposition}
The corresponding results are reported in Tables~\ref{tab:app-rq2-decomp-mimic-macro}, \ref{tab:app-rq2-decomp-mimic-micro}, \ref{tab:app-rq2-decomp-iu-macro}, \ref{tab:app-rq2-decomp-iu-micro}.

Counts below sum the exported label-wise newly asserted and withdrawn positives
over fourteen findings on reference-scored cells. New/withdrawn TP/FP identify whether
those assertions agree with the reference. They diagnose how free decisions change
the constrained operating point without treating every added assertion as beneficial.

\begin{table*}[!tp]
\centering
\begin{minipage}[t]{0.48\textwidth}
\centering\footnotesize

\setlength{\tabcolsep}{2.5pt}\renewcommand{\arraystretch}{1.0}

\begin{tabular}{@{}c|cc|cc|cc@{}}\toprule

\textbf{Generator} & \shortstack{\textbf{New}\\\textbf{TP}} & \shortstack{\textbf{New}\\\textbf{FP}} & \shortstack{\textbf{Wdr}\\\textbf{TP}} & \shortstack{\textbf{Wdr}\\\textbf{FP}} & \shortstack{\textbf{Net}\\\textbf{TP}} & \shortstack{\textbf{Net}\\\textbf{FP}}\\ \midrule

Q25-3 & 207 & 587 & 26 & 62 & 181 & 525 \\

Q25-1 & 323 & 1153 & 21 & 70 & 302 & 1083 \\

Q35 & 257 & 2278 & 100 & 535 & 157 & 1743 \\

IV35 & 234 & 939 & 50 & 174 & 184 & 765 \\

ESC-RL & 184 & 501 & 22 & 57 & 162 & 444 \\

MPO & 197 & 575 & 31 & 76 & 166 & 499 \\

EditGRPO & 168 & 489 & 20 & 66 & 148 & 423 \\

DPO-G & 311 & 905 & 28 & 61 & 283 & 844 \\

R2Gen & 292 & 1139 & 20 & 74 & 272 & 1065 \\

R2CMN & 268 & 854 & 16 & 68 & 252 & 786 \\

L15 & 140 & 356 & 23 & 36 & 117 & 320 \\

LN-M & 112 & 283 & 21 & 45 & 91 & 238 \\

LN-V & 122 & 359 & 13 & 39 & 109 & 320 \\

\bottomrule\end{tabular}

\caption{MIMIC: macro-oriented free-minus-constrained assertion decomposition.}\label{tab:app-rq2-decomp-mimic-macro}
\end{minipage}
\hfill
\begin{minipage}[t]{0.48\textwidth}
\centering\footnotesize

\setlength{\tabcolsep}{2.5pt}\renewcommand{\arraystretch}{1.0}

\begin{tabular}{@{}c|cc|cc|cc@{}}\toprule

\textbf{Generator} & \shortstack{\textbf{New}\\\textbf{TP}} & \shortstack{\textbf{New}\\\textbf{FP}} & \shortstack{\textbf{Wdr}\\\textbf{TP}} & \shortstack{\textbf{Wdr}\\\textbf{FP}} & \shortstack{\textbf{Net}\\\textbf{TP}} & \shortstack{\textbf{Net}\\\textbf{FP}}\\ \midrule

Q25-3 & 66 & 152 & 13 & 28 & 53 & 124 \\

Q25-1 & 155 & 221 & 19 & 42 & 136 & 179 \\

Q35 & 195 & 282 & 81 & 476 & 114 & -194 \\

IV35 & 152 & 224 & 48 & 132 & 104 & 92 \\

ESC-RL & 75 & 151 & 13 & 25 & 62 & 126 \\

MPO & 86 & 150 & 11 & 27 & 75 & 123 \\

EditGRPO & 70 & 140 & 17 & 32 & 53 & 108 \\

DPO-G & 156 & 255 & 13 & 31 & 143 & 224 \\

R2Gen & 113 & 209 & 15 & 28 & 98 & 181 \\

R2CMN & 130 & 216 & 15 & 48 & 115 & 168 \\

L15 & 56 & 84 & 8 & 13 & 48 & 71 \\

LN-M & 46 & 105 & 9 & 30 & 37 & 75 \\

LN-V & 45 & 93 & 8 & 24 & 37 & 69 \\

\bottomrule\end{tabular}

\caption{MIMIC: micro-oriented free-minus-constrained assertion decomposition.}\label{tab:app-rq2-decomp-mimic-micro}
\end{minipage}
\end{table*}

\begin{table*}[!tp]
\centering
\begin{minipage}[t]{0.48\textwidth}
\centering\footnotesize

\setlength{\tabcolsep}{2.5pt}\renewcommand{\arraystretch}{1.0}

\begin{tabular}{@{}c|cc|cc|cc@{}}\toprule

\textbf{Generator} & \shortstack{\textbf{New}\\\textbf{TP}} & \shortstack{\textbf{New}\\\textbf{FP}} & \shortstack{\textbf{Wdr}\\\textbf{TP}} & \shortstack{\textbf{Wdr}\\\textbf{FP}} & \shortstack{\textbf{Net}\\\textbf{TP}} & \shortstack{\textbf{Net}\\\textbf{FP}}\\ \midrule

Q35 & 85 & 1789 & 7 & 18 & 78 & 1771 \\

IV35 & 50 & 1540 & 1 & 15 & 49 & 1525 \\

Q25-3 & 249 & 3511 & 0 & 14 & 249 & 3497 \\

Q25-1 & 361 & 3816 & 1 & 19 & 360 & 3797 \\

ESC-RL & 270 & 2896 & 0 & 7 & 270 & 2889 \\

MPO & 199 & 3728 & 3 & 8 & 196 & 3720 \\

R2Gen & 200 & 4548 & 3 & 19 & 197 & 4529 \\

EditGRPO & 198 & 4609 & 2 & 2 & 196 & 4607 \\

DPO-G & 192 & 3688 & 4 & 6 & 188 & 3682 \\

R2CMN & 255 & 4001 & 0 & 12 & 255 & 3989 \\

L15 & 348 & 4885 & 0 & 1 & 348 & 4884 \\

LN-M & 174 & 4199 & 2 & 17 & 172 & 4182 \\

LN-V & 419 & 4338 & 0 & 1 & 419 & 4337 \\

\bottomrule\end{tabular}

\caption{IU: macro-oriented free-minus-constrained assertion decomposition.}\label{tab:app-rq2-decomp-iu-macro}
\end{minipage}
\hfill
\begin{minipage}[t]{0.48\textwidth}
\centering\footnotesize

\setlength{\tabcolsep}{2.5pt}\renewcommand{\arraystretch}{1.0}

\begin{tabular}{@{}c|cc|cc|cc@{}}\toprule

\textbf{Generator} & \shortstack{\textbf{New}\\\textbf{TP}} & \shortstack{\textbf{New}\\\textbf{FP}} & \shortstack{\textbf{Wdr}\\\textbf{TP}} & \shortstack{\textbf{Wdr}\\\textbf{FP}} & \shortstack{\textbf{Net}\\\textbf{TP}} & \shortstack{\textbf{Net}\\\textbf{FP}}\\ \midrule

Q35 & 21 & 38 & 8 & 17 & 13 & 21 \\

IV35 & 10 & 19 & 10 & 14 & 0 & 5 \\

Q25-3 & 30 & 64 & 1 & 3 & 29 & 61 \\

Q25-1 & 53 & 104 & 1 & 4 & 52 & 100 \\

ESC-RL & 27 & 63 & 0 & 3 & 27 & 60 \\

MPO & 37 & 82 & 3 & 2 & 34 & 80 \\

R2Gen & 50 & 100 & 2 & 7 & 48 & 93 \\

EditGRPO & 50 & 106 & 0 & 1 & 50 & 105 \\

DPO-G & 46 & 99 & 1 & 2 & 45 & 97 \\

R2CMN & 35 & 81 & 1 & 6 & 34 & 75 \\

L15 & 41 & 88 & 0 & 0 & 41 & 88 \\

LN-M & 43 & 95 & 1 & 15 & 42 & 80 \\

LN-V & 40 & 88 & 0 & 0 & 40 & 88 \\

\bottomrule\end{tabular}

\caption{IU: micro-oriented free-minus-constrained assertion decomposition.}\label{tab:app-rq2-decomp-iu-micro}
\end{minipage}
\end{table*}

\subsection{Threshold and Candidate-Pool Diagnostics}
\label{app:rq2-diagnostics}
\paragraph{Frozen thresholds.}
The corresponding results are reported in Tables~\ref{tab:app-threshold-mimic}, \ref{tab:app-threshold-iu}, \ref{tab:app-threshold-label-mimic}, \ref{tab:app-threshold-label-iu}.

We compute $\tau_L=b_L/(a_L+2b_L)$ from the final frozen coefficients.
This exactly reproduces the complementary-probability free rule, including
numerical coefficient floors; it is not reconstructed from empirical test F1.
Near-zero thresholds are defined by $\tau_L<10^{-8}$.
On IU, every configuration has one to eleven such macro thresholds.
For Qwen2.5-VL-7B (3 epoch), four labels (Lung Lesion, Consolidation, Pneumonia, Fracture)
have near-zero macro thresholds and a free positive rate of one.
These patterns depend on the soft-count scoring procedure and fitted signals;
they do not establish that training prevalence alone causes threshold collapse.
For reference-only mask $m$, precision of added positives is
\begin{equation}
P_{\mathrm{add}}=\frac{\sum_{i,L}m_{iL}\mathbf1[z^F_{iL}=1,z^C_{iL}=0,y_{iL}=1]}
 {\sum_{i,L}m_{iL}\mathbf1[z^F_{iL}=1,z^C_{iL}=0]}.
\end{equation}
An empty denominator is undefined. Positive rates and their summed assertions
in this diagnostic use all test inputs/labels, without reference masking;
added-positive counts and precision use reference-scored cells.
Thus the diagnostic assertion statistic can differ from the original RQ2 export.

\begin{table*}[!htbp]
\centering\scriptsize
\setlength{\tabcolsep}{4pt}\renewcommand{\arraystretch}{1.0}
\begin{tabular}{@{}ccccccccc@{}}
\toprule
\textbf{Generator} & \textbf{Ma near-zero} & \textbf{Ma assertions} & \textbf{Ma added} & \textbf{Ma add prec.} & \textbf{Mi near-zero} & \textbf{Mi assertions} & \textbf{Mi added} & \textbf{Mi add prec.}\\
\midrule
Q25-3 & 0 & 2.6851 & 794 & 0.2607 & 0 & 2.1522 & 218 & 0.3028 \\
Q25-1 & 0 & 3.0677 & 1476 & 0.2188 & 0 & 2.1620 & 376 & 0.4122 \\
Q35 & 0 & 3.9679 & 2535 & 0.1014 & 0 & 2.2849 & 477 & 0.4088 \\
IV35 & 0 & 2.9958 & 1173 & 0.1995 & 0 & 2.2004 & 376 & 0.4043 \\
ESC-RL & 0 & 2.6110 & 685 & 0.2686 & 0 & 2.1613 & 226 & 0.3319 \\
MPO & 0 & 2.6578 & 772 & 0.2552 & 0 & 2.1508 & 236 & 0.3644 \\
EditGRPO & 0 & 2.6208 & 657 & 0.2557 & 0 & 2.1557 & 210 & 0.3333 \\
DPO-G & 0 & 2.8729 & 1216 & 0.2558 & 0 & 2.1802 & 411 & 0.3796 \\
R2Gen & 0 & 3.0719 & 1431 & 0.2041 & 0 & 2.1669 & 322 & 0.3509 \\
R2CMN & 0 & 2.8499 & 1122 & 0.2389 & 0 & 2.1522 & 346 & 0.3757 \\
L15 & 0 & 2.5098 & 496 & 0.2823 & 0 & 2.1054 & 140 & 0.4000 \\
LN-M & 0 & 2.4197 & 395 & 0.2835 & 0 & 2.0915 & 151 & 0.3046 \\
LN-V & 0 & 2.4804 & 481 & 0.2536 & 0 & 2.0810 & 138 & 0.3261 \\
\bottomrule
\end{tabular}
\caption{MIMIC: frozen-threshold diagnostics. Near-zero counts use $\tau<10^{-8}$. Assertions sum unmasked per-label positive rates; added counts and precision pool reference-scored cells. Ma/Mi refer to scoring objectives.}\label{tab:app-threshold-mimic}
\end{table*}

\begin{table*}[!htbp]
\centering\scriptsize
\setlength{\tabcolsep}{4pt}\renewcommand{\arraystretch}{1.0}
\begin{tabular}{@{}ccccccccc@{}}
\toprule
\textbf{Generator} & \textbf{Ma near-zero} & \textbf{Ma assertions} & \textbf{Ma added} & \textbf{Ma add prec.} & \textbf{Mi near-zero} & \textbf{Mi assertions} & \textbf{Mi added} & \textbf{Mi add prec.}\\
\midrule
Q35 & 1 & 5.6121 & 1874 & 0.0454 & 0 & 1.1996 & 59 & 0.3559 \\
IV35 & 1 & 4.9798 & 1590 & 0.0314 & 0 & 1.1525 & 29 & 0.3448 \\
Q25-3 & 4 & 9.7601 & 3760 & 0.0662 & 0 & 1.2534 & 94 & 0.3191 \\
Q25-1 & 6 & 10.5448 & 4177 & 0.0864 & 0 & 1.3117 & 157 & 0.3376 \\
ESC-RL & 2 & 8.3879 & 3166 & 0.0853 & 0 & 1.2466 & 90 & 0.3000 \\
MPO & 2 & 10.1323 & 3927 & 0.0507 & 0 & 1.2735 & 119 & 0.3109 \\
R2Gen & 9 & 11.7713 & 4748 & 0.0421 & 0 & 1.3117 & 150 & 0.3333 \\
EditGRPO & 5 & 11.8924 & 4807 & 0.0412 & 0 & 1.3206 & 156 & 0.3205 \\
DPO-G & 2 & 9.8430 & 3880 & 0.0495 & 0 & 1.2960 & 145 & 0.3172 \\
R2CMN & 3 & 10.9283 & 4256 & 0.0599 & 0 & 1.2870 & 116 & 0.3017 \\
L15 & 11 & 12.9462 & 5233 & 0.0665 & 0 & 1.2960 & 129 & 0.3178 \\
LN-M & 6 & 11.1323 & 4373 & 0.0398 & 0 & 1.3027 & 138 & 0.3116 \\
LN-V & 10 & 11.8386 & 4757 & 0.0881 & 0 & 1.2937 & 128 & 0.3125 \\
\bottomrule
\end{tabular}
\caption{IU: frozen-threshold diagnostics. Near-zero counts use $\tau<10^{-8}$. Assertions sum unmasked per-label positive rates; added counts and precision pool reference-scored cells. Ma/Mi refer to scoring objectives.}\label{tab:app-threshold-iu}
\end{table*}

\begin{table*}[!htbp]
\centering\scriptsize
\setlength{\tabcolsep}{4pt}\renewcommand{\arraystretch}{1.0}
\begin{tabular}{@{}cccccccc@{}}
\toprule
\textbf{Label} & \textbf{Train prev.} & \textbf{Ma threshold} & \textbf{Ma pos. rate} & \textbf{Ma add prec.} & \textbf{Mi threshold} & \textbf{Mi pos. rate} & \textbf{Mi add prec.}\\
\midrule
No Finding & 0.2375 & 0.3540 & 0.4118 & 0.3614 & 0.3172 & 0.4375 & 0.2077 \\
Enlarged Cardiomediastinum & 0.0931 & 0.0802 & 0.2312 & 0.1115 & 0.3172 & 0.0045 & 0.8824 \\
Cardiomegaly & 0.2576 & 0.3132 & 0.2978 & 0.4626 & 0.3172 & 0.2948 & 0.4140 \\
Lung Lesion & 0.0395 & 0.1100 & 0.0488 & 0.1896 & 0.3172 & 0.0063 & 0.2857 \\
Lung Opacity & 0.3045 & 0.2864 & 0.4273 & 0.3912 & 0.3172 & 0.3990 & 0.3393 \\
Edema & 0.0955 & 0.2771 & 0.1332 & 0.5294 & 0.3172 & 0.1161 & 0.5587 \\
Consolidation & 0.0383 & 0.1507 & 0.0911 & 0.1148 & 0.3172 & 0.0283 & 0.3058 \\
Pneumonia & 0.0704 & 0.2106 & 0.1193 & 0.2565 & 0.3172 & 0.0792 & 0.3063 \\
Atelectasis & 0.2123 & 0.2880 & 0.3370 & 0.4172 & 0.3172 & 0.3066 & 0.4403 \\
Pneumothorax & 0.0334 & 0.3966 & 0.0100 & 0.4000 & 0.3172 & 0.0119 & 0.5000 \\
Pleural Effusion & 0.1839 & 0.3691 & 0.2120 & 0.5670 & 0.3172 & 0.2259 & 0.4554 \\
Pleural Other & 0.0169 & 0.0629 & 0.0826 & 0.0987 & 0.3172 & 0.0037 & 0.2381 \\
Fracture & 0.0387 & 0.0608 & 0.2100 & 0.0491 & 0.3172 & 0.0070 & 0.3030 \\
Support Devices & 0.2622 & 0.4180 & 0.2196 & 0.4737 & 0.3172 & 0.2364 & 0.2969 \\
\bottomrule
\end{tabular}
\caption{MIMIC: label-wise threshold diagnostics. Thresholds and positive rates are equal-weight means over thirteen configurations; added precision pools added scored cells across configurations. These aggregate associations are descriptive, not causal.}\label{tab:app-threshold-label-mimic}
\end{table*}

\begin{table*}[!htbp]
\centering\scriptsize
\setlength{\tabcolsep}{4pt}\renewcommand{\arraystretch}{1.0}
\begin{tabular}{@{}cccccccc@{}}
\toprule
\textbf{Label} & \textbf{Train prev.} & \textbf{Ma threshold} & \textbf{Ma pos. rate} & \textbf{Ma add prec.} & \textbf{Mi threshold} & \textbf{Mi pos. rate} & \textbf{Mi add prec.}\\
\midrule
No Finding & 0.6270 & 0.3676 & 0.8360 & 0.6433 & 0.2710 & 0.9283 & 0.2550 \\
Enlarged Cardiomediastinum & 0.0268 & 0.0044 & 0.8420 & 0.0195 & 0.2710 & 0.0000 & -- \\
Cardiomegaly & 0.0875 & 0.1132 & 0.4250 & 0.1722 & 0.2710 & 0.1261 & 0.4353 \\
Lung Lesion & 0.0446 & 0.0107 & 0.9719 & 0.0402 & 0.2710 & 0.0000 & -- \\
Lung Opacity & 0.1258 & 0.1537 & 0.4310 & 0.1754 & 0.2710 & 0.1578 & 0.2811 \\
Edema & 0.0126 & 0.0037 & 0.9922 & 0.0107 & 0.2710 & 0.0000 & -- \\
Consolidation & 0.0091 & 0.0036 & 0.6175 & 0.0035 & 0.2710 & 0.0000 & -- \\
Pneumonia & 0.0170 & 0.0042 & 0.6171 & 0.0210 & 0.2710 & 0.0000 & -- \\
Atelectasis & 0.0481 & 0.0371 & 0.8441 & 0.0642 & 0.2710 & 0.0021 & -- \\
Pneumothorax & 0.0043 & 0.0104 & 0.8013 & 0.0025 & 0.2710 & 0.0000 & -- \\
Pleural Effusion & 0.0296 & 0.0931 & 0.3130 & 0.0662 & 0.2710 & 0.0524 & 0.2167 \\
Pleural Other & 0.0138 & 0.0098 & 0.7018 & 0.0123 & 0.2710 & 0.0000 & -- \\
Fracture & 0.0248 & 0.0024 & 0.9934 & 0.0327 & 0.2710 & 0.0000 & -- \\
Support Devices & 0.0410 & 0.0496 & 0.5959 & 0.0285 & 0.2710 & 0.0060 & 0.0000 \\
\bottomrule
\end{tabular}
\caption{IU: label-wise threshold diagnostics. Thresholds and positive rates are equal-weight means over thirteen configurations; added precision pools added scored cells across configurations. These aggregate associations are descriptive, not causal.}\label{tab:app-threshold-label-iu}
\end{table*}
\paragraph{Reference-informed candidate oracle.}
The corresponding results are reported in Tables~\ref{tab:app-oracle-mimic}, \ref{tab:app-oracle-f1-mimic}, \ref{tab:app-oracle-iu}, \ref{tab:app-oracle-f1-iu}.

For each input, the oracle minimizes the number of errors on reference-scored cells:
\begin{equation}
z_i^{\mathrm{orc}}\in\arg\min_{z\in\mathcal Z_i}
\sum_Lm_{iL}\mathbf1[z_L\ne y_{iL}].
\end{equation}
It uses test references only for diagnosis, and is not deployable or a strict
upper bound on corpus F1. Minimizing per-input Hamming error does not maximize
macro/micro F1. Zero error concerns scored finding cells, not complete reports.
A missed-better fraction counts inputs where constrained selection has strictly
more masked errors than the oracle. It is relative to Hamming, not a proof that
the selection algorithm fails to optimize its own score.
Across configurations, zero-error candidates are available on 63.8\%/70.5\%
of MIMIC/IU inputs on average. Macro selection misses a lower-error candidate
on 51.3\%/47.2\%, and micro selection on 48.4\%/25.4\%.

\begin{table*}[!htbp]
\centering\scriptsize
\setlength{\tabcolsep}{4pt}\renewcommand{\arraystretch}{1.0}
\begin{tabular}{@{}cccccccc@{}}
\toprule
\textbf{Generator} & \textbf{Unique} & \textbf{Zero error (\%)} & \textbf{Oracle errors} & \textbf{Ma errors} & \textbf{Mi errors} & \textbf{Ma missed (\%)} & \textbf{Mi missed (\%)}\\
\midrule
Q25-3 & 8.4525 & 67.5978 & 0.4043 & 1.3436 & 1.2325 & 56.8436 & 53.7011 \\
Q25-1 & 8.6711 & 62.9888 & 0.4839 & 1.3561 & 1.2626 & 52.6536 & 51.0475 \\
Q35 & 4.7968 & 38.0587 & 1.1746 & 1.6948 & 1.5656 & 30.2374 & 24.8603 \\
IV35 & 6.0810 & 50.2095 & 0.8080 & 1.4560 & 1.3226 & 38.1983 & 33.1704 \\
ESC-RL & 8.9246 & 69.4134 & 0.3757 & 1.3547 & 1.2318 & 56.2151 & 54.0503 \\
MPO & 9.3149 & 67.2486 & 0.4351 & 1.3715 & 1.2584 & 53.8408 & 51.0475 \\
EditGRPO & 9.4260 & 70.8101 & 0.3638 & 1.3520 & 1.2381 & 56.3547 & 54.1899 \\
DPO-G & 6.7989 & 59.9860 & 0.5587 & 1.3729 & 1.2277 & 49.1620 & 44.8324 \\
R2Gen & 7.3296 & 64.5251 & 0.5000 & 1.3785 & 1.2549 & 49.5112 & 46.5084 \\
R2CMN & 7.6683 & 64.8743 & 0.4735 & 1.3568 & 1.2451 & 53.2821 & 49.5810 \\
L15 & 8.7570 & 70.2514 & 0.3673 & 1.3254 & 1.2263 & 57.4022 & 54.8184 \\
LN-M & 9.4637 & 72.4162 & 0.3352 & 1.2989 & 1.2200 & 56.4944 & 56.0056 \\
LN-V & 8.6913 & 71.5782 & 0.3554 & 1.2989 & 1.2193 & 56.7737 & 54.9581 \\
\bottomrule
\end{tabular}
\caption{MIMIC: candidate-oracle diagnostics. Errors are mismatched scored labels per input, not normalized Hamming rates. Missed denotes a strictly better available candidate under masked Hamming error.}\label{tab:app-oracle-mimic}
\end{table*}

\begin{table*}[!tp]
\centering\scriptsize
\setlength{\tabcolsep}{3pt}\renewcommand{\arraystretch}{1.0}
\begin{minipage}[t]{0.48\textwidth}
\centering
\begin{tabular}{@{}cc|cccc@{}}
\toprule
\textbf{Generator} & \textbf{Cond.} & \textbf{Ma-14} & \textbf{Mi-14} & \textbf{Ma-5} & \textbf{Mi-5}\\
\midrule
Q25-3 & Oracle & 0.6822 & 0.8797 & 0.8344 & 0.9016 \\
Q25-3 & Macro C & 0.4722 & 0.6465 & 0.6158 & 0.6602 \\
Q25-3 & Micro C & 0.4382 & 0.6683 & 0.5992 & 0.6752 \\
Q25-1 & Oracle & 0.5996 & 0.8520 & 0.7825 & 0.8675 \\
Q25-1 & Macro C & 0.4607 & 0.6318 & 0.5938 & 0.6443 \\
Q25-1 & Micro C & 0.4073 & 0.6516 & 0.5808 & 0.6620 \\
Q35 & Oracle & 0.5539 & 0.6955 & 0.7296 & 0.7536 \\
Q35 & Macro C & 0.4586 & 0.5997 & 0.5829 & 0.6214 \\
Q35 & Micro C & 0.4338 & 0.6108 & 0.5759 & 0.6273 \\
IV35 & Oracle & 0.5996 & 0.7705 & 0.7612 & 0.7849 \\
IV35 & Macro C & 0.4668 & 0.6306 & 0.5933 & 0.6262 \\
IV35 & Micro C & 0.4269 & 0.6473 & 0.6016 & 0.6466 \\
ESC-RL & Oracle & 0.7087 & 0.8888 & 0.8531 & 0.9127 \\
ESC-RL & Macro C & 0.4838 & 0.6436 & 0.6128 & 0.6573 \\
ESC-RL & Micro C & 0.4520 & 0.6684 & 0.6122 & 0.6780 \\
MPO & Oracle & 0.7039 & 0.8717 & 0.8442 & 0.8931 \\
MPO & Macro C & 0.4771 & 0.6399 & 0.6080 & 0.6526 \\
MPO & Micro C & 0.4333 & 0.6597 & 0.5814 & 0.6612 \\
EditGRPO & Oracle & 0.7371 & 0.8924 & 0.8346 & 0.9022 \\
EditGRPO & Macro C & 0.4849 & 0.6474 & 0.6112 & 0.6619 \\
EditGRPO & Micro C & 0.4478 & 0.6677 & 0.5978 & 0.6779 \\
\bottomrule\end{tabular}
\end{minipage}\hfill
\begin{minipage}[t]{0.48\textwidth}
\centering
\begin{tabular}{@{}cc|cccc@{}}
\toprule
\textbf{Generator} & \textbf{Cond.} & \textbf{Ma-14} & \textbf{Mi-14} & \textbf{Ma-5} & \textbf{Mi-5}\\
\midrule
DPO-G & Oracle & 0.6557 & 0.8313 & 0.7275 & 0.8115 \\
DPO-G & Macro C & 0.4657 & 0.6288 & 0.5791 & 0.6347 \\
DPO-G & Micro C & 0.4390 & 0.6594 & 0.5759 & 0.6586 \\
R2Gen & Oracle & 0.6493 & 0.8522 & 0.8349 & 0.8832 \\
R2Gen & Macro C & 0.4626 & 0.6305 & 0.5947 & 0.6416 \\
R2Gen & Micro C & 0.4300 & 0.6563 & 0.5808 & 0.6623 \\
R2CMN & Oracle & 0.6556 & 0.8568 & 0.8116 & 0.8801 \\
R2CMN & Macro C & 0.4735 & 0.6344 & 0.6119 & 0.6553 \\
R2CMN & Micro C & 0.4267 & 0.6576 & 0.5875 & 0.6661 \\
L15 & Oracle & 0.7542 & 0.8916 & 0.8544 & 0.9023 \\
L15 & Macro C & 0.5126 & 0.6547 & 0.6290 & 0.6715 \\
L15 & Micro C & 0.4586 & 0.6698 & 0.6066 & 0.6783 \\
LN-M & Oracle & 0.7780 & 0.9022 & 0.8799 & 0.9082 \\
LN-M & Macro C & 0.5201 & 0.6607 & 0.6347 & 0.6738 \\
LN-M & Micro C & 0.4697 & 0.6706 & 0.6008 & 0.6725 \\
LN-V & Oracle & 0.7683 & 0.8958 & 0.8574 & 0.9049 \\
LN-V & Macro C & 0.5190 & 0.6592 & 0.6295 & 0.6739 \\
LN-V & Micro C & 0.4661 & 0.6702 & 0.6013 & 0.6743 \\
\bottomrule\end{tabular}
\end{minipage}
\caption{MIMIC: F1 of the Hamming-minimizing oracle and constrained selections. Oracle F1 is diagnostic and is not an F1 upper bound.}\label{tab:app-oracle-f1-mimic}
\end{table*}

\begin{table*}[!htbp]
\centering\scriptsize
\setlength{\tabcolsep}{4pt}\renewcommand{\arraystretch}{1.0}
\begin{tabular}{@{}cccccccc@{}}
\toprule
\textbf{Generator} & \textbf{Unique} & \textbf{Zero error (\%)} & \textbf{Oracle errors} & \textbf{Ma errors} & \textbf{Mi errors} & \textbf{Ma missed (\%)} & \textbf{Mi missed (\%)}\\
\midrule
Q35 & 2.0673 & 67.0404 & 0.6009 & 0.9709 & 0.8072 & 16.1435 & 13.4529 \\
IV35 & 3.2601 & 71.3004 & 0.4753 & 0.9664 & 0.7691 & 23.0942 & 18.1614 \\
Q25-3 & 3.9776 & 73.9910 & 0.3498 & 1.5179 & 0.8274 & 61.6592 & 33.6323 \\
Q25-1 & 2.9641 & 68.6099 & 0.5022 & 1.7892 & 0.8274 & 73.0942 & 27.3543 \\
ESC-RL & 4.1143 & 76.9058 & 0.3251 & 1.5785 & 0.8117 & 66.1435 & 33.1839 \\
MPO & 3.5000 & 72.1973 & 0.3969 & 1.2578 & 0.8072 & 48.2063 & 29.3722 \\
R2Gen & 2.0538 & 66.1435 & 0.6659 & 0.9821 & 0.8341 & 18.1614 & 11.6592 \\
EditGRPO & 2.1502 & 67.2646 & 0.4731 & 0.8475 & 0.8206 & 33.8565 & 34.0807 \\
DPO-G & 2.2892 & 67.7130 & 0.4933 & 0.9753 & 0.8117 & 36.7713 & 29.3722 \\
R2CMN & 3.5628 & 71.0762 & 0.4170 & 1.5583 & 0.8274 & 57.6233 & 28.2511 \\
L15 & 2.9798 & 71.0762 & 0.5291 & 1.6502 & 0.8229 & 56.7265 & 16.1435 \\
LN-M & 3.7108 & 68.3857 & 0.4395 & 1.1054 & 0.8453 & 42.1525 & 34.5291 \\
LN-V & 3.9305 & 74.4395 & 0.4350 & 1.9933 & 0.8184 & 79.8206 & 21.0762 \\
\bottomrule
\end{tabular}
\caption{IU: candidate-oracle diagnostics. Errors are mismatched scored labels per input, not normalized Hamming rates. Missed denotes a strictly better available candidate under masked Hamming error.}\label{tab:app-oracle-iu}
\end{table*}

\begin{table*}[!tp]
\centering\scriptsize
\setlength{\tabcolsep}{3pt}\renewcommand{\arraystretch}{1.0}
\begin{minipage}[t]{0.48\textwidth}
\centering
\begin{tabular}{@{}cc|cccc@{}}
\toprule
\textbf{Generator} & \textbf{Cond.} & \textbf{Ma-14} & \textbf{Mi-14} & \textbf{Ma-5} & \textbf{Mi-5}\\
\midrule
Q35 & Oracle & 0.3715 & 0.7248 & 0.5009 & 0.5793 \\
Q35 & Macro C & 0.2530 & 0.6110 & 0.2581 & 0.3779 \\
Q35 & Micro C & 0.2499 & 0.6386 & 0.3146 & 0.3151 \\
IV35 & Oracle & 0.4754 & 0.7850 & 0.5556 & 0.7200 \\
IV35 & Macro C & 0.2656 & 0.6114 & 0.3044 & 0.4071 \\
IV35 & Micro C & 0.2479 & 0.6567 & 0.3641 & 0.4217 \\
Q25-3 & Oracle & 0.3365 & 0.8223 & 0.2586 & 0.4615 \\
Q25-3 & Macro C & 0.1459 & 0.3446 & 0.1144 & 0.1806 \\
Q25-3 & Micro C & 0.1207 & 0.6168 & 0.0877 & 0.1774 \\
Q25-1 & Oracle & 0.1766 & 0.7346 & 0.0000 & 0.0000 \\
Q25-1 & Macro C & 0.0582 & 0.1807 & 0.0000 & 0.0000 \\
Q25-1 & Micro C & 0.0665 & 0.6019 & 0.0000 & 0.0000 \\
ESC-RL & Oracle & 0.3461 & 0.8354 & 0.2928 & 0.6099 \\
ESC-RL & Macro C & 0.1221 & 0.3057 & 0.1058 & 0.1600 \\
ESC-RL & Micro C & 0.1255 & 0.6245 & 0.1308 & 0.2615 \\
MPO & Oracle & 0.3089 & 0.7991 & 0.3022 & 0.3937 \\
MPO & Macro C & 0.1603 & 0.4537 & 0.2011 & 0.2454 \\
MPO & Micro C & 0.1199 & 0.6203 & 0.0778 & 0.1296 \\
R2Gen & Oracle & 0.1406 & 0.6761 & 0.1299 & 0.3968 \\
R2Gen & Macro C & 0.0828 & 0.5370 & 0.0547 & 0.1576 \\
R2Gen & Micro C & 0.0773 & 0.6034 & 0.0529 & 0.1525 \\
\bottomrule\end{tabular}
\end{minipage}\hfill
\begin{minipage}[t]{0.48\textwidth}
\centering
\begin{tabular}{@{}cc|cccc@{}}
\toprule
\textbf{Generator} & \textbf{Cond.} & \textbf{Ma-14} & \textbf{Mi-14} & \textbf{Ma-5} & \textbf{Mi-5}\\
\midrule
EditGRPO & Oracle & 0.0994 & 0.7305 & 0.0581 & 0.0792 \\
EditGRPO & Macro C & 0.0877 & 0.5837 & 0.0496 & 0.0702 \\
EditGRPO & Micro C & 0.0815 & 0.6056 & 0.0572 & 0.0769 \\
DPO-G & Oracle & 0.1610 & 0.7317 & 0.0571 & 0.1509 \\
DPO-G & Macro C & 0.1081 & 0.5397 & 0.0754 & 0.1186 \\
DPO-G & Micro C & 0.0799 & 0.6108 & 0.0400 & 0.1091 \\
R2CMN & Oracle & 0.2425 & 0.7872 & 0.1728 & 0.3471 \\
R2CMN & Macro C & 0.1019 & 0.3375 & 0.1039 & 0.1771 \\
R2CMN & Micro C & 0.1078 & 0.6152 & 0.0867 & 0.1983 \\
L15 & Oracle & 0.2143 & 0.7479 & 0.0645 & 0.0980 \\
L15 & Macro C & 0.0521 & 0.2236 & 0.0000 & 0.0000 \\
L15 & Micro C & 0.0808 & 0.6108 & 0.0000 & 0.0000 \\
LN-M & Oracle & 0.1468 & 0.7586 & 0.0000 & 0.0000 \\
LN-M & Macro C & 0.0849 & 0.5200 & 0.0077 & 0.0182 \\
LN-M & Micro C & 0.0788 & 0.6044 & 0.0000 & 0.0000 \\
LN-V & Oracle & 0.2807 & 0.7905 & 0.0444 & 0.1165 \\
LN-V & Macro C & 0.0227 & 0.0553 & 0.0291 & 0.0741 \\
LN-V & Micro C & 0.0866 & 0.6129 & 0.0157 & 0.0400 \\
\bottomrule\end{tabular}
\end{minipage}
\caption{IU: F1 of the Hamming-minimizing oracle and constrained selections. Oracle F1 is diagnostic and is not an F1 upper bound.}\label{tab:app-oracle-f1-iu}
\end{table*}

\section{Additional Results for Text Realization}\label{app:rq3}

RQ3 is restricted to Qwen2.5-VL-7B (3 epoch) on both benchmarks.
All targets match RQ2, and retrieval and generation share the same target within each
condition. Tables report point estimates, not bootstrap significance.
Constrained verbatim belongs to our framework: it outputs the original macro-oriented
constrained selection, whereas the twelve realization configurations transform targets
into new reports or retrieve training reports.

\subsection{Absolute Targets and Realized Findings}
\label{app:rq3-absolute}
Tables~\ref{tab:app-rq3-abs-mimic}, \ref{tab:app-rq3-abs-iu} report decision-level targets and the findings extracted from retrieved and generated reports.

\begin{table*}[!tp]
\centering
\begin{minipage}[t]{0.48\textwidth}
\centering\scriptsize

\setlength{\tabcolsep}{1.5pt}\renewcommand{\arraystretch}{1.0}

\begin{tabular}{@{}ccccccc@{}}\toprule

\textbf{Obj.} & \textbf{Target} & \textbf{Stage} & \textbf{Ma-14} & \textbf{Mi-14} & \textbf{Ma-5} & \textbf{Mi-5}\\ \midrule

macro & C & D & .4722 & .6465 & .6158 & .6602 \\

macro & C & R & .4607 & .6353 & .6010 & .6464 \\

macro & C & G & .4567 & .6370 & .6158 & .6604 \\

macro & P & D & .4937 & .6303 & .6030 & .6469 \\

macro & P & R & .4673 & .6154 & .5907 & .6335 \\

macro & P & G & .4792 & .6220 & .6005 & .6484 \\

macro & F & D & .5030 & .6311 & .6214 & .6664 \\

macro & F & R & .4761 & .6205 & .5982 & .6487 \\

macro & F & G & .4937 & .6332 & .6184 & .6653 \\

\addlinespace[3pt]
micro & C & D & .4382 & .6683 & .5992 & .6752 \\

micro & C & R & .4239 & .6499 & .5868 & .6583 \\

micro & C & G & .4248 & .6505 & .5966 & .6718 \\

micro & P & D & .4602 & .6673 & .6152 & .6729 \\

micro & P & R & .4381 & .6466 & .6010 & .6561 \\

micro & P & G & .4416 & .6502 & .6078 & .6706 \\

micro & F & D & .4625 & .6661 & .6207 & .6772 \\

micro & F & R & .4517 & .6466 & .6124 & .6616 \\

micro & F & G & .4424 & .6505 & .6100 & .6717 \\

\bottomrule\end{tabular}

\caption{MIMIC: full Qwen2.5-VL-7B (3 epoch) finding results. D/R/G denote decision/retrieval/generation.}\label{tab:app-rq3-abs-mimic}
\end{minipage}
\hfill
\begin{minipage}[t]{0.48\textwidth}
\centering\scriptsize

\setlength{\tabcolsep}{1.5pt}\renewcommand{\arraystretch}{1.0}

\begin{tabular}{@{}ccccccc@{}}\toprule

\textbf{Obj.} & \textbf{Target} & \textbf{Stage} & \textbf{Ma-14} & \textbf{Mi-14} & \textbf{Ma-5} & \textbf{Mi-5}\\ \midrule

macro & C & D & .1459 & .3446 & .1144 & .1806 \\

macro & C & R & .1301 & .3491 & .1195 & .1867 \\

macro & C & G & .1271 & .3783 & .1094 & .1750 \\

macro & P & D & .1353 & .1471 & .1200 & .0393 \\

macro & P & R & .1001 & .0619 & .1462 & .0617 \\

macro & P & G & .0699 & .0630 & .0884 & .0599 \\

macro & F & D & .1657 & .1787 & .1639 & .0936 \\

macro & F & R & .1035 & .0994 & .1585 & .1121 \\

macro & F & G & .0770 & .0813 & .1272 & .1255 \\

\addlinespace[3pt]
micro & C & D & .1207 & .6168 & .0877 & .1774 \\

micro & C & R & .1207 & .6175 & .0877 & .1774 \\

micro & C & G & .1212 & .6207 & .0885 & .1789 \\

micro & P & D & .1207 & .6181 & .0877 & .1789 \\

micro & P & R & .1207 & .6181 & .0877 & .1789 \\

micro & P & G & .1183 & .6191 & .0885 & .1803 \\

micro & F & D & .1386 & .6192 & .1575 & .3563 \\

micro & F & R & .1389 & .6256 & .1558 & .3452 \\

micro & F & G & .1394 & .6260 & .1590 & .3584 \\

\bottomrule\end{tabular}

\caption{IU: full Qwen2.5-VL-7B (3 epoch) finding results. D/R/G denote decision/retrieval/generation.}\label{tab:app-rq3-abs-iu}
\end{minipage}
\end{table*}

\subsection{Fidelity and Full Report Metrics}
Target fidelity and reference-based report metrics are reported in Tables~\ref{tab:app-rq3-quality-mimic}, \ref{tab:app-rq3-quality-iu}.

\begin{table*}[!htbp]\centering\scriptsize

\setlength{\tabcolsep}{3pt}\renewcommand{\arraystretch}{1.0}

\begin{tabular}{@{}ccccccccccc@{}}\toprule

\textbf{Objective} & \textbf{Target} & \textbf{Route} & \textbf{Ret.} & \textbf{Ins.} & \textbf{Ham.} & \textbf{GREEN} & \textbf{RG} & \textbf{RaTE} & \textbf{RL} & \textbf{B4}\\ \midrule

macro & C & R & .8946 & .0360 & .0211 & .4375 & .2722 & .6010 & .2886 & .0799 \\

macro & C & G & .8919 & .0309 & .0208 & .4890 & .3200 & .6453 & .3575 & .1161 \\

macro & P & R & .8702 & .0315 & .0262 & .4274 & .2687 & .5992 & .2864 & .0777 \\

macro & P & G & .8645 & .0271 & .0264 & .4817 & .3147 & .6434 & .3542 & .1167 \\

macro & F & R & .8286 & .0284 & .0375 & .4243 & .2647 & .5973 & .2825 & .0723 \\

macro & F & G & .7958 & .0221 & .0426 & .4813 & .3156 & .6442 & .3529 & .1184 \\

\addlinespace[3pt]
micro & C & R & .9222 & .0377 & .0164 & .4389 & .2690 & .5990 & .2884 & .0782 \\

micro & C & G & .9219 & .0345 & .0160 & .4841 & .3175 & .6458 & .3583 & .1150 \\

micro & P & R & .9196 & .0366 & .0168 & .4392 & .2688 & .5991 & .2886 & .0781 \\

micro & P & G & .9172 & .0339 & .0168 & .4843 & .3164 & .6450 & .3574 & .1140 \\

micro & F & R & .8991 & .0362 & .0207 & .4347 & .2680 & .5981 & .2869 & .0769 \\

micro & F & G & .8984 & .0305 & .0200 & .4823 & .3162 & .6441 & .3571 & .1158 \\

\bottomrule\end{tabular}

\caption{MIMIC: fidelity and all five report metrics for the twelve main realization configurations.}\label{tab:app-rq3-quality-mimic}\end{table*}

\begin{table*}[!htbp]\centering\scriptsize

\setlength{\tabcolsep}{3pt}\renewcommand{\arraystretch}{1.0}

\begin{tabular}{@{}ccccccccccc@{}}\toprule

\textbf{Objective} & \textbf{Target} & \textbf{Route} & \textbf{Ret.} & \textbf{Ins.} & \textbf{Ham.} & \textbf{GREEN} & \textbf{RG} & \textbf{RaTE} & \textbf{RL} & \textbf{B4}\\ \midrule

macro & C & R & .9634 & .0019 & .0034 & .4867 & .2617 & .5680 & .2707 & .0494 \\

macro & C & G & .8626 & .0673 & .0175 & .4963 & .3154 & .5953 & .3329 & .0961 \\

macro & P & R & .4443 & .0031 & .1954 & .4367 & .2182 & .5460 & .2303 & .0277 \\

macro & P & G & .4589 & .1876 & .2269 & .2920 & .2195 & .5162 & .2706 & .0612 \\

macro & F & R & .3639 & .0038 & .4444 & .0774 & .1049 & .4286 & .2083 & .0198 \\

macro & F & G & .3469 & .1332 & .4925 & .1981 & .1975 & .4991 & .2402 & .0454 \\

\addlinespace[3pt]
micro & C & R & .9979 & .0000 & .0002 & .5609 & .2882 & .5985 & .2844 & .0563 \\

micro & C & G & .9807 & .0065 & .0019 & .5378 & .3356 & .6065 & .3579 & .0998 \\

micro & P & R & 1.0000 & .0000 & .0000 & .5605 & .2879 & .5984 & .2842 & .0562 \\

micro & P & G & .9741 & .0088 & .0026 & .5401 & .3349 & .6065 & .3573 & .0988 \\

micro & F & R & .9177 & .0000 & .0074 & .5486 & .2856 & .5969 & .2826 & .0569 \\

micro & F & G & .8730 & .0041 & .0117 & .5346 & .3358 & .6108 & .3571 & .0986 \\

\bottomrule\end{tabular}

\caption{IU: fidelity and all five report metrics for the twelve main realization configurations.}\label{tab:app-rq3-quality-iu}\end{table*}

\subsection{Realization Gaps and Gain Changes}
Tables~\ref{tab:app-rq3-gap-mimic}, \ref{tab:app-rq3-attenuation-mimic}, \ref{tab:app-rq3-gap-iu}, \ref{tab:app-rq3-attenuation-iu} report realization gaps and changes in gains relative to constrained targets.

\begin{table*}[!tp]
\centering
\begin{minipage}[t]{0.48\textwidth}
\centering\scriptsize

\setlength{\tabcolsep}{1.5pt}\renewcommand{\arraystretch}{1.0}

\begin{tabular}{@{}ccccccc@{}}\toprule

\textbf{Obj.} & \textbf{Target} & \textbf{Route} & \textbf{Ma-14} & \textbf{Mi-14} & \textbf{Ma-5} & \textbf{Mi-5}\\ \midrule

macro & C & R & -.0116 & -.0111 & -.0148 & -.0138 \\

macro & C & G & -.0155 & -.0094 & .0000 & +.0002 \\

macro & P & R & -.0264 & -.0149 & -.0123 & -.0134 \\

macro & P & G & -.0145 & -.0082 & -.0025 & +.0015 \\

macro & F & R & -.0269 & -.0106 & -.0232 & -.0177 \\

macro & F & G & -.0093 & +.0021 & -.0030 & -.0011 \\

\addlinespace[3pt]
micro & C & R & -.0142 & -.0184 & -.0124 & -.0170 \\

micro & C & G & -.0134 & -.0178 & -.0026 & -.0034 \\

micro & P & R & -.0221 & -.0207 & -.0142 & -.0168 \\

micro & P & G & -.0186 & -.0171 & -.0074 & -.0022 \\

micro & F & R & -.0108 & -.0195 & -.0083 & -.0156 \\

micro & F & G & -.0200 & -.0155 & -.0107 & -.0055 \\

\bottomrule\end{tabular}

\caption{MIMIC: realization gaps $L_r=M_r-M_{\mathrm{dec}}$. Negative values indicate lower reference-based finding scores after realization.}\label{tab:app-rq3-gap-mimic}
\end{minipage}
\hfill
\begin{minipage}[t]{0.48\textwidth}
\centering\scriptsize

\setlength{\tabcolsep}{1.5pt}\renewcommand{\arraystretch}{1.0}

\begin{tabular}{@{}ccccccc@{}}\toprule

\textbf{Obj.} & \textbf{Contrast} & \textbf{Route} & \textbf{Ma-14} & \textbf{Mi-14} & \textbf{Ma-5} & \textbf{Mi-5}\\ \midrule

macro & P--C & R & -.0149 & -.0037 & +.0025 & +.0004 \\

macro & P--C & G & +.0010 & +.0012 & -.0025 & +.0013 \\

macro & F--C & R & -.0153 & +.0006 & -.0084 & -.0039 \\

macro & F--C & G & +.0062 & +.0115 & -.0030 & -.0013 \\

\addlinespace[3pt]
micro & P--C & R & -.0079 & -.0023 & -.0018 & +.0002 \\

micro & P--C & G & -.0052 & +.0007 & -.0048 & +.0012 \\

micro & F--C & R & +.0034 & -.0011 & +.0041 & +.0013 \\

micro & F--C & G & -.0066 & +.0022 & -.0081 & -.0021 \\

\bottomrule\end{tabular}

\caption{MIMIC: complete gain-change values $A_r$ for partial/free relative to constrained targets.}\label{tab:app-rq3-attenuation-mimic}
\end{minipage}
\end{table*}

\begin{table*}[!tp]
\centering
\begin{minipage}[t]{0.48\textwidth}
\centering\scriptsize

\setlength{\tabcolsep}{1.5pt}\renewcommand{\arraystretch}{1.0}

\begin{tabular}{@{}ccccccc@{}}\toprule

\textbf{Obj.} & \textbf{Target} & \textbf{Route} & \textbf{Ma-14} & \textbf{Mi-14} & \textbf{Ma-5} & \textbf{Mi-5}\\ \midrule

macro & C & R & -.0158 & +.0045 & +.0051 & +.0060 \\

macro & C & G & -.0188 & +.0336 & -.0051 & -.0056 \\

macro & P & R & -.0351 & -.0853 & +.0263 & +.0224 \\

macro & P & G & -.0654 & -.0841 & -.0315 & +.0206 \\

macro & F & R & -.0622 & -.0793 & -.0054 & +.0185 \\

macro & F & G & -.0887 & -.0974 & -.0367 & +.0319 \\

\addlinespace[3pt]
micro & C & R & .0000 & +.0006 & .0000 & .0000 \\

micro & C & G & +.0005 & +.0039 & +.0008 & +.0014 \\

micro & P & R & .0000 & .0000 & .0000 & .0000 \\

micro & P & G & -.0024 & +.0010 & +.0008 & +.0015 \\

micro & F & R & +.0003 & +.0064 & -.0017 & -.0111 \\

micro & F & G & +.0008 & +.0068 & +.0015 & +.0021 \\

\bottomrule\end{tabular}

\caption{IU: realization gaps $L_r=M_r-M_{\mathrm{dec}}$. Negative values indicate lower reference-based finding scores after realization.}\label{tab:app-rq3-gap-iu}
\end{minipage}
\hfill
\begin{minipage}[t]{0.48\textwidth}
\centering\scriptsize

\setlength{\tabcolsep}{1.5pt}\renewcommand{\arraystretch}{1.0}

\begin{tabular}{@{}ccccccc@{}}\toprule

\textbf{Obj.} & \textbf{Contrast} & \textbf{Route} & \textbf{Ma-14} & \textbf{Mi-14} & \textbf{Ma-5} & \textbf{Mi-5}\\ \midrule

macro & P--C & R & -.0193 & -.0897 & +.0212 & +.0164 \\

macro & P--C & G & -.0466 & -.1177 & -.0265 & +.0263 \\

macro & F--C & R & -.0464 & -.0838 & -.0105 & +.0125 \\

macro & F--C & G & -.0699 & -.1310 & -.0317 & +.0376 \\

\addlinespace[3pt]
micro & P--C & R & .0000 & -.0006 & .0000 & .0000 \\

micro & P--C & G & -.0029 & -.0029 & .0000 & .0000 \\

micro & F--C & R & +.0003 & +.0058 & -.0017 & -.0111 \\

micro & F--C & G & +.0003 & +.0030 & +.0007 & +.0006 \\

\bottomrule\end{tabular}

\caption{IU: complete gain-change values $A_r$ for partial/free relative to constrained targets.}\label{tab:app-rq3-attenuation-iu}
\end{minipage}
\end{table*}

\subsection{Reference-Aware Realization Change Decomposition}
The corresponding results are reported in Table~\ref{tab:rq3-change-decomposition}.

\label{app:rq3-decomposition}
This analysis is restricted to Qwen2.5-VL-7B (3 epoch), matching the main RQ3 scope.
On reference-scored cells, $D_{\mathrm{TP}}$ and $D_{\mathrm{FP}}$ count deleted
target true/false positives; $A_+$ and $A_-$ count added reference-positive/negative
findings. The accounting identities are
\begin{equation}
\begin{aligned}
\Delta TP&=A_+-D_{\mathrm{TP}},\\
\Delta FP&=A_--D_{\mathrm{FP}},\\
\Delta FN&=D_{\mathrm{TP}}-A_+.
\end{aligned}
\end{equation}
These automatic-label diagnostics are not clinician-verified errors.
Unlike target fidelity, they exclude reference-uncertain cells.
The identities and per-label aggregation are checked against exported counts.
Removing more false than true positives does not by itself imply F1 improvement;
macro effects additionally depend on the label distribution of the changes.
For MIMIC macro free generation, Mi-14 rises from 0.6311 to 0.6332 despite target
changes. For IU, the same condition deletes 2,448 false positives but loses
345 true positives and adds 220 false positives; Mi-14 falls from 0.1787 to 0.0813.

\begin{table*}[!tp]
\centering\scriptsize
\setlength{\tabcolsep}{1.5pt}\renewcommand{\arraystretch}{1.0}
\begin{tabular}{@{}ccccccc@{}}
\toprule
\textbf{Data} & \textbf{Obj.} & \textbf{T/R} & \textbf{$D_{\mathrm{TP}}$} & \textbf{$D_{\mathrm{FP}}$} & \textbf{$A_+$} & \textbf{$A_-$}\\
\midrule
MIMIC & macro & C/R & 121 & 167 & 31 & 69 \\
MIMIC & macro & C/G & 117 & 173 & 26 & 59 \\
MIMIC & macro & P/R & 158 & 235 & 23 & 69 \\
MIMIC & macro & P/G & 149 & 265 & 21 & 57 \\
MIMIC & macro & F/R & 206 & 381 & 19 & 70 \\
MIMIC & macro & F/G & 215 & 488 & 20 & 47 \\
MIMIC & micro & C/R & 107 & 103 & 22 & 77 \\
MIMIC & micro & C/G & 113 & 97 & 28 & 66 \\
MIMIC & micro & P/R & 117 & 102 & 22 & 75 \\
MIMIC & micro & P/G & 119 & 111 & 29 & 65 \\
MIMIC & micro & F/R & 140 & 149 & 25 & 74 \\
MIMIC & micro & F/G & 135 & 159 & 25 & 62 \\
IU & macro & C/R & 2 & 18 & 1 & 0 \\
IU & macro & C/G & 10 & 63 & 20 & 14 \\
IU & macro & P/R & 150 & 1049 & 0 & 3 \\
IU & macro & P/G & 162 & 1002 & 21 & 206 \\
IU & macro & F/R & 326 & 2406 & 1 & 5 \\
IU & macro & F/G & 345 & 2448 & 8 & 220 \\
IU & micro & C/R & 0 & 1 & 0 & 0 \\
IU & micro & C/G & 1 & 8 & 1 & 2 \\
IU & micro & P/R & 0 & 0 & 0 & 0 \\
IU & micro & P/G & 3 & 9 & 1 & 3 \\
IU & micro & F/R & 11 & 35 & 0 & 0 \\
IU & micro & F/G & 18 & 53 & 0 & 2 \\
\bottomrule
\end{tabular}
\caption{Pooled reference-scored changes for Qwen2.5-VL-7B (3 epoch): 1,432 MIMIC and 446 IU test inputs. C/P/F denote targets, R/G denote routes. Counts are descriptive.}\label{tab:rq3-change-decomposition}
\end{table*}

\subsection{Extended Report Benchmark and Computational Requirements}
\label{app:rq3-benchmark}
Tables~\ref{tab:app-rq3-benchmark-mimic} and~\ref{tab:app-rq3-benchmark-iu} report extended benchmarks.
Tables~\ref{tab:app-rq3-scale-mimic} and~\ref{tab:app-rq3-scale-iu} report alternate-generation-scale results.

The main analysis reports all configurations rather than replacing them with a
test-selected winner. Development-selected CCS and SDR variants use development Ma-14;
their test scores are evaluated after selection is frozen. The development-selected
ours row is included only as a supplementary deployment rule, with twelve configurations
searched using the same criterion. Random selection is excluded from this appendix.
Scale-ablation rows are not included in that configuration search.

\begin{table*}[!htbp]\centering\scriptsize

\setlength{\tabcolsep}{3pt}\renewcommand{\arraystretch}{1.0}

\begin{tabular}{@{}cccccccccc@{}}\toprule

\textbf{Method} & \textbf{Ma-14} & \textbf{Mi-14} & \textbf{Ma-5} & \textbf{Mi-5} & \textbf{GREEN} & \textbf{RG} & \textbf{RaTE} & \textbf{RL} & \textbf{B4}\\ \midrule

Plurality & .3900 & .6214 & .5229 & .6209 & .4947 & .3227 & .6307 & .3518 & .1124 \\

Most-typical & .3900 & .6214 & .5229 & .6209 & .4947 & .3227 & .6307 & .3518 & .1124 \\

Consensus-CB & .4005 & .6331 & .5338 & .6361 & .4876 & .3158 & .6425 & .3282 & .1016 \\

CCS-Emb & .3881 & .6092 & .5178 & .6059 & .4952 & .3313 & .6465 & .3617 & .1203 \\

Classifier probe & .4400 & .6163 & .5625 & .6348 & .4783 & .3106 & .6385 & .3234 & .1063 \\

CLIPScore & .4037 & .6046 & .5295 & .6101 & .4812 & .3118 & .6266 & .3346 & .1131 \\

MBR-1 & .3725 & .5903 & .5111 & .5775 & .4977 & .3382 & .6458 & .3728 & .1186 \\

MBR-2 & .3742 & .5833 & .5145 & .5762 & .4942 & .3347 & .6458 & .3766 & .1200 \\

Self-consistency & .3464 & .5629 & .4657 & .5449 & .4784 & .3114 & .6359 & .3241 & .0968 \\

Median length & .3677 & .5786 & .4727 & .5540 & .4839 & .3102 & .6285 & .3494 & .1100 \\

SDR-Dev & .3478 & .5511 & .4740 & .5426 & .4689 & .2914 & .6135 & .3243 & .0930 \\

CCS-Dev & .3789 & .5898 & .5145 & .5824 & .4970 & .3310 & .6463 & .3747 & .1220 \\

Ours: macro C verbatim & .4562 & .6306 & .5799 & .6390 & .4845 & .3115 & .6437 & .3246 & .1055 \\

Greedy & .3531 & .5807 & .4712 & .5729 & .4866 & .3146 & .6336 & .3545 & .1131 \\

Ours: dev-selected & .4761 & .6205 & .5982 & .6487 & .4243 & .2647 & .5973 & .2825 & .0723 \\

\bottomrule\end{tabular}

\caption{MIMIC: extended benchmark, including all five reference-based report metrics and additional baselines.}\label{tab:app-rq3-benchmark-mimic}\end{table*}

\begin{table*}[!htbp]\centering\scriptsize

\setlength{\tabcolsep}{3pt}\renewcommand{\arraystretch}{1.0}

\begin{tabular}{@{}cccccccccc@{}}\toprule

\textbf{Configuration} & \textbf{Ma-14} & \textbf{Mi-14} & \textbf{Ma-5} & \textbf{Mi-5} & \textbf{GREEN} & \textbf{RG} & \textbf{RaTE} & \textbf{RL} & \textbf{B4}\\ \midrule

Macro, C $\to$ G & .4334 & .6260 & .5901 & .6556 & .4818 & .3428 & .6454 & .3613 & .1195 \\

Macro, P $\to$ G & .4669 & .6198 & .5843 & .6444 & .4796 & .3419 & .6452 & .3586 & .1208 \\

Macro, F $\to$ G & .4777 & .6338 & .6031 & .6622 & .4797 & .3417 & .6469 & .3585 & .1231 \\

Micro, C $\to$ G & .4130 & .6415 & .5627 & .6705 & .4818 & .3436 & .6450 & .3622 & .1194 \\

Micro, P $\to$ G & .4315 & .6402 & .5811 & .6684 & .4818 & .3424 & .6440 & .3619 & .1200 \\

Micro, F $\to$ G & .4287 & .6408 & .5803 & .6721 & .4843 & .3431 & .6448 & .3615 & .1207 \\

\bottomrule\end{tabular}

\caption{MIMIC: alternate-generation-scale results. These are supplementary and excluded from development configuration selection; the implementation notes identify a 1.5B image-label realization checkpoint (see Appendix~\ref{app:realization-implementation}).}\label{tab:app-rq3-scale-mimic}\end{table*}

\begin{table*}[!htbp]\centering\scriptsize

\setlength{\tabcolsep}{3pt}\renewcommand{\arraystretch}{1.0}

\begin{tabular}{@{}cccccccccc@{}}\toprule

\textbf{Method} & \textbf{Ma-14} & \textbf{Mi-14} & \textbf{Ma-5} & \textbf{Mi-5} & \textbf{GREEN} & \textbf{RG} & \textbf{RaTE} & \textbf{RL} & \textbf{B4}\\ \midrule

Plurality & .0586 & .5940 & .0082 & .0202 & .5678 & .3129 & .6373 & .3009 & .0523 \\

Most-typical & .0875 & .5966 & .0082 & .0202 & .6028 & .3810 & .6510 & .3397 & .0751 \\

Consensus-CB & .0663 & .5924 & .0000 & .0000 & .5668 & .3122 & .6370 & .3003 & .0511 \\

CCS-Emb & .0775 & .5842 & .0000 & .0000 & .6022 & .3740 & .6454 & .3383 & .0785 \\

Classifier probe & .1266 & .5156 & .1046 & .1795 & .5259 & .2977 & .6131 & .2930 & .0578 \\

CLIPScore & .1762 & .5635 & .0987 & .1653 & .5689 & .3257 & .6222 & .3262 & .0796 \\

MBR-1 & .0544 & .5825 & .0000 & .0000 & .5993 & .3960 & .6565 & .3588 & .0860 \\

MBR-2 & .0629 & .5843 & .0000 & .0000 & .5692 & .3585 & .6301 & .3571 & .0956 \\

Self-consistency & .0982 & .5806 & .0565 & .1197 & .5564 & .3061 & .6330 & .2980 & .0508 \\

Median length & .0680 & .5785 & .0370 & .0962 & .5955 & .3270 & .6241 & .3311 & .0731 \\

SDR-Dev & .0562 & .5046 & .0157 & .0396 & .6255 & .3543 & .6357 & .3337 & .0795 \\

CCS-Dev & .1022 & .5787 & .0707 & .1565 & .5553 & .2997 & .6300 & .2962 & .0496 \\

Ours: macro C verbatim & .2105 & .5596 & .1529 & .2667 & .5369 & .3027 & .6214 & .2939 & .0624 \\

Greedy & .1020 & .5521 & .0467 & .0727 & .5731 & .3122 & .5947 & .3360 & .0855 \\

Ours: dev-selected & .1389 & .6256 & .1558 & .3452 & .5486 & .2856 & .5969 & .2826 & .0569 \\

\bottomrule\end{tabular}

\caption{IU: extended benchmark, including all five reference-based report metrics and additional baselines.}\label{tab:app-rq3-benchmark-iu}\end{table*}

\begin{table*}[!htbp]\centering\scriptsize

\setlength{\tabcolsep}{3pt}\renewcommand{\arraystretch}{1.0}

\begin{tabular}{@{}cccccccccc@{}}\toprule

\textbf{Configuration} & \textbf{Ma-14} & \textbf{Mi-14} & \textbf{Ma-5} & \textbf{Mi-5} & \textbf{GREEN} & \textbf{RG} & \textbf{RaTE} & \textbf{RL} & \textbf{B4}\\ \midrule

Macro, C $\to$ G & .1565 & .5549 & .1663 & .3425 & .5822 & .3772 & .6436 & .3551 & .0947 \\

Macro, P $\to$ G & .0747 & .1387 & .1050 & .1714 & .2947 & .2459 & .5138 & .2674 & .0581 \\

Macro, F $\to$ G & .0854 & .1258 & .1416 & .2183 & .2197 & .2327 & .5092 & .2520 & .0514 \\

Micro, C $\to$ G & .1275 & .6262 & .1510 & .3239 & .6033 & .4002 & .6667 & .3670 & .0934 \\

Micro, P $\to$ G & .1275 & .6275 & .1510 & .3262 & .6072 & .4020 & .6679 & .3675 & .0938 \\

Micro, F $\to$ G & .1402 & .6203 & .2131 & .3676 & .5983 & .3992 & .6643 & .3667 & .0948 \\

\bottomrule\end{tabular}

\caption{IU: alternate-generation-scale results. These are supplementary and excluded from development configuration selection; the implementation notes identify a 1.5B image-label realization checkpoint (see Appendix~\ref{app:realization-implementation}).}\label{tab:app-rq3-scale-iu}\end{table*}

\subsection{Development Selection Records}
\label{app:dev-selection}
Tables~\ref{tab:app-rq3-dev-mimic}, \ref{tab:app-rq3-dev-iu} list the complete development selection records for MIMIC and IU.

\begin{table*}[!tp]
\centering
\begin{minipage}[t]{0.48\textwidth}
\centering\scriptsize

\setlength{\tabcolsep}{1.5pt}\renewcommand{\arraystretch}{1.0}

\begin{tabular}{@{}cccc@{}}\toprule

\textbf{Family} & \textbf{Variant} & \textbf{Dev} & \textbf{Sel.}\\ \midrule

SDR & Cham.-avg & 0.342217 &  \\

SDR & Cham.-knn & 0.310034 &  \\

SDR & Cham.-min & 0.304517 &  \\

SDR & Haus.-avg & 0.336153 &  \\

SDR & Haus.-knn & 0.306052 &  \\

SDR & Haus.-min & 0.304919 &  \\

SDR & Hung.-avg & 0.303909 &  \\

SDR & Hung.-knn & 0.320099 &  \\

SDR & Hung.-min & 0.346134 & Yes \\

CCS & chexbert14 & 0.372566 &  \\

CCS & chexbert5 & 0.348229 &  \\

CCS & rougeL & 0.375075 &  \\

CCS & bleu4 & 0.380896 & Yes \\

CCS & embed & 0.375046 &  \\

CCS & bertscore & 0.350811 &  \\

CCS & radgraph & 0.362794 &  \\

CCS & ratescore & 0.352996 &  \\

CCS & radeval\_bert & 0.351614 &  \\

CCS & best & 0.375046 &  \\

Ours & Macro, C $\to$ R & 0.473436 &  \\

Ours & Macro, C $\to$ G & 0.466512 &  \\

Ours & Macro, P $\to$ R & 0.488058 &  \\

Ours & Macro, P $\to$ G & 0.490667 &  \\

Ours & Macro, F $\to$ R & 0.507305 & Yes \\

Ours & Macro, F $\to$ G & 0.494486 &  \\

Ours & Micro, C $\to$ R & 0.431744 &  \\

Ours & Micro, C $\to$ G & 0.436125 &  \\

Ours & Micro, P $\to$ R & 0.439162 &  \\

Ours & Micro, P $\to$ G & 0.446720 &  \\

Ours & Micro, F $\to$ R & 0.446516 &  \\

Ours & Micro, F $\to$ G & 0.443025 &  \\

\bottomrule\end{tabular}

\caption{MIMIC: complete development selection records. Six decimals retain the small difference between the leading IU configurations.}\label{tab:app-rq3-dev-mimic}
\end{minipage}
\hfill
\begin{minipage}[t]{0.48\textwidth}
\centering\scriptsize

\setlength{\tabcolsep}{1.5pt}\renewcommand{\arraystretch}{1.0}

\begin{tabular}{@{}cccc@{}}\toprule

\textbf{Family} & \textbf{Variant} & \textbf{Dev} & \textbf{Sel.}\\ \midrule

SDR & Cham.-avg & 0.059820 &  \\

SDR & Cham.-knn & 0.051774 &  \\

SDR & Cham.-min & 0.057812 &  \\

SDR & Haus.-avg & 0.061546 & Yes \\

SDR & Haus.-knn & 0.051641 &  \\

SDR & Haus.-min & 0.051601 &  \\

SDR & Hung.-avg & 0.050881 &  \\

SDR & Hung.-knn & 0.051429 &  \\

SDR & Hung.-min & 0.051429 &  \\

CCS & chexbert14 & 0.061173 &  \\

CCS & chexbert5 & 0.066396 &  \\

CCS & rougeL & 0.055240 &  \\

CCS & bleu4 & 0.055102 &  \\

CCS & embed & 0.055650 &  \\

CCS & bertscore & 0.051409 &  \\

CCS & radgraph & 0.052124 &  \\

CCS & ratescore & 0.055091 &  \\

CCS & radeval\_bert & 0.092002 & Yes \\

CCS & best & 0.092002 &  \\

Ours & Macro, C $\to$ R & 0.148213 &  \\

Ours & Macro, C $\to$ G & 0.138365 &  \\

Ours & Macro, P $\to$ R & 0.080907 &  \\

Ours & Macro, P $\to$ G & 0.068323 &  \\

Ours & Macro, F $\to$ R & 0.143220 &  \\

Ours & Macro, F $\to$ G & 0.092045 &  \\

Ours & Micro, C $\to$ R & 0.113436 &  \\

Ours & Micro, C $\to$ G & 0.104375 &  \\

Ours & Micro, P $\to$ R & 0.113436 &  \\

Ours & Micro, P $\to$ G & 0.104508 &  \\

Ours & Micro, F $\to$ R & 0.149160 & Yes \\

Ours & Micro, F $\to$ G & 0.148962 &  \\

\bottomrule\end{tabular}

\caption{IU: complete development selection records. Six decimals retain the small difference between the leading IU configurations.}\label{tab:app-rq3-dev-iu}
\end{minipage}
\end{table*}

\paragraph{Computational requirements.}
Greedy uses one generator pass. Sample-based selectors use the same pool of
$N=32$ sampled reports. Retrieval adds a search over training reports;
generation adds one realization-model pass. Classifier and CLIP-based reranking \cite{hessel-etal-2021-clipscore}
require their respective auxiliary models, and SDR uses set-distance representations.
These are procedural cost descriptions, not measured latency, memory, or FLOP comparisons.

\end{document}